\documentclass[a4paper,fleqn]{cas-dc}

\usepackage[numbers]{natbib}
\usepackage{booktabs}
\usepackage{multirow}
\usepackage{algpseudocode} % 伪代码格式
\usepackage{algorithm}
\usepackage{booktabs}   % 表格格式、
\usepackage{xcolor}  % 颜色支持
\usepackage{bm}
\usepackage{url}
\usepackage{verbatim}
\usepackage{graphicx}
\usepackage{hyperref}
\hypersetup{
	colorlinks=true,
	linkcolor=black
}

\def\tsc#1{\csdef{#1}{\textsc{\lowercase{#1}}\xspace}}
\tsc{WGM}
\tsc{QE}
\tsc{EP}
\tsc{PMS}
\tsc{BEC}
\tsc{DE}
\begin{document}
	
	% 定义颜色
	\definecolor{keywordcolor}{RGB}{0, 120, 0} % 关键词颜色
	\definecolor{commentcolor}{RGB}{128, 128, 128} % 注释颜色
	\definecolor{stringcolor}{RGB}{0, 0, 0} % 变量和操作符颜色

	\let\WriteBookmarks\relax
	\def\floatpagepagefraction{1}
	\def\textpagefraction{.001}
	\shorttitle{FMaMIL: Frequency-Driven Mamba MIL}
	\shortauthors{Hangbei Cheng et~al.}
	
	\title [mode = title]{FMaMIL: Frequency-Driven Mamba Multi-Instance Learning for Weakly Supervised Lesion Segmentation in Medical Images}                     
	\tnotemark[1]
	
	\author[1]{Hangbei Cheng}[type=editor,
	auid=000,bioid=1,
	orcid=0009-0006-7310-7636]
	\ead{chenghangbei0702@163.com}
	\credit{Conceptualization, Investigation, Formal analysis, Writing – Original Draft, Visualization}
	
	\author[1]{Xiaorong Dong}[type=editor,
	auid=000,bioid=1,
	orcid=0009-0007-1337-1678]
	\ead{dongxiaorong0607@163.com}
	\credit{Investigation, Data Curation, Formal analysis}
	
	\author[2]{Xueyu Liu}[type=editor,
	auid=000,bioid=1,
	orcid=0000-0001-5745-4722]
	\ead{snowrain01@outlook.com}
	\credit{Formal analysis, Methodology, Validation}
	
	\author[3]{Jianan Zhang}[type=editor,
	auid=000,bioid=1,
	orcid=0009-0001-0147-0990]
	\ead{zhangjiananm0@sina.com}
	\credit{Supervision, Conceptualization, Methodology, Writing – Review \& Editing, Project administration}
	\cormark[1]
	
	\author[4]{Xuetao Ma}[type=editor,
	auid=000,bioid=1,
	orcid=0000-0003-1182-709X]
	\ead{maxuetao@mail.bnu.edu.cn}
	\credit{Resources, Investigation}
	
	\author[2]{Mingqiang Wei}[type=editor,
	auid=000,bioid=1,
	orcid=0000-0002-5572-1126]
	\ead{weimingqiang@tyut.edu.cn}
	\credit{Conceptualization, Methodology, Supervision}

	\author[5]{Liansheng Wang}[type=editor,
	auid=000,bioid=1,
	orcid=0000-0002-2096-454X]
	\ead{lswang@xmu.edu.cn}
	\credit{Conceptualization, Writing – Review \& Editing}
	
	\author[6]{Junxin Chen}[type=editor,
	auid=000,bioid=1,
	orcid=0009-0007-1337-1678]
	\ead{chenjunxin@dlut.edu.cn}
	\credit{Conceptualization, Writing – Review \& Editing}

	\author[2]{Yongfei Wu}[type=editor,
	auid=000,bioid=1,
	orcid=0000-0002-5010-2561]
	\ead{wuyongfei@tyut.edu.cn}
	\credit{Supervision, Project administration, Writing – Review \& Editing, Funding acquisition}
	\cormark[1]

	\address[1]{College of Computer Science and Technology (College of Data Science), Taiyuan University of Technology, Taiyuan, Shanxi, 030024, China}
	\address[2]{College of Artificial Intelligence, Taiyuan University of Technology, Taiyuan, Shanxi, 030024, China}
	\address[3]{School of mathematics, Taiyuan University of Technology, Taiyuan, 030024, China.}
	\address[4]{School of Artificial Intelligence, Beijing Normal University, Beijing, 100875, China}
	\address[5]{National Institute for Data Science in Health and Medicine, and the Department of Computer Science, School of Informatics, Xiammen, 361005, China}
	\address[6]{School of Software, Dalian University of Technology, Dalian,116621, China}

	\tnotetext[1]{This work was supported in part by the National Natural Science Foundation of China under Grant 61901292, in part by the Natural Science Foundation of Shanxi Province under Grant 202303021211082, and in part by the Graduate Scientific Research and Innovation Project of Shanxi Province under Grant RC2400005593.}
	
	\cortext[1]{Corresponding authors: Jianan Zhang and Yongfei Wu.}
	
	\begin{abstract}
		Accurate lesion segmentation in histopathology images is essential for diagnostic interpretation and quantitative analysis, yet it remains challenging due to the limited availability of costly pixel-level annotations. To address this, we propose FMaMIL, a novel two-stage framework for weakly supervised lesion segmentation based solely on image-level labels. In the first stage, a lightweight Mamba-based encoder is introduced to capture long-range dependencies across image patches under the MIL paradigm. To enhance spatial sensitivity and structural awareness, we design a learnable frequency-domain encoding module that supplements spatial-domain features with spectrum-based information. CAMs generated in this stage are used to guide segmentation training. In the second stage, we refine the initial pseudo labels via a CAM-guided soft-label supervision and a self-correction mechanism, enabling robust training even under label noise. Extensive experiments on both public and private histopathology datasets demonstrate that FMaMIL outperforms state-of-the-art weakly supervised methods without relying on pixel-level annotations, validating its effectiveness and potential for digital pathology applications.  The code can be accessed at \url{https://github.com/chenghangbei0702/FmaMIL}.
	\end{abstract}

	\begin{highlights}
		\item We propose FMaMIL, the first Mamba-based MIL framework enhanced with learnable frequency-domain encoding for weakly supervised lesion segmentation using only image-level labels.
		\item A bidirectional spatial scanning strategy is introduced to model contextual structures in pathology images more effectively.
		\item A CAM-guided pseudo-label refinement strategy with soft supervision and self-correction enhances segmentation under label noise.
		\item Our method achieves state-of-the-art segmentation performance on both public and private histopathology datasets without relying on pixel-level annotations.
	\end{highlights}
	
	
	\maketitle
	
	\section{Introduction}
	\label{Introduction}
	
	\begin{figure}[!t]
		\includegraphics[width=\linewidth]{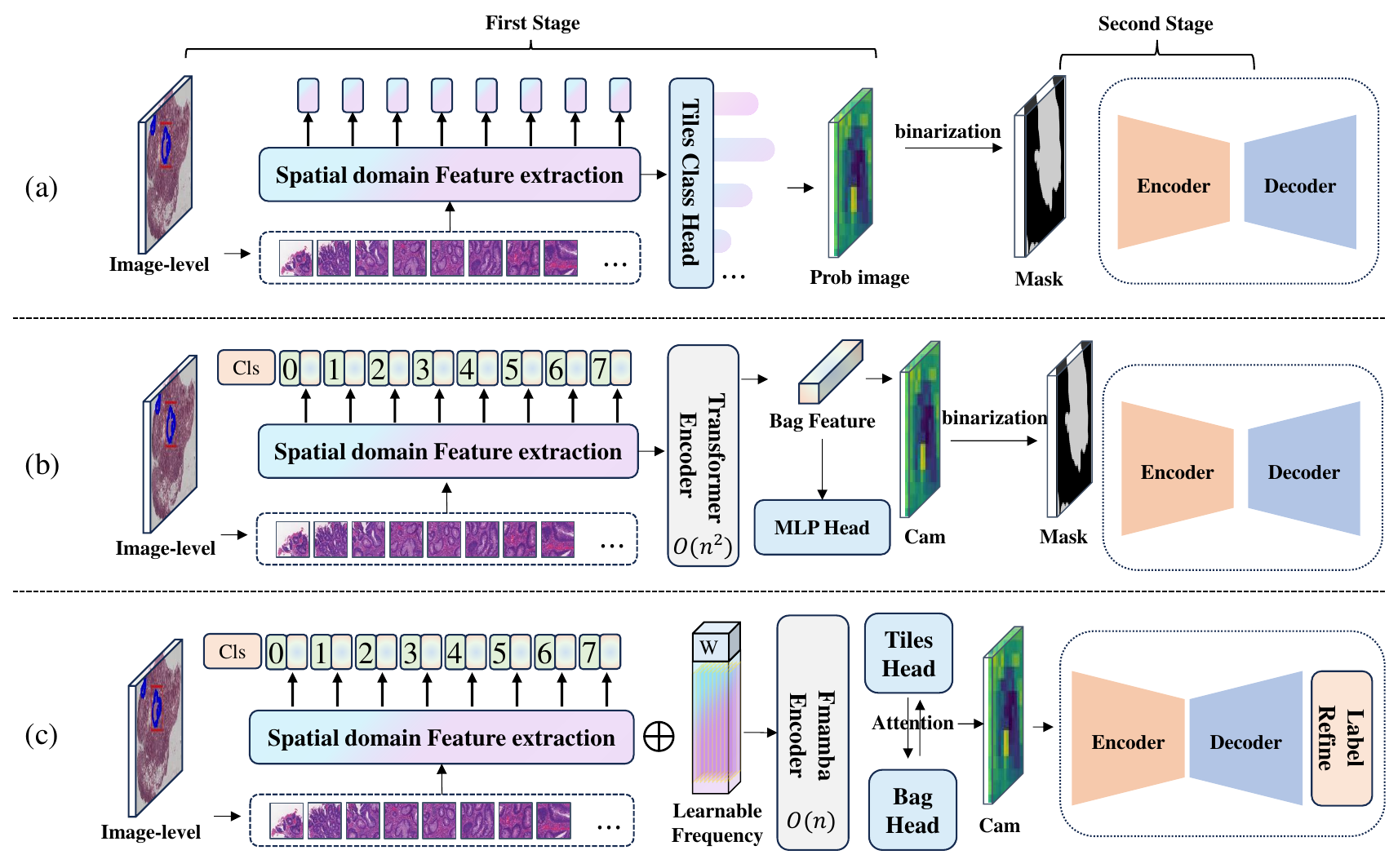}
		\centering
		\caption{Comparison of three two-stage WSS frameworks for pathology images.
			(a) Tile classification-based method for generating pseudo masks;
			(b) Transformer-based MIL with attention map generation;
			(c) Our proposed method, which incorporates learnable frequency features and a lightweight Mamba encoder to achieve efficient and accurate mask generation with label refinement..}
		\label{frameworkcompare}
	\end{figure}
	
\begin{figure}[!t]
		\includegraphics[width=\linewidth]{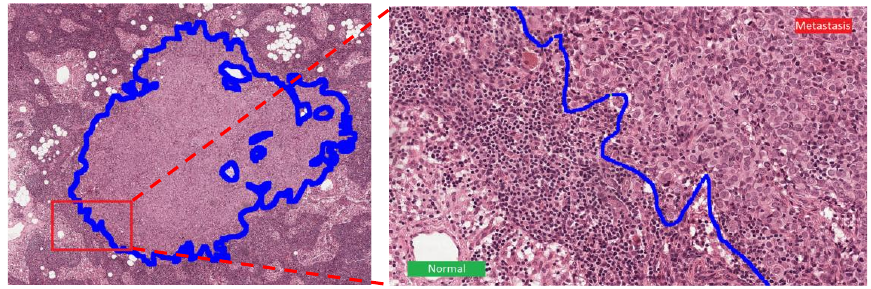}
		\centering
		\caption{illustrates the common diffuse characteristics of lesions.}
		\label{mil}
	\end{figure}

	In the field of digital medicine, ultra-high-resolution whole-slide images (WSIs) have become a mainstream imaging modality for modern disease diagnosis due to their comprehensive and rich diagnostic information\cite{wang2019pathology,xu2017large,hou2016patch,lin2018scannet,dong2018reinforced}. However, accurately identifying abnormal tissue regions in such high-resolution and structurally complex pathological images poses a significant challenge. These abnormal regions often exhibit subtle and diffuse structural features, as illustrated in Fig.~\ref{mil}, creating substantial difficulties for automated analysis. Traditional supervised segmentation methods rely heavily on high-quality pixel-level annotations provided by experienced medical experts, which are both time-consuming and expensive\cite{wang2021annotation}. Moreover, the subjectivity involved in annotation can lead to substantial variability in labeling results for the same lesion, further undermining the generalization capability of models\cite{liao2024modeling}. Consequently, patient-level labels, as the most readily available form of coarse annotation in clinical practice, have gained attention for weakly supervised lesion segmentation, offering an effective strategy to alleviate the reliance on high-quality data annotations\cite{wang2020ud,bakalo2021weakly}.
	
	Segmentation methods based on patient-level labels provide only pathological category information without specifying lesion locations, making this approach one of the most challenging problems in pixel-level segmentation tasks\cite{litjens2017survey,liu2021review}. Due to GPU memory constraints, traditional weakly supervised segmentation(WSS) methods cannot directly process WSIs comprising billions of pixels\cite{wang2019weakly}, and multiple instance learning (MIL) has emerged as the predominant solution\cite{quellec2017multiple}. MIL operates under a bag-based learning paradigm, where each bag comprises multiple instances, and labels are provided only at the bag level while individual instance labels remain unknown\cite{dietterich1997solving}. In pathological image analysis, a WSI is considered positive if it contains any lesion areas\cite{litjens2017survey,ronneberger2015u}. By leveraging this strategy, feature extraction is performed on small instances within the MIL framework rather than on the gigapixel WSI, alleviating GPU memory limitations\cite{quellec2017multiple,shao2021transmil}.
	
	Conventional MIL tasks typically encode instances within a bag into low-dimensional features using pre-trained models and aggregate these features into bag-level representations~\cite{ilse2018attention,cheplygina2015multiple}. In this process, instances within the bag are often assumed to be independent and identically distributed (i.i.d.), resulting in loss functions that backpropagate only to a subset of prominent positive instances, potentially leading to false segmentation issues~\cite{ilse2018attention,wang2018revisiting}, as exemplified in tile classification-based frameworks(Fig.~\ref{frameworkcompare}(a)). In contrast, clinical pathologists consider not only contextual information around regions but also the interrelations between different regions during diagnosis\cite{litjens2017survey,quellec2017multiple}. Consequently, recent studies have reformulated WSI analysis as a long-sequence modeling problem\cite{yang2024mambamil,shao2021transmil}, employing models like Transformers (Fig.~\ref{frameworkcompare}(b)) to capture discriminative features by exploring inter-instance correlations and global contextual information at the bag level. Despite improvements in performance, such methods face challenges of high computational cost and overfitting, limiting further advancements\cite{zhang2025attention}. Recently, the selective state-space model (Mamba) was introduced as a novel approach to address the computational complexity bottleneck of Transformers\cite{qu2024survey,hu2024state,waleffe2024empirical}. Mamba achieves linear computational complexity without compromising the global receptive field, significantly enhancing efficiency. Studies have demonstrated Mamba's superior ability to capture long-range dependencies, surpassing traditional models. Unlike typical visual tasks, WSIs exhibit weak spatial correlations and sparse distributions of positive patches, making Mamba’s robust sequential modeling capabilities particularly advantageous for WSI analysis. Interestingly, although some methods have combined Mamba with MIL, these efforts have primarily focused on image classification, with limited attention to the more challenging segmentation tasks\cite{yang2024mambamil,fang2024mammil}. Furthermore, existing research predominantly extracts features in the spatial domain, neglecting the potential value of frequency-domain information\cite{hao2024t}. Frequency-domain features uniquely excel in capturing texture and edge details in pathological images, especially in noisy or complex backgrounds\cite{cao2024multi,xu2020learning}. Therefore, integrating multi-level features from both spatial and frequency domains holds promise for further performance improvements.
	
	Based on this background, we propose FMaMIL, a novel Mamba MIL framework that integrates frequency-domain and spatial-domain information for weakly supervised lesion segmentation in digital pathology, as illustrated in Fig.~\ref{frameworkcompare}(c). The proposed framework leverages Mamba’s linear computational advantages in long-range feature extraction to thoroughly explore inter-instance correlations, while frequency-domain information enhances feature representation to better understand the complex characteristics of lesions. FMaMIL demonstrates excellent scalability, effectively mining instance-level features while ensuring bag-level classification accuracy. We evaluate the proposed method on two  datasets, including one private and one public datasets. Experimental results show that our model significantly outperforms state-of-the-art methods, achieving superior performance and strong generalization in lesion segmentation tasks.
	
	The major contributions of this paper are summarized as follows:
	\begin{itemize} 
		\item To the best of our knowledge, we are the first to propose FMaMIL—a novel weakly supervised lesion segmentation framework that integrates Mamba-based long-range modeling and learnable frequency-domain encoding under the MIL paradigm, using only image-level labels.
		
		\item A learnable frequency-domain encoding module and a bidirectional spatial scanning mechanism are introduced to enhance the model’s ability to capture texture details and contextual structure.
		
		\item We design a CAM-guided pseudo-label refinement strategy with soft supervision and self-correction, enabling coarse-to-fine training and achieving state-of-the-art performance on two histopathology datasets.
		
	\end{itemize}
	
	The remainder of this paper is organized as follows: 
	Section \ref{Related works}  reviews the related work, highlighting the limitations and challenges in the field.
	Section \ref{Proposedmethod} focuses on the proposed method,  detailing its design and key technical aspects.
	Section \ref{Experiment} presents experimental results on three datasets and analyzes ablation studies.
	Section \ref{discussion} discusses the strengths and limitations of the model
	Finally, Section \ref{Conclusion} concludes the paper and outlines future research directions.
	
	\section{Related works}
	\label{Related works}
	\subsection{Weakly supervised medical image seg.}
	
	Weakly supervised medical image segmentation primarily relies on points \cite{bearman2016s}, bounding boxes \cite{dai2015boxsup}, and image-level annotations \cite{xu2014weakly,wang2017learning}. Image-level annotations, being the easiest form of labels to obtain clinically in medicine, have become the mainstream approach. Existing WSSS methods based on image-level labels can mainly be divided into single-stage methods \cite{araslanov2020single} and two-stage methods \cite{wu2024dupl}. Single-stage methods directly use image-level labels as supervision to train end-to-end segmentation networks, while two-stage methods typically involve two separate training steps: converting image-level labels into offline class activation maps (CAM) \cite{zhou2016learning}, which highlight the regions in the image that are most influential for classification decisions through heatmap generation, thereby transforming image-level labels into pixel-level pseudo-labels. These pseudo-labels are then used to train segmentation models to improve segmentation accuracy. However, pseudo-labels often contain noise \cite{xiao2015learning}, which directly affects the output quality of segmentation models. To improve the quality of attention maps, several improved CAM methods have been proposed, including adversarial erasing \cite{wei2017object}, online attention accumulation \cite{jiang2019integral}, and region growing. Although these methods have made improvements, they usually take the entire image as the only input, are prone to background noise interference, and mainly extract global image information. Especially in digital pathology images, the lesion areas are similar to the background \cite{campanella2019clinical}, and digital pathology images have the characteristics of high resolution and complex structure \cite{xu2024whole}, making traditional deep learning methods difficult to effectively deal with. MIL \cite{maron1997framework} has emerged as an effective solution, which processes bags containing multiple instances and relies only on bag-level labels, effectively solving the problem of pathology image segmentation \cite{ilse2018attention,lerousseau2020weakly}.
	
	\subsection{Application of MIL in Pathological Seg.}

	MIL utilizes image-level annotations as labels and segments pathological images into multiple patches. These patches do not have independent labels but are associated only with image-level annotations. MIL effectively addresses the challenges posed by the high resolution and sparse annotations in WSIs\cite{campanella2019clinical}, achieving a balance between efficiency and effectiveness \cite{li2023weakly}. Traditional MIL frameworks can be divided into instance-level algorithms and embedding-level algorithms. Kanavati et al. \cite{kanavati2020weakly} trained a CNN based on the EfficientNet-B3 architecture using transfer learning and MIL, with a focus on the most representative instances. Ilse et al. \cite{ilse2018attention} extracted instance features using a CNN and aggregated these features via an attention mechanism to identify key instances. However, traditional MIL methods typically rely on the assumption of iid instances, extracting features from individual instances only. In clinical practice, pathologists not only rely on features within regions but also pay attention to the relationships between different regions. The aforementioned methods all neglect the correlations between instances \cite{yang2024mambamil}.
	
	To address this issue, several solutions have been explored. Zhuchen Shao et al. proposed a novel model, TransMIL, based on a correlated multi-instance learning framework \cite{shao2021transmil}. This model integrates morphological and spatial information through manually designed transformation functions and models the correlations between instances using the self-attention mechanism of Transformers, thereby capturing comprehensive image information. Z. Chen et al. introduced the SA-MIL model \cite{li2023weakly}, which incorporates self-attention and deep supervision mechanisms into the standard MIL framework to capture contextual information from the entire image, overcoming the limitations of independent instances. Fang et al. proposed the SAM-MIL framework\cite{fang2024sam}, which addresses the issue of ignoring spatial relationships between instances in traditional MIL methods by introducing spatial information and context-aware mechanisms. However, these methods consume substantial memory resources and have high computational complexity when dealing with higher-dimensional data and complex pathological image structures. Enhancing the model's ability to process long sequences and improving memory efficiency in WSS of pathological images have become critical issues that urgently need to be resolved in the scientific community.
	
	\subsection{Applications of Mamba in Medical Imaging}
	
	To achieve high-precision segmentation, multi-instance segmentation is often combined with computationally intensive models such as Transformers, which pose significant challenges when processing long-sequence inputs, such as medical images\cite{vaswani2017attention,khan2022transformers,han2022survey}. To improve efficiency, State-Space Models (SSM)\cite{li2025videomamba,gu2020hippo} have emerged. This model maps sequential data to a lower-dimensional state space and utilizes matrix or convolution operations for modeling, thereby achieving linear complexity. The Mamba model\cite{wang2024state,gu2023mamba}, based on SSM, incorporates global receptive fields and selective scanning operators, expanding the range of information capture while reducing computational load and capturing critical features. These advantages have led to renewed interest in SSM within the academic community.
	
	As Mamba technology has evolved, it has increasingly integrated with other techniques. In visual tasks, Vision Mamba \cite{zhu2024vision} enhances the efficiency of visual feature extraction by processing long-sequence medical images using selective scanning mechanisms, position-awareness, and bidirectional SSM. VMamba\cite{DBLP:journals/corr/abs-2401-10166} effectively collects contextual information from different perspectives through four scanning paths. In MIL tasks, the MamMIL model\cite{fang2024mammil} applies SSM for the first time within the WSI MIL framework, achieving WSI classification with linear complexity. The MambaMIL model \cite{yang2024mambamil}enhances long-sequence modeling through a sequence reordering module to improve WSI analysis performance. Mamba2MIL\cite{zhang2024mamba2mil} further refines the MambaMIL model by introducing multi-scale analysis to capture additional details. These studies demonstrate that the Mamba framework significantly enhances medical image segmentation accuracy while improving computational efficiency, signaling its broad application potential in medical image analysis. However, it is worth noting that most current research focuses on spatial domain information, whereas frequency domain information is equally critical in medical image analysis\cite{liu2021feddg,feng2022fiba}.
	
	\subsection{Frequency-Domain Information in WSIs Seg.}
	
	In recent years, frequency-domain information has gained increasing recognition for its ability to effectively integrate global and local feature modeling, highlighting its importance in pathological image segmentation \cite{ding2023slf,ruan2025learning,chang2024net,zheng2024fouriermil,hao2024t,zhou2024spatial}. For instance, FourierMIL \cite{zheng2024fouriermil} leverages Discrete Fourier Transform (DFT) and All-Pass Frequency Filtering to effectively capture both global and local dependencies in the frequency domain, significantly improving classification and segmentation performance for pathological images, particularly in high-resolution WSI analysis. Similarly, T-Mamba \cite{hao2024t} integrates frequency-domain features with spatial-domain features by employing a frequency enhancement module and an adaptive gating unit, achieving substantial improvements in the segmentation accuracy of dental CBCT images. This approach demonstrates strong advantages, especially when handling noisy and low-contrast pathological images. Furthermore, Zhou et al. proposed the Spatial-Frequency Dual-Domain Attention Network \cite{zhou2024spatial}, which utilizes 2D Discrete Fourier Transform (2D-DFT) to separate low- and high-frequency components of images, combined with adaptive filtering to enhance global feature learning, thereby improving the segmentation performance of pathological images.
	
	Despite the significant potential of frequency-domain information in pathological image segmentation, its application still faces several challenges. For example, most filter thresholds require manual tuning, which limits adaptability, and integrating frequency-domain information with spatial-domain features remains difficult. Future research should focus on optimizing adaptive filtering strategies and developing efficient fusion mechanisms to further promote the application of frequency-domain techniques in pathological image segmentation.

	\section{Proposed method}
	\label{Proposedmethod}
	%在本节中，我们将首先介绍相关的先验知识，之后详细介绍我们提出的整体框架，包括各个核心组件的技术细节，涉及到mamba块的设计、可学习的频域编码、多实例分类头等。
	In this section, we first introduce the relevant prior knowledge, followed by a detailed explanation of the proposed framework. Specifically, we analyze the key components of the framework, including the design of the FMamba block, the learnable frequency-domain encoding, the multi-instance classification head and the Cam-seg Model, along with their technical details.

	\subsection{Preliminaries}
	\subsubsection{Mamba}
	Recently, SSM-based methods have demonstrated significant potential in the field of sequential data processing, with particular attention to structured state-space sequence models (S4) and selective state-space models (Mamba). These models rely on classical continuous system theory and leverage efficient algorithms and architectures to process one-dimensional functions or sequences $x(t) \in \mathbb{R}$ through hidden states $h(t) \in \mathbb{R}^N$ and map them to outputs $y(t) \in \mathbb{R}$. This mapping process is achieved through carefully designed evolution parameters $\mathbf{A} \in \mathbb{R}^{N \times N}$ and projection parameters $\mathbf{B} \in \mathbb{R}^{N \times 1}$ and $\mathbf{C} \in \mathbb{R}^{1 \times N}$
	
	In the continuous-time domain, this process can be represented as a linear ordinary differential equation:
	\begin{equation}
		\begin{aligned}
			h'(t) &= \mathbf{A}h(t) + \mathbf{B}x(t), \\
			y(t) &= \mathbf{C}h(t).
		\end{aligned}
	\end{equation}
	where $\mathbf{A}$ serves as the evolution matrix for hidden states, $\mathbf{B}$ describes the influence of input signal $x(t)$ on hidden states, and $\mathbf{C}$ is responsible for extracting the output from hidden states, $h'(t)$ denotes the rate of change of the hidden state over time.
	
	To apply this continuous system to discrete-time scenarios, S4 and Mamba models introduce a time-scale parameter $\Delta$ and use discretization methods such as Zero-Order Hold (ZOH) to convert continuous parameters $\mathbf{A}$ and $\mathbf{B}$ into discrete parameters $\overline{\mathbf{A}}$ and $\overline{\mathbf{B}}$. This process typically follows the following formulas:
	\begin{equation}
		\begin{split}
			\overline{\mathbf{A}} &= \exp(\Delta \mathbf{A}), \\
			\overline{\mathbf{B}} &= (\Delta\mathbf{A})^{-1}  (\exp(\Delta \mathbf{A}) - \mathbf{I} ) \cdot \Delta\mathbf{B}.
		\end{split}
	\end{equation}
	$\exp(\cdot)$ denotes the matrix exponential, serving as a bridge between continuous and discrete-time systems.
	
	In the discrete-time domain, the state equation and output equation of the state-space model can be expressed as:
	\begin{equation}
		\begin{split}
			h_t &= \overline{\mathbf{A}}h_{t-1} + \overline{\mathbf{B}}x_t, \\
			y_t &= \mathbf{C}h_t.
		\end{split}
	\end{equation}
	At the end of the process, by computing the convolution kernel $\overline{\mathbf{K}}\in \mathbb{R}^{M}$ of the SSM, the model can efficiently extract features from the input sequence $\text{x}$ and generate the output sequence $\text{y}$. This process can be represented as:
	\begin{equation}
		\begin{split}
			\overline{\mathbf{K}} &= (\mathbf{C}\overline{\mathbf{B}}, \mathbf{C}\overline{\mathbf{A}}\overline{\mathbf{B}}, \ldots, \mathbf{C}{\overline{\mathbf{A}}}^{M-1}\overline{\mathbf{B}}), \\
			\text{y} &= \text{x} * \overline{\mathbf{K}}.
		\end{split}
	\end{equation}
	Here, $M$ denotes the length of the input sequence $x$.
	
	\begin{figure*}[!t]
		\includegraphics[width=\linewidth]{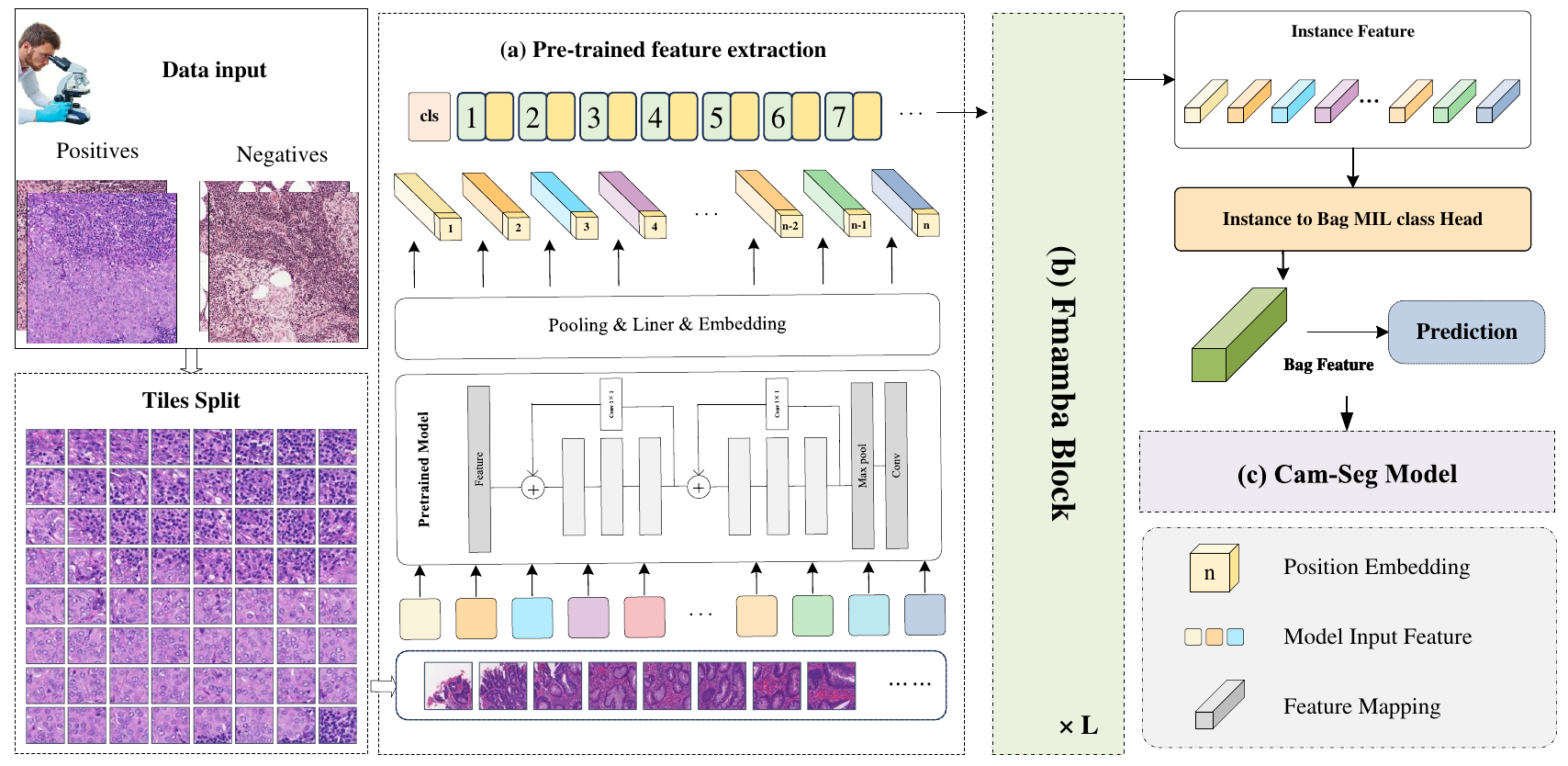}
		\centering
		\caption{The overall framework of our proposed FMaMIL primarily consists of three stages.}
		\label{framwork}
	\end{figure*}
	
	\subsubsection{Multiple Instance Learning}
	%在多示例的学习过程中，数据以包为单位，表示为
	In the process of MIL, the data is organized into bags rather than instances, The instances do not have labels. let $W = \{(x_1,y_1),...,(x_n,y_n)\}$ represent a WSI bag $W$, where $n$ represents the total number of patches contained in $W$. Each individual yield instance-level features $x_i \in \{x_{1},x_{2},\ldots,x_{n}\}$  has a corresponding latent label $y_i \in  \{y_{1},y_{2},\ldots,y_{N}\}$  (where $y_i$ = 1 denotes positive, and $y_i$ = 0 denotes negative). The labels $y$ for bag $W$ are known to the model and are represented as follows:
	
	\begin{equation}
		\label{mileq}
		y(W)=\begin{cases}0,&\text{if}  \sum_{i=1}^K y_i=0,\\1,&\text{otherwise}.\end{cases}
	\end{equation}
	When at least one instance in the bag  $W$ is classified as positive, the WSI is classified as positive; otherwise, it is classified as negative. MIL decomposes high-resolution images into multiple small image patches, which has been shown to extract more microstructural information about the lesions.
	
	\subsection{The overall framework}
	%我们创新性地提出了一种融合多示例学习与选择性状态空间方程的弱监督分割策略，该策略巧妙地结合了频域与空域的特征信息，我们称之为FMaMIL。如图x所示，我们所提出的方法框架精心设计了三个核心阶段：预训练特征提取、示例相关性深度挖掘、以及示例级至包级的分类头。在示例相关性的深度挖掘阶段，我们进一步引入了可学习的频域编码技术和空域特征的双向扫描机制。在接下来的小节中，我们将逐一详细阐述每个阶段的具体实施细节。
	We propose an innovative WSS strategy that integrates MIL with the Mamba, cleverly combining frequency-domain and spatial-domain feature information. We refer to this approach as FMaMIL. As shown in Fig.~\ref{framwork}, the proposed method framework is carefully designed with three core stages: Pre-trained Feature Extraction, deep exploration of instance correlations, and a pseudo-label training strategy based on the CAM.
	
	In the deep exploration of instance correlations stage, we further introduce a learnable frequency-domain encoding technique and a bidirectional scanning mechanism for spatial-domain features. In the following subsections, we will detail the specific implementation of each stage.
	
	\subsection{pre-trained feature extraction}
	Our framework is designed to process pathological slice data containing both positive and negative samples, including positive sample set $D^+ = \{\mathbf{S}_1^+, \mathbf{S}_2^+, \ldots, \mathbf{S}_{a}^+\}$ and the negative sample set  $D^- = \{\mathbf{S}_1^-, \mathbf{S}_2^-, \ldots, \mathbf{S}_{b}^-\}$. Each pathological slice $\mathbf{S}_a^+$ or $\mathbf{S}_b^- \in \mathbb{R}^{\mathtt{H \times W \times C}}$ is further divided into $N \times M$ non-overlapping image patches. $N$ for the number of rows, $M$ for the number of columns. $(H, W)$ represents the dimensions of the input image, and $C$ is the number of channels. Each patch $\mathbf{p}_{(n,m)} \in \{\mathbf{p}_{(1,1)}, \mathbf{p}_{(1,2)}, \ldots, \mathbf{p}_{(N,M)}\} \in \mathbb{R}^{\mathtt{q \times q \times C}}$ carries detailed information and tissue structure from a local region of the pathological slice. $q$ represents the size of the side length of a patch.
	
	Next, we use a pre-trained Convolutional Neural Network, such as ResNet, VGG, or EfficientNet, to extract features from each patch $\mathbf{p}_{(n,m)}$.  This process captures diverse features ranging from low-level details (e.g., edges and textures) to high-level semantic information (e.g., tissue types and lesion characteristics), encoding each patch into a fixed-dimensional vector \( \mathbf{f}_{(n,m)} \in \mathbb{R}^d \). The feature extraction process can be formulated as:
	\begin{equation}
		\mathbf{f}_{(n,m)} = F_{\text{CNN}}(\mathbf{p}_{(n,m)}),  n = 1, 2, \dots, N,  m = 1, 2, \dots, M,
	\end{equation}
	where $ F_{\text{CNN}} $ represents the pre-trained CNN model, and $ d $ is the dimension of the feature vector.
	
	After extracting features for all patches, These features are arranged into a sequential order by scanning the spatial positions of the patches row by row in the original image, forming a sequence:
	\begin{equation}
		\mathbf{F}_{row} = [\mathbf{f}_{(1,1)}, \mathbf{f}_{(1,2)}, \dots, \mathbf{f}_{(n,m)}] \in \mathbb{R}^{N  \times M \times d}.
	\end{equation}
	
	Next, positional encoding is added to each feature vector \( \mathbf{f}_{(n,m)} \) to retain spatial positional information of the image. Additionally, a learnable classification token (cls token), initialized as a zero vector \( \mathbf{f}_{\text{cls}} \in \mathbb{R}^d \), is prepended to the sequence. The final input sequence after adding the CLS token can be expressed as:
	\begin{equation}
		\begin{aligned}
			\mathbf{F}'_{row} &= [\mathbf{f}_{\text{cls}}, \mathbf{f}_{(1,1)} + \mathbf{e}_1, \mathbf{f}_{(1,2)} + \mathbf{e}_2, \dots, \mathbf{f}_{(n,m)} + \mathbf{e}_{N \times M}] \\
			&\in \mathbb{R}^{(N \times M +1) \times d}
		\end{aligned}
	\end{equation}
	where \( \mathbf{e}_i \in \mathbb{R}^d \) represents the positional encoding for the \( i \)-th patch.
	
	\subsection{deep exploration of instance correlations}
	%为了充分挖掘示例之间的相关性，并结合现实的视觉扫描模式，还原图像在二维空间中的结构信息，我们将在这一节首先介绍句子重排序技术、之后介绍可学习的频域编码设计，最后介绍整个基于mamba与双域信息的编码器结构。
	To fully explore the correlations between samples and reconstruct the structural information of images in two-dim space based on realistic visual scanning patterns, this section will first introduce the sentence reordering technique, followed by the design of a learnable frequency-domain encoding. Finally, the overall encoder architecture based on Mamba and dual-domain information will be presented.
	
	\subsubsection{Sentence reordering}
	\begin{figure}[!t]
		\centering
		\includegraphics[width=\linewidth]{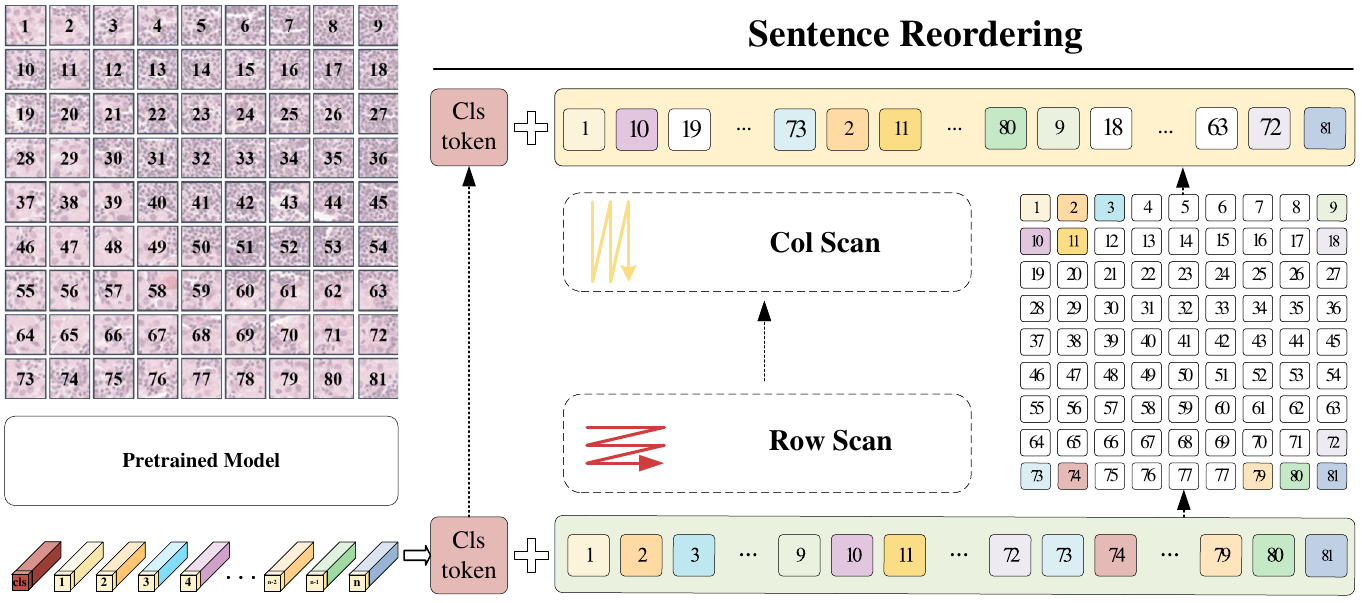}
		\caption{\small{Sentence Reordering Structure.}}
		\label{sequence}
	\end{figure}
	
	%在预训练阶段结束后，我们得到了每张病理切片的特征序列 ，该序列通过对二维空间逐行扫描得到。然而，由于模型在推理时可能会忽略来自上下空间的信息，为此我们设计了一种句子重排序技术，将逐行扫描的特征序列  转换为逐列扫描的特征序列 。具体过程如图 X 所示。具体而言，句子重排序技术将依据patch在原始病理切片中的列位置，而非行位置，来重新组织这些patch的序列。每个patch在新的逐列扫描序列中都会保持其原有的特征向量，但它们的排列顺序会根据列索引进行调整。同时，我们也会为这个新的逐列扫描序列添加相应的位置编码与一个额外的类标记，以便模型能够区分并处理两个不同的扫描模式，捕获图像的空间结构信息。
	After the pre-training phase, we obtain the feature sequence $\mathbf{F}'_{row}$ for each pathology slide, which is generated through row-wise scanning in two-dimensional space. However, during inference, the model may overlook information from the vertical spatial dimension. To address this, we designed a sentence reordering technique to transform the row-wise feature sequence $\mathbf{F}'_{row}$ into a column-wise feature sequence $\mathbf{F}'_{col}$. The detailed process is illustrated in Fig.~\ref{sequence}.
	
	Specifically, the sentence reordering technique reorganizes the patch feature sequence based on their column positions in the original pathology slide, rather than their row positions. In the newly generated column-wise sequence, each patch retains its original feature vector, but their arrangement is adjusted according to the column indices. The reordered feature sequence can be expressed as:
	\begin{equation}
		\begin{aligned}
			\mathbf{F}'_{col} &= [\mathbf{f}_{\text{cls}}, \mathbf{f}_{(1,1)} + \mathbf{e}_1, \mathbf{f}_{(2,1)} + \mathbf{e}_2, \dots, \mathbf{f}_{(m,n)} + \mathbf{e}_{M \times N}] \\
			&\in \mathbb{R}^{(N \times M +1) \times d}
		\end{aligned}
	\end{equation}
	
	where $\mathbf{f}_{(m,n)}$ represents the feature vector of the patch located at the $m$-th row and $n$-th column.
	
	In addition, we incorporate positional encodings and an extra learnable classification token (CLS token) into the new column-wise sequence $\mathbf{F}'_{col}$. This allows the model to differentiate and effectively process the two scanning modes (row-wise and column-wise), thereby capturing the spatial structural information of the pathology slides more comprehensively.

	\subsubsection{learnable frequency-domain encoding}
	\begin{figure}[!t]
		\centering
		\includegraphics[width=\linewidth]{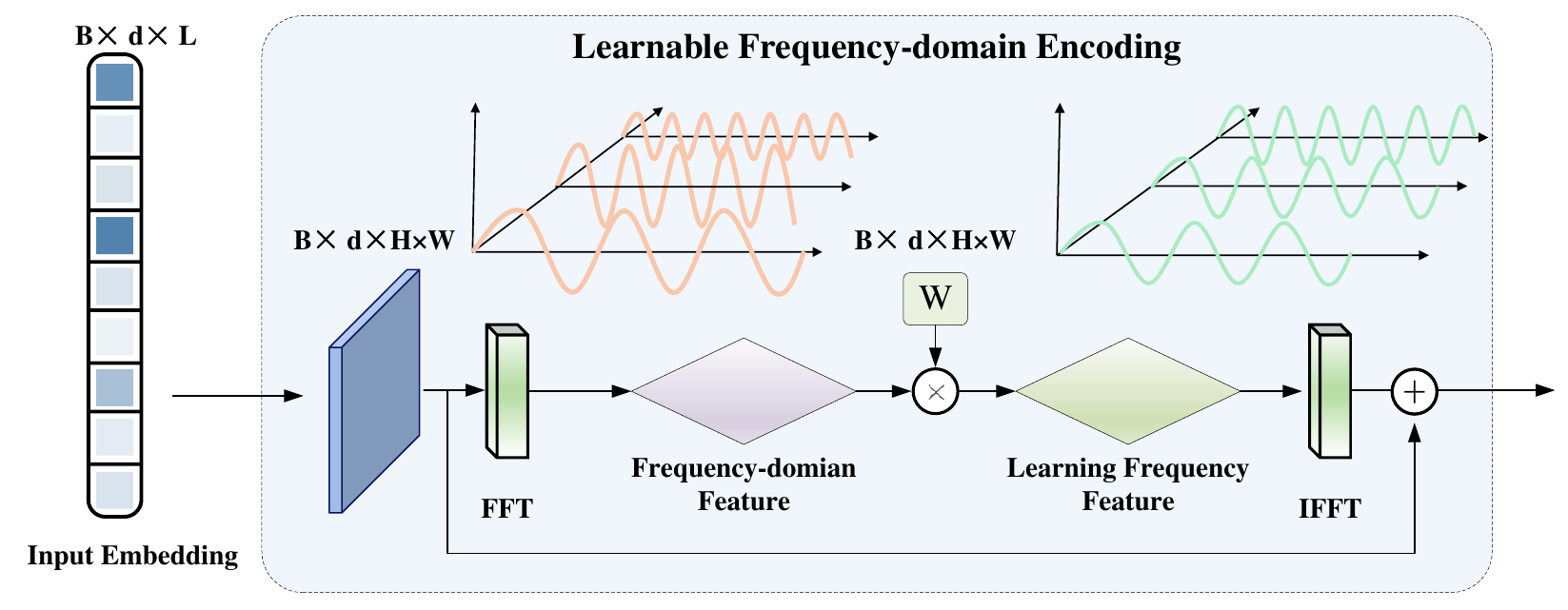}
		\caption{\small{Learnable Frequency Domain Encoding Structure.}}
		\label{frequency}
	\end{figure}
	
	Features extracted in the spatial domain typically capture local spatial details and dependencies. However, the frequency domain offers a different perspective, focusing on the frequency components and global patterns of the data. This is crucial for modeling long-range dependencies. High-resolution digital pathology images often contain substantial local details, such as cellular structures, interstitial spaces, and blood vessels. Unlike traditional classification methods, MIL relies more heavily on understanding local texture details (corresponding to high-frequency signals in the frequency domain) and global structural features (corresponding to low-frequency signals in the frequency domain). The fusion of these multi-scale features is particularly important for effective pseudo-segmentation tasks.
	
	Frequency domain encoding helps models capture hierarchical spatial information and organizational structures more effectively. Based on this idea, we propose the use of Fast Fourier Transform (FFT) to build learnable frequency domain encoding with logarithmic-linear computational complexity, achieving more comprehensive feature extraction. This approach can be viewed as a trainable frequency filter that automatically learns the relevant frequency components for classification while suppressing unrelated signals. This operation also helps reduce the computational burden on the network.
	
	\begin{algorithm}[H]
		\caption{Learnable Frequency Domain Encoding}
		\textbf{Require:} token sequence $\mathbf{x}_{f} : \textcolor{keywordcolor}{\mathtt{(B, M, E)}}$ \\
		\textbf{Ensure:} token sequence $\mathbf{y}_{f} : \textcolor{keywordcolor}{\mathtt{(B, M, E)}}$
		\begin{algorithmic}[1]
			\State $\textcolor{commentcolor}{\text{/* FFT branch for frequency domain processing */}}$
			\State $\mathbf{x'}_{f} : \textcolor{keywordcolor}{\mathtt{(B, M, E)}} \gets \mathbf{FFT}(\mathbf{x}_{f})$
			\State $\mathbf{x'}_{f} : \textcolor{keywordcolor}{\mathtt{(B, M, E)}} \gets  \mathbf{x'}_{f} \bm{\odot} \mathbf{W}_f$
			\State $\textcolor{commentcolor}{\text{/* dd learnable parameters */}}$
			\State $\mathbf{x'}_{f} : \textcolor{keywordcolor}{\mathtt{(B, M, E)}} \gets \mathbf{LearnableParameters}(\mathbf{x'}_{f})$
			\State $\textcolor{commentcolor}{\text{/* IFFT branch and follow-up actions */}}$
			\State $\mathbf{x'}_{f} : \textcolor{keywordcolor}{\mathtt{(B, M, E)}} \gets \mathbf{IFFT}(\mathbf{x'}_{f})$
			\State $\textcolor{commentcolor}{\text{/* add skip connection */}}$
			\State $\mathbf{y}_{f} : \textcolor{keywordcolor}{\mathtt{(B, M, E)}} \gets \mathbf{x'}_{f} + \mathbf{x}_{f}$
			\State $\mathbf{Return}: \mathbf{y}_{f}  $
		\end{algorithmic}
		\label{lfft}
	\end{algorithm}

	\textbf{Architecture:} The learnable frequency domain encoding approach is based on a spectral network composed of an FFT layer, a learnable encoding module, and an inverse FFT (IFFT) layer. The detailed process is illustrated in Fig.~\ref{frequency} and Algorithm.~\ref{lfft}.

	First, the input feature tensor $x \in \mathbb{R}^{B \times d \times L}$, where \( B \) is the batch size,and \( L \) is the sequence length. We apply FFT along the sequence dimension \( L \), converting the feature from the spatial domain to the frequency domain, yielding the complex frequency domain representation \( x_{\text{fft}} \), which consists of real part \( x_{\text{fft\_real}} \) and imaginary part \( x_{\text{fft\_imag}} \). This process can be represented as:
	
	\begin{equation}
		\begin{split}
			x_{\text{fft}} & = \text{FFT}(x),\\
			X[k] &=\sum_{n=0}^{L-1} x[n] \cdot e^{-j \frac{2 \pi}{L} k n}, \quad k=0,1, \ldots, L-1, \\
			x_{\text{fft}} &= x_{\text{fft\_real}} + j \cdot x_{\text{fft\_imag}}
		\end{split}
	\end{equation}
	where $x[n]$ is the $n$-th element of the input sequence, $X[k]$ is the $k$-th frequency component in the frequency domain,
	$j$ is the imaginary unit.
	
	Secondly,unlike traditional filtering methods, we use learnable weight parameters to automatically learn the importance of each frequency component, thus capturing both coarse and fine spatial features. This approach enhances the model's flexibility and robustness. Specifically, we introduce learnable frequency weights \( w \in \mathbb{R}^{D \times L}\) for each channel. The complex weights include both real and imaginary parts. These weights are applied to the frequency domain features using complex arithmetic rules, as follows:
	\begin{equation}
		x_{\text{fft\_out}} = x_{\text{fft}} \cdot w
	\end{equation}
	
	Finally, we apply the IFFT to transform the frequency domain features back to the spatial domain $x_{\mathrm{fft\_space}}$. To address the potential loss of local details—particularly the incomplete representation of edges and small objects—caused by the dispersion of information across the entire spectrum during the Fourier Transform (FT), we introduce a skip connection into the operation process. This process can be represented as:
	
	\begin{equation}
		\begin{split}
			x_{\text{fft\_space}} & = x + x_{\text{ifft}} \\
			& = x + \frac{1}{N}\sum_{k=0}^{N-1}X_{\text{fft\_out}}[k]e^{j2\pi\frac{kn}{N}}
		\end{split}
	\end{equation}

	\subsubsection{FMamba block}
	
	\begin{figure}[!t]
		\centering
		\includegraphics[width=\linewidth]{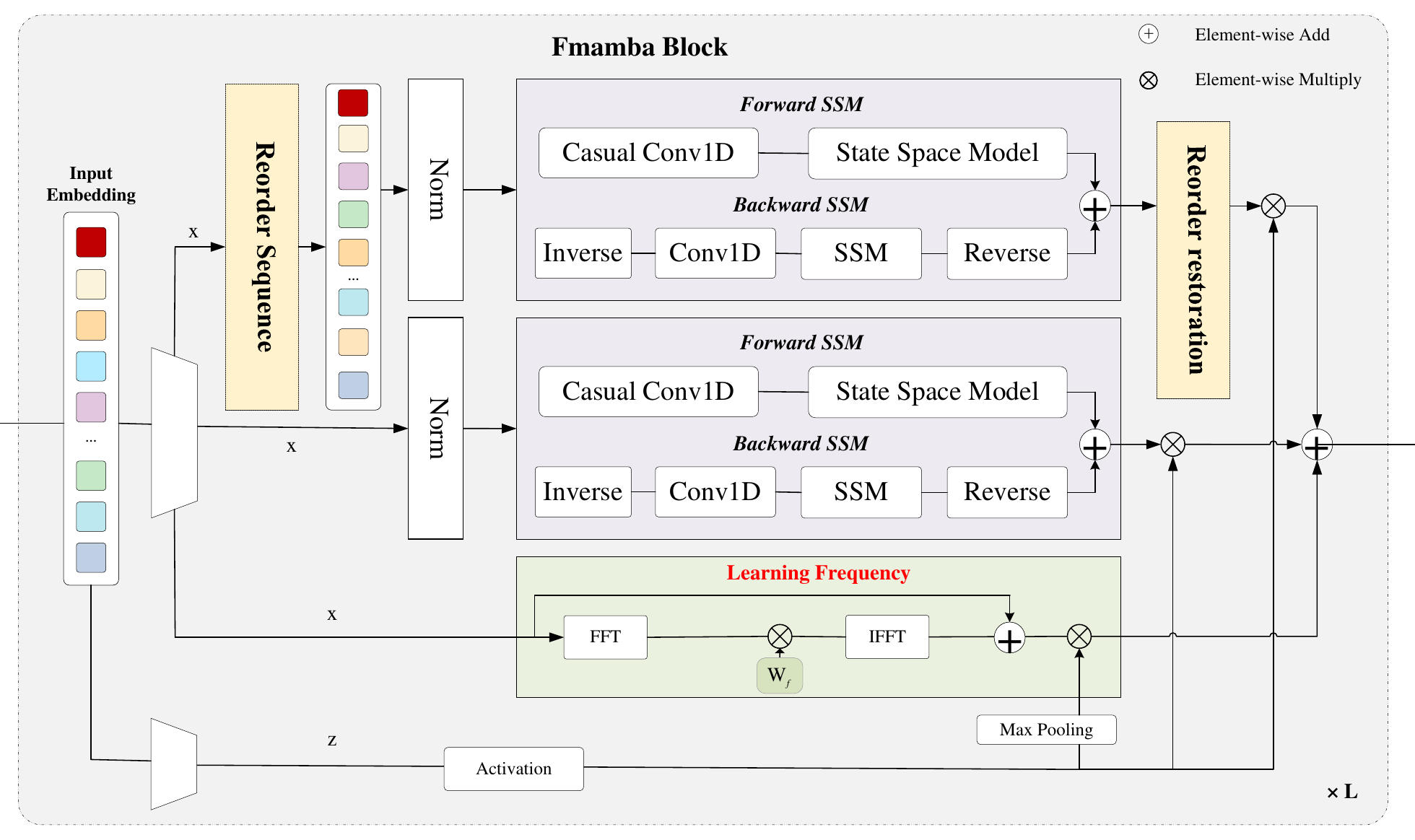}
		\caption{\small{Fmamba Block Structure.}}
		\label{fmamba}
	\end{figure}
	
	We have previously outlined the fundamental components of the mamba-based encoder within the FaMIL framework. Building upon this foundation, we further present the basic architecture of an enhanced mamba encoding block, which integrates spatial and frequency domain information, as illustrated in Fig.~\ref{fmamba} and Algorithm.~\ref{fmambaalg} . In this structure, we extract row-wise scanning sequences ($\mathbf{F}'_{row}$), column-wise scanning sequences ($\mathbf{F}'_{col}$), and the output of a learnable frequency domain encoding module ($x_{\text{fft\_space}}$) from the input.

	In the design of the novel mamba-based module, we draw inspiration from Ref. \cite{yue2024medmamba} and introduce a bidirectional sequence modeling strategy to reconstruct finer details in the visual 2D space. Specifically, for the sequences $\mathbf{F}'_{row}$ and $\mathbf{F}'_{col}$, we first standardize them using normalization layers. Subsequently, the normalized sequences are linearly projected into vectors $x$ and $z$ of dimension $E$. Next, we apply forward and backward SSM operations on $x$. Each directional sequence modeling involves passing through a Casual Convolution Layer followed by an SSM. For the output of sequence $\mathbf{F}'_{col}$, we go through the recorder restoration module before converting it into a sequence that corresponds to the progressive scan $\mathbf{F}'_{row}$ one by one. Finally, we aggregate the results of the bidirectional sequence modeling using a gating mechanism with $z$, yielding the corresponding outputs $\mathrm{out_\text{row}}$ and $\mathrm{out_\text{col}}$ for  $\mathbf{F}'_{row}$ and $\mathbf{F}'_{col}$, respectively. This process can be mathematically described as follows:

	\begin{equation}
		\begin{split}
			\mathbf{F}'' &= \mathrm{Norm}(\mathbf{F}'),\\
			\mathrm{out_\text{row/col}}&=\mathrm{SiLu}(z) \cdot(\mathrm{SSM}(\mathrm{Conv}1\mathrm{D}(\mathbf{F}'')) \\
			& \quad +\mathrm{SSM}(\mathrm{Conv}1\mathrm{D}(\mathbf{F}''[-1]))[-1])
		\end{split}
	\end{equation}
	where $\mathbf{F}'$ represents the input sequence ( $\mathbf{F}'_{row}$ or $\mathbf{F}'_{col}$), and $\mathrm{Conv}1\mathrm{D}$ denotes a one-dimensional convolutional layer.

	As a supplement, for $x_{\text{fft\_space}}$, the frequency feature is aggregated by the activated $z'$ with maxpooling operation. This approach preserves the global information adjusted in the frequency domain while maintaining the local patterns of spatial features. The process can be expressed mathematically as follows:
	\begin{equation}
		\mathrm{out}_\text{fft}=\mathrm{LayerNorm}(|x_{\text{fft\_space}} \cdot \mathrm{Max\_pool}(\mathrm{SiLu}(z))|)
	\end{equation}

	Ultimately, we obtain three key outputs:  $\mathrm{out_\text{row}}$ , $\mathrm{out_\text{col}}$ and $\mathrm{out}_\text{fft}$. Unlike the gated weighted fusion operation proposed in Ref. \cite{hao2024t}, we adopt a more straightforward and concise strategy, directly summing these three outputs to produce a new fused sequence $F_{\mathrm{new}}$. This approach not only preserves spatial information and frequency-domain features more comprehensively but also effectively reduces the computational complexity of the model. The process can be expressed mathematically as:

	\begin{equation}
		F_{\mathrm{new}}=\mathrm{out_\text{col}}+\mathrm{out_\text{row}}+\mathrm{out_\text{fft}}
	\end{equation}

	\begin{algorithm}[H]
		\caption{Fmamba Block Process}
		\textbf{Require:} token sequence $\mathbf{T}_{l-1} : \textcolor{keywordcolor}{\mathtt{(B, M, D)}}$ \\
		\textbf{Ensure:} token sequence $\mathbf{T}_{l} :  \textcolor{keywordcolor}{\mathtt{(B, M, D)}}$
		\begin{algorithmic}[1]
			\State $\textcolor{commentcolor}{\text{/* normalize the input sequence } \mathbf{T'}_{l-1} \text{ */}}$
			\State $\mathbf{T'}_{l-1} : \textcolor{keywordcolor}{\mathtt{(B, M, D)}} \gets \mathbf{Norm}(\mathbf{T}_{l-1})$
			\State $\mathbf{x} : \textcolor{keywordcolor}{\mathtt{(B, M, E)}} \gets \mathbf{Linear}^\mathbf{x}(\mathbf{T'}_{l-1})$
			\State $\mathbf{z} : \textcolor{keywordcolor}{\mathtt{(B, M, E)}} \gets \mathbf{Linear}^\mathbf{z}(\mathbf{T'}_{l-1})$
			
			\State $\textcolor{commentcolor}{\text{/* split x into three parts for different directions */}}$
			\State $\mathbf{x}_{h}, \mathbf{x}_{v}, \mathbf{x}_{f}: \textcolor{keywordcolor}{\mathtt{(B, M, E)}} \gets \mathbf{Split}(\mathbf{x})$
			\State $\textcolor{commentcolor}{\text{/* reorder $x_v$ for the vertical direction */}}$
			\State $\mathbf{x}_{v} : \textcolor{keywordcolor}{\mathtt{(B, M, E)}}\gets \mathbf{Reorder}(\mathbf{x}_{v})$
			\State $\textcolor{commentcolor}{\text{/* perform bidirectional propagation for $x_h$ and $x_v$ separately */}}$
			\For{$o \in \{\text{forward}, \text{backward}\}$}
			\State $\mathbf{x'}_{o}: \textcolor{keywordcolor}{\mathtt{(B, M, E)}} \gets \mathbf{SiLU}(\mathbf{Conv1d}_o(\mathbf{x}))$
			\State $\mathbf{B}_{o} : \textcolor{keywordcolor}{\mathtt{(B, M, N)}} \gets \mathbf{Linear}^\mathbf{B}_o(\mathbf{x'}_{o})$
			\State $\mathbf{C}_{o} : \textcolor{keywordcolor}{\mathtt{(B, M, N)}} \gets \mathbf{Linear}^\mathbf{C}_o(\mathbf{x'}_{o})$
			\State $\mathbf{\bm{\Delta}}_{o} : \textcolor{keywordcolor}{\mathtt{(B, M, E)}} \gets \log(1 + \exp(\mathbf{Linear}^{\mathbf{\bm{\Delta}
			}}_o(\mathbf{x'}_{o}) + \mathbf{Parameter}_o^\mathbf{\bm{\Delta}
			}))$
			\State $\overline{\mathbf{A}_{o}} : \textcolor{keywordcolor}{\mathtt{(B, M, E, N)}} \gets \mathbf{\bm{\Delta}
			}_{o} \boldsymbol{\otimes} \mathbf{Parameter}_o^\mathbf{A}$
			\State $\overline{\mathbf{B}_{o}} : \textcolor{keywordcolor}{\mathtt{(B, M, E, N)}} \gets \mathbf{\bm{\Delta}
			}_{o} \boldsymbol{\otimes}\mathbf{B}_{o}$
			\State $\mathbf{y}_{o} : \textcolor{keywordcolor}{\mathtt{(B, M, E)}} \gets \mathbf{SSM}(\overline{\mathbf{A}_{o}}, \overline{\mathbf{B}_{o}}, \mathbf{C}_{o})(\mathbf{x'}_{o})$
			\EndFor
			\State $\mathbf{y}_{h} : \textcolor{keywordcolor}{\mathtt{(B, M, E)}} \gets \mathbf{y}_{h\_forward} + \mathbf{y}_{h\_backward}$
			\State $\mathbf{y}_{v} : \textcolor{keywordcolor}{\mathtt{(B, M, E)}} \gets \mathbf{y}_{v\_forward} + \mathbf{y}_{v\_backward}$
			\State $\textcolor{commentcolor}{\text{/* perform reorder restoration on $\mathbf{y}_v$ */}}$
			\State $\mathbf{y}_{v} : \textcolor{keywordcolor}{\mathtt{(B, M, E)}}  \gets \mathbf{ReorderRestoration}  (\mathbf{y}_v) $
			\State $\textcolor{commentcolor}{\text{/* processing information in frequency domain */}}$
			\State $\mathbf{y}_{f} : \textcolor{keywordcolor}{\mathtt{(B, M, E)}}  \gets \mathbf{LFDE}  (\mathbf{x}_f) $
			\State $\mathbf{y'}_{h} : \textcolor{keywordcolor}{\mathtt{(B, M, E)}} \gets \mathbf{y}_{h} \bm{\odot} \mathbf{SiLU(z)}$
			\State $\mathbf{y'}_{v} : \textcolor{keywordcolor}{\mathtt{(B, M, E)}} \gets \mathbf{y}_{v} \bm{\odot} \mathbf{SiLU(z)}$
			\State $\mathbf{y'}_{f} : \textcolor{keywordcolor}{\mathtt{(B, M, E)}} \gets \mathbf{y}_{f} \bm{\odot} \mathbf{SiLU(Max Pooling(z))}$
			\State $\mathbf{T}_{l} : \textcolor{keywordcolor}{\mathtt{(B, M, D)}} \gets \mathbf{Linear}^{\mathbf{T}} (\mathbf{y'}_{h} + \mathbf{y'}_{v} + \mathbf{y'}_{f}) + \mathbf{T}_{l-1}$		
			\State $\mathbf{Return}: \mathbf{T}_{l}  $
		\end{algorithmic}
		\label{fmambaalg}
	\end{algorithm}
	
	\subsection{Instance-level to Bag-level Classification Head}  
	%经过若干个 Fmaba 块后，我们得到了形状为 $B \times L \times D$ 的特征张量，其中 $L$ 代表 patch 的数量加上 CLS token。为了专注于 patch 级的表示，我们去掉 CLS token，得到形状为 $B \times (L-1) \times D$ 的特征张量。这些 $L-1$ 个特征向量分别对应于每个单独的 patch。为了将这些实例级特征聚合成包级特征，我们首先使用共享的实例编码器将其映射到一个固定维度的潜在空间。该编码器由一系列线性变换和非线性激活函数（如 ReLU）组成，以增强特征的表达能力。在实例编码之后，我们引入注意力机制来计算每个 patch 在包中的重要性。具体而言，注意力模块由线性层和 $\tanh$ 激活函数组成，以学习 patch 级别的注意力分数。这些分数随后通过 softmax 归一化，确保它们在一个包内的总和为 1。最终的包级特征通过实例特征的加权求和获得，其中学习到的注意力分数作为权重。该聚合后的特征作为整个包的紧凑表示，并进一步通过分类头进行最终的标签预测。这种基于注意力的多实例学习（MIL）分类头能够使模型聚焦于最相关的 patch，在仅有包级标签的任务中提升可解释性和分类性能。
	In the MIL framework for pathology image segmentation, we first extract feature representations through a series of Fmaba blocks, resulting in a feature tensor of shape $B \times L \times D$, where $L$ denotes the number of patches extracted from the image along with a CLS token. To focus on local tissue characteristics, we remove the CLS token, obtaining a refined feature tensor of shape $B \times (L-1) \times D$, where each of the $L-1$ feature vectors corresponds to a distinct patch within the pathology image:
	
	\begin{figure}[!t]
		\centering
		\includegraphics[width=\linewidth]{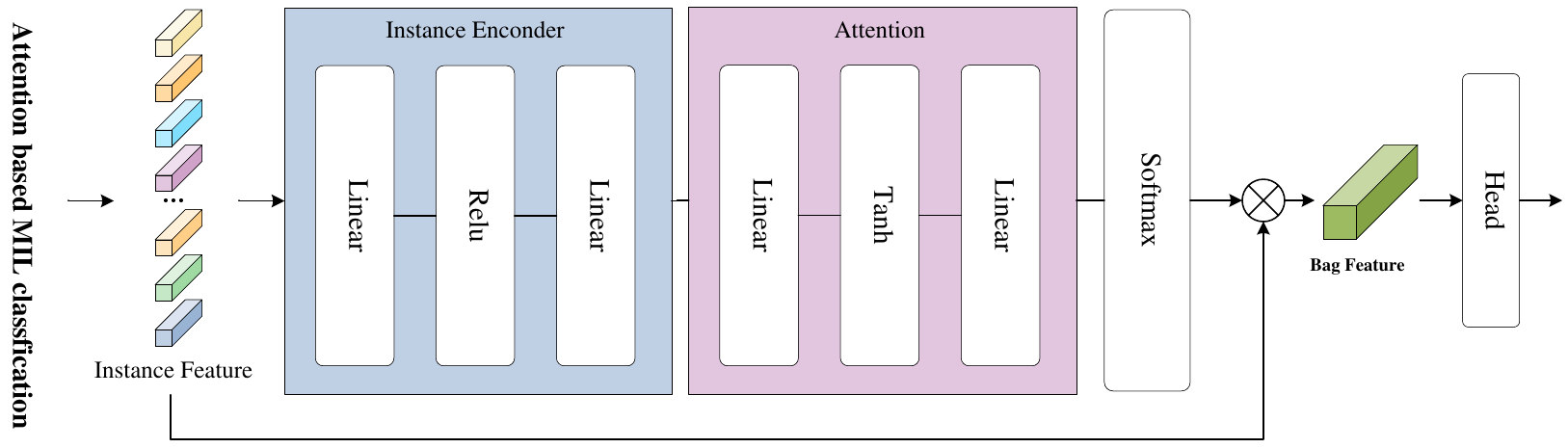}
		\caption{\small{Instance-level to Bag-level Classification Head Block Structure.}}
		\label{clshead}
	\end{figure}
	
	\begin{equation}
		X = [x_1, x_2, \dots, x_{L-1}] \in \mathbb{R}^{B \times (L-1) \times D}
	\end{equation}
	
	To construct a bag-level representation from these instance-level features, we employ a shared instance encoder that maps each patch feature into a fixed-dimensional latent space. This encoder consists of linear layers followed by non-linear activation functions, such as ReLU, to enhance feature expressiveness and capture underlying pathological patterns. Specifically, instance-level features undergo a linear transformation $f_{\text{enc}}(\cdot)$ for feature mapping:
	
	\begin{equation}
		Z_i = f_{\text{enc}}(x_i) = \sigma(W_{\text{enc}} x_i + b_{\text{enc}})
	\end{equation}
	
	where $W_{\text{enc}}$ and $b_{\text{enc}}$ are learnable parameters of the linear transformation, and $\sigma(\cdot)$ represents the ReLU non-linear activation function.
	
	Following instance encoding, an attention mechanism is introduced to learn the significance of each patch in the overall pathology classification task. Specifically, the attention module consists of linear layers and a $\tanh$ activation function to compute the attention scores for each patch:
	
	\begin{equation}
		a_i = \tanh(W_{\text{att}} Z_i + b_{\text{att}})
	\end{equation}
	
	where $W_{\text{att}}$ and $b_{\text{att}}$ are the parameters of the attention module. These scores are then normalized using a softmax function to ensure they sum to one, allowing the model to focus on the most informative pathological regions:
	
	\begin{equation}
		\alpha_i = \frac{\exp(a_i)}{\sum_{j=1}^{L-1} \exp(a_j)}
	\end{equation}
	
	The final bag-level feature is obtained through a weighted sum of the instance features, where the learned attention weights $\alpha_i$ determine the contribution of each patch to the final decision:
	
	\begin{equation}
		Z_{\text{bag}} = \sum_{i=1}^{L-1} \alpha_i Z_i
	\end{equation}
	
	This aggregated feature serves as a comprehensive representation of the entire pathology image and is subsequently processed by the classification head for the final diagnostic prediction. The classification head typically consists of a fully connected layer followed by a softmax function:
	
	\begin{equation}
		y = \text{softmax}(W_{\text{cls}} Z_{\text{bag}} + b_{\text{cls}})
	\end{equation}
	
	where $W_{\text{cls}}$ and $b_{\text{cls}}$ are the parameters of the final classification layer, and $y$ represents the predicted class probabilities.
	
	In WSS tasks, relying solely on bag-level predictions is often insufficient, as it tends to capture only the most discriminative regions while ignoring other informative parts of the instance. Therefore, a well-designed instance-level classifier plays a vital role in enhancing model performance. Inspired by ~\cite{qu2022bi}, we leverage the attention scores derived from the MIL attention mechanism as soft labels to guide the training of the instance classifier. This design encourages the model to focus not only on the most discriminative instances but also on those that are moderately informative.
	
	After obtaining the final class probability distribution $y$ from the bag-level classification head, we optimize the model using the standard cross-entropy loss between the predicted probability and the ground truth label $y_{\text{true}}$:
	
	\begin{equation} \mathcal{L}_{\text{bag}} = -\sum_{c=1}^{C} y_{\text{true}, c} \log y_{c} \end{equation}
	where $C$ is the number of classes, $y_{\text{true}, c}$ represents the one-hot encoded ground truth label, and $y_c$ is the predicted class probability.  
	
	In parallel, the instance classifier is trained using the attention weights as soft supervision. Let $\alpha_i$ denote the attention score of the $i$-th instance, and $\hat{y}_i$ be the predicted class probability of that instance. The instance-level loss is formulated as:
	
	\begin{equation} \mathcal{L}_{\text{instance}} = -\sum_{i=1}^{N} \alpha_i \log \hat{y}_i \end{equation}
	
	The final loss function combines both the bag-level and instance-level objectives:
	
	\begin{equation} \mathcal{L}_{cls} = \lambda \mathcal{L}_{\text{bag}} + (1-\lambda) \mathcal{L}_{\text{instance}} \end{equation}
	
	where $\lambda$ is a balancing coefficient controlling the trade-off between bag and instance classification.

	We employ the Adam optimizer to update the model parameters $\theta$ via backpropagation to minimize the loss:  
	
	\begin{equation}
		\theta \leftarrow \theta - \eta \frac{\partial \mathcal{L}_{cls}}{\partial \theta}
	\end{equation}
	
	where $\eta$ is the learning rate. Through iterative optimization, the model improves classification accuracy, enabling effective pathology image recognition.

	This attention-based MIL classification head enables the model to effectively highlight critical pathological regions, improving interpretability and facilitating accurate lesion identification even in the absence of pixel-level annotations. The architecture of this classification head is illustrated in Fig.~\ref{clshead}, which depicts the instance-to-bag transformation process.

	\subsection{Cam-seg Model}  
	\begin{figure}[!t]
		\centering
		\includegraphics[width=\linewidth]{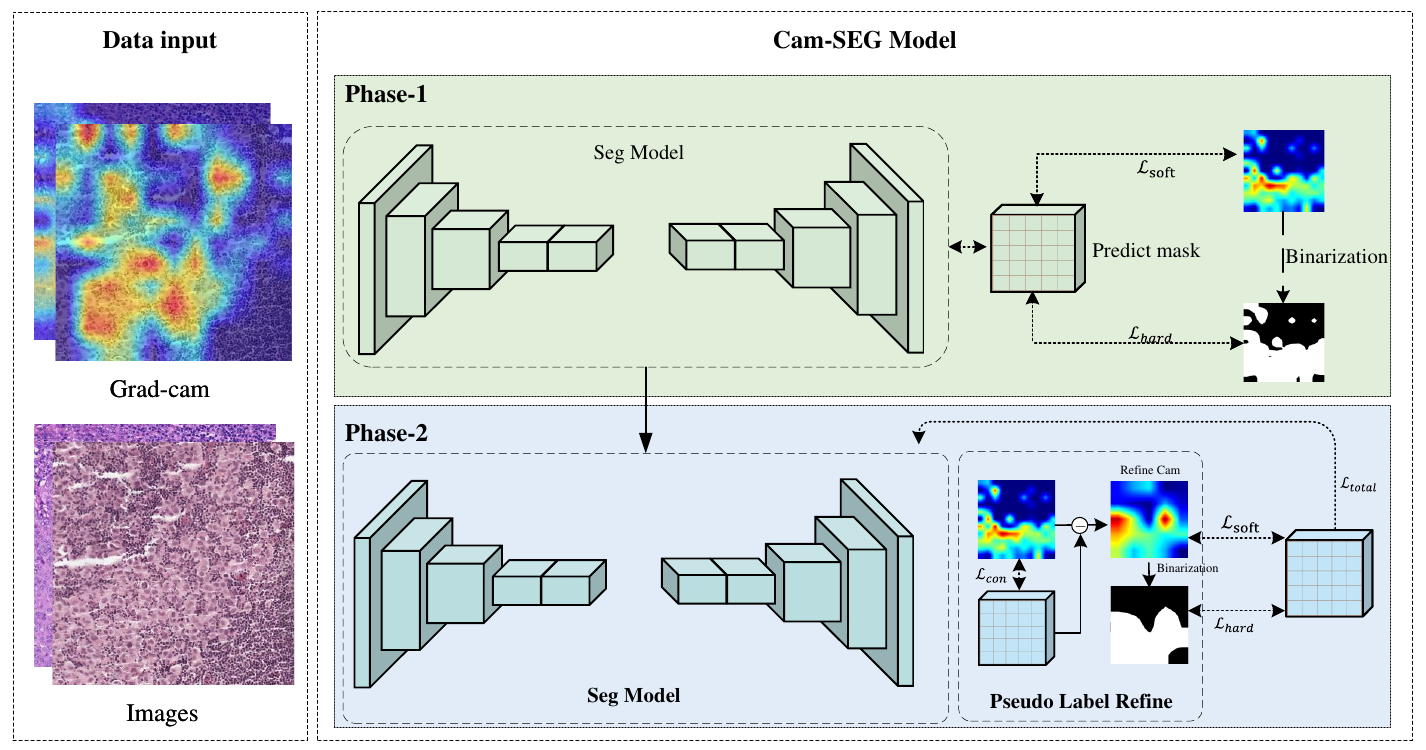}
		\caption{\small{Cam-seg Model Structure.}}
		\label{camseg}
	\end{figure}
	
	%在传统的两阶段弱监督分割任务中，我们通常通过将 CAM 图二值化得到一张伪标签，作为强监督来引导模型学习。然而，这张伪标签往往包含噪声，这些噪声标签可能会对模型产生错误的指导，影响训练的有效性。为了解决这一问题，本文提出了一种基于 CAM 图的伪标签训练策略，旨在增强伪标签的可靠性，从而提升模型的分割效果。
	In traditional two-stage weakly-supervised segmentation tasks, a pseudo-label is typically generated by binarizing the CAM, which is then used as strong supervision to guide the model's learning. However, these pseudo-labels often contain noise, and such noisy labels may mislead the model, thereby affecting the effectiveness of the training process. To address this issue, this paper proposes a pseudo-label training strategy based on the CAM, aiming to enhance the reliability of the pseudo-labels and thus improve the model's segmentation performance. The architecture of Cam-seg Model is illustrated in Fig.~\ref{camseg}

	%本模型采用分期训练策略，在不同的训练阶段使用不同的伪标签生成方法。在初期阶段，伪标签直接由 CAM 图生成；而在后期阶段，逐步结合模型的预测结果修正 CAM 图，最终生成更加精确的伪标签。具体来说，训练过程可以分为以下几个阶段：
	
	The model adopts a staged training strategy, using different pseudo-label generation methods at different training stages. In the early stage, pseudo-labels are directly generated from the CAM. In the later stage, the CAM is progressively refined by incorporating the model's prediction results, ultimately generating more accurate pseudo-labels. Specifically, the training process can be divided into the following stages:
	
	%初期阶段： 在模型的初期阶段，伪标签直接通过 CAM 图生成，作为软伪标签引导学习。在这一阶段，通过计算交叉熵损失 (BCE) 衡量 CAM 图与模型预测结果之间的差异。此时，CAM 图二值化后作为硬标签，使用 Dice 损失和结构感知的mIoU损失来衡量硬标签与模型预测之间的差异。整体损失函数可以表示为和：
	
	\textbf{Early stage}: In the initial stage of the model, pseudo-labels are directly generated from the CAM, serving as soft pseudo-labels to guide learning. During this phase, the difference between the CAM and the model's prediction results is measured using the binary cross-entropy (BCE) loss. At this point, the CAM is binarized and used as a hard label, and the differences between the hard label and the model's predictions are evaluated using the Dice loss and the structure-aware mean Intersection over Union (mIoU) loss. The overall loss function can be expressed as:
	
	\begin{equation}
		\mathcal{L}_{total} = \alpha \ \mathcal{L}_{soft} + (1 - \alpha) \ \mathcal{L}_{hard}
	\end{equation}

	%中后期阶段： 随着模型训练的深入，逐步结合模型的预测结果来修正 CAM 图。在一些区域，模型的预测结果具有较高的置信度时，我们优先使用模型预测结果替代 CAM 图标签，从而减少噪声标签的影响。这一过程有助于发现并纠正 CAM 图中的错误标注或未标注区域。具体地，伪标签的生成采用置信度加权的方式，对于每个像素点，我们根据模型预测的置信度加权选择使用 CAM 图或模型预测作为该像素的标签。假设 $y_1$ 和 $y_2$ 分别为 CAM 图和模型预测的标签，置信度加权伪标签可以表示为：
	
	\textbf{Mid-to-late stage:} As the model training progresses, the CAM is gradually refined by incorporating the model's prediction results. In regions where the model's predictions have high confidence, the model's predictions are prioritized over the CAM labels, thereby reducing the impact of noisy labels. This process helps identify and correct errors or unannotated areas in the CAM. Specifically, the generation of pseudo-labels is performed using confidence-weighted selection. For each pixel, we select the label for that pixel based on a weighted combination of the CAM and the model's prediction, where the weights are determined by the model's prediction confidence $\beta$ ($\beta$ > 0.7). Let $y_{cam}$ and $y_{pred}$ represent the labels from the CAM and the model's predictions, respectively. The confidence-weighted pseudo-label can be expressed as:
	\begin{equation}
		y_{refined} = \beta \ y_{cam} + (1 - \beta) \ y_{pred}
	\end{equation}

	%值得注意的是，为了更好地约束模型，避免错误的伪标签修正，我们引入了一致性损失。对于 CAM 图中的高响应区域，模型需要保持较高的预测准确度。从而确保模型不会在这些重要区域出现过度纠正。该损失函数可以表述为：
	
	It is worth noting that, to better constrain the model and avoid incorrect pseudo-label corrections, we introduce a consistency loss. For the high-response areas in the CAM, the model is required to maintain high prediction accuracy, ensuring that the model does not overcorrect in these critical regions. This loss function can be expressed as:
	
	\begin{equation}
		\mathcal{L}_{\text{con}} = - \frac{1}{N} \sum_{i=1}^{N} \left[ y_i \log(\hat{y}_{ i}) + (1 - y_i) \log(1 - \hat{y}_{i}) \right]
	\end{equation}
	Where $y_i$ is the high-response region in the CAM map, and $\hat{y}_{ i}$
	is the predicted probability value of the model for the corresponding region.

	%模型在中后期阶段的总损失函数可以表达为：
	The total loss function of the model in the mid-to-late stage can be expressed as:
	
	\begin{equation}
		\mathcal{L}_{total} = \alpha \ \mathcal{L}_{soft} + (1 - \alpha) \ \mathcal{L}_{hard} +  \mathcal{L}_{con}
	\end{equation}

	%通过这种训练策略，我们不仅能够引入更强的结构感知能力，还能确保模型在伪标签生成过程中保持对低置信度区域的稳定性。这使得模型能够在强噪声环境下仍然有效地进行训练，并引导模型在复杂场景下进行更精准的分割。
	
	Through this training strategy, we are able to introduce stronger structural awareness while ensuring that the model maintains stability in low-confidence areas during the pseudo-label generation process. This allows the model to be effectively trained even in the presence of strong noise, guiding it to perform more accurate segmentation in complex scenarios.

	\section{Experimental results}
	\label{Experiment}
	
	\subsection{Dataset description}
\begin{figure}[!t]
		\centering
		\includegraphics[width=\linewidth]{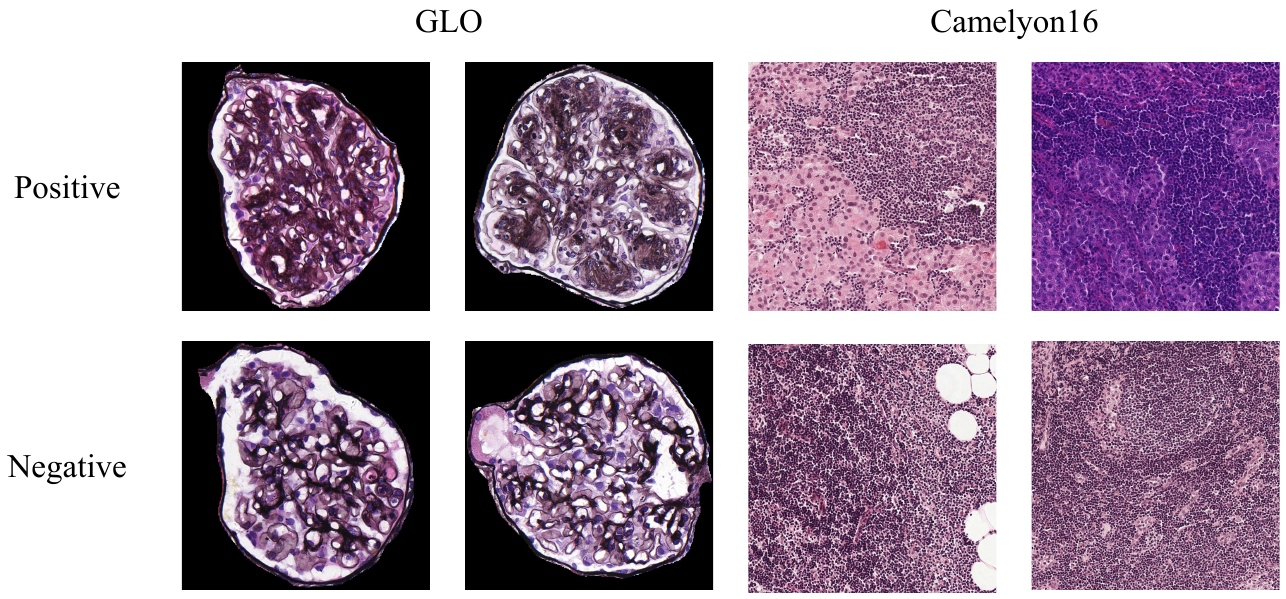}
		\caption{\small{Dataset samples}}
		\label{dataset}
	\end{figure}

	%我们在一个公开可用的数据集（CAMELYON16）和一个私有数据集（肾小球病变）上验证了我们的方法。每个数据集的详细信息如表X所示，具体介绍如下：1）肾小球病变：该数据集包括我们在2017至2020年间从山西省人民医院（SPPH）收集的281例PAS染色肾活检样本，以及在2017至2019年间从山西医科大学第二医院（SHSXMU）收集的30例肾活检样本。经过肾小球分割后，共获得1593个含有K-W结节的阳性肾小球和1660个不含K-W结节的阴性肾小球，这些数据用于训练分类和病变识别模型。每一张肾小球病变的标注均由三位专业病理学家共同完成，数据采集得到了伦理委员会批准（SPPH：No.127，SHSXMU：YX.026）。在我们的评估过程中，选取了398张阳性图像和414张阴性图像作为内部数据集进行验证。2）CAMELYON16数据集：CAMELYON16（Cancer Metastases in Lymph Nodes, 2016）是一个专为癌症转移检测任务设计的病理图像数据集，该数据集由国际医疗影像学会（MICCAI）2016年举办的挑战赛提供，并成为数字病理研究中的常用基准数据集。数据集包含400张来自200名乳腺癌患者的全视野数字病理切片图像，所有图像均使用苏木精和伊红染色，涵盖了丰富的组织细节。每张切片图像均由病理专家进行标注，明确指示出癌症转移的区域。
	We evaluated our method on both a publicly available dataset (CAMELYON16) and a private dataset (glomerular lesions). Detailed information about each dataset is provided in Tab.~\ref{dataset_tab}, some samples are shown in Fig.~\ref{dataset}, with the following descriptions:
	
	\textbf{1) Glomerular Lesions:} This dataset consists of 281 PAS-stained kidney biopsy samples collected from Shanxi Provincial People's Hospital (SPPH) between 2017 and 2020, and 30 kidney biopsy samples collected from the Second Hospital of Shanxi Medical University (SHSXMU) between 2017 and 2019. After glomerular segmentation, a total of 1593 glomeruli containing K-W nodules (positive) and 1660 glomeruli without K-W nodules (negative) were obtained for training classification and lesion detection models. Each glomerular lesion annotation was performed collaboratively by three professional pathologists. Data collection was approved by the ethics committees (SPPH: No.127, SHSXMU: YX.026). For our evaluation, 398 positive images and 414 negative images were selected from the internal dataset.
	
	\textbf{2) CAMELYON16 Dataset\cite{10.1001/jama.2017.14585}:} The CAMELYON16 (Cancer Metastases in Lymph Nodes, 2016) dataset is specifically designed for cancer metastasis detection in pathology images. Provided by the MICCAI 2016 challenge, it has become a widely used benchmark in digital pathology research. The dataset includes 400 whole-slide digital pathology images from 200 breast cancer patients, all stained with hematoxylin and eosin, containing rich tissue details. Each slide image was annotated by pathologists, identifying the regions of cancer metastasis.
	
	%值得注意的是，为了便于测试我们的模型，我们对CAMELYON16数据集进行了预处理。具体而言，我们通过滑动窗口的方式对图像进行了降采样，将每张数字切片划分为更小的patch。每个patch的标签均采用了患者级别的标签，这样处理可以确保标签的合理性与一致性，同时提升了模型的训练效率和测试的便利性。此预处理步骤不仅减少了计算资源的消耗，还使得模型能够更好地捕捉图像中的局部特征，有助于提高分类和转移灶检测的准确性。

	It is worth noting that we performed some preprocessing on the CAMELYON16 dataset to facilitate testing our model. Specifically, we downsampled the images using a sliding window approach, dividing each whole-slide image into smaller patches. The label for each patch was assigned based on the patient-level label, ensuring label consistency and accuracy. This preprocessing step not only reduced the computational resources required but also enabled the model to better capture local features in the images.
	
	% Table generated by Excel2LaTeX from sheet 'dataset'
	\begin{table}[htbp]
		\centering
		\caption{Dataset Information}
		\resizebox{\linewidth}{!}{
			\begin{tabular}{cccc}
				\toprule
				\textbf{Category} & \textbf{Attribute} & \textbf{Glomerular} & \textbf{Camelyon16} \\
				\midrule
				\multirow{6}[2]{*}{\textbf{Dataset Information}} & Source & private\_data & ISBI challenge \\
				& Target & Glomerular spikes & Breast cancer metastases \\
				& Annotation Type & \multicolumn{2}{c}{Patient-level \& Pixel-level (Only Eval)} \\
				& Number of Classes  & \multicolumn{2}{c}{2  (positive \& negative)} \\
				& Input Image Size & \multicolumn{2}{c}{1100 * 1100} \\
				& train:test:val & \multicolumn{2}{c}{7:2:1} \\
				\midrule
				\multirow{2}[2]{*}{\textbf{WSI Distribution}} & Positive WSIs & 212   & 110 \\
				& Negative WSIs & 300   & 160 \\
				\midrule
				\multirow{6}[2]{*}{\textbf{Patch Statistics}} & Total  (Train/test) & 2846/813 & 13184/3776 \\
				& Total  (Val) & 407   & 1883 \\
				& Positive  (Train/test) & 1394/398 & 6410/1832 \\
				& Positive  (Val) & 199   & 916 \\
				& Negative  (Train/test) & 1452/415 & 6774/1944 \\
				& Negative (Val) & 208   & 967 \\
				\bottomrule
			\end{tabular}%
		}
		\label{dataset_tab}%
	\end{table}%

	\subsection{Implementation details}
	%在本研究中，我们在两台不同配置的服务器上进行了实验。第一台服务器配置了6块NVIDIA V100显卡，第二台服务器则配置了一块NVIDIA A6000显卡。两台服务器均安装了Ubuntu 22.04.3 LTS（x86_64）操作系统，操作系统版本为GNU/Linux 6.2.0-35-generic，并且都配置了CUDA 11.8驱动以支持深度学习加速。在实验过程中，所有输入图像都统一调整为1100×1100像素的大小，以确保数据的标准化和一致性，并优化了模型的训练与测试效率。
	
\textbf{Server Configuration:} In this study, we conducted experiments on two servers with different configurations. The first server is equipped with six NVIDIA V100 GPUs, while the second server is equipped with a single NVIDIA A6000 GPU. Both servers are running Ubuntu 22.04.3 LTS (x86\_64) with the GNU/Linux version 6.2.0-35-generic operating system, and they are configured with CUDA 11.8 drivers to support deep learning acceleration. During the experiments, all input images were standardized to a size of 1100 × 1100 pixels to ensure consistency and uniformity in the data, optimizing both the training and testing efficiency of the model.
	
	%所有模型均在训练集上进行训练，并报告验证集上的最佳精度。为了保证公平比较，我们的训练设置遵循了基准模型Vision Mamba的配置。具体而言，我们采用了随机裁剪、随机水平翻转、标签平滑正则化、混合增强和随机擦除等数据增强策略。在训练过程中，我们使用ResNet50作为预训练的特征提取器，提取每个patch的特征。优化器选用AdamW，动量设置为0.9，批次大小为12，权重衰减为0.05。模型训练过程中使用了余弦学习率调度器，初始学习率为$1 \times 10^{-3}$，并应用了EMA（Exponential Moving Average）训练策略，共训练300个epoch。在测试阶段，我们对验证集进行了中心裁剪。为了充分发挥Vision Mamba模型在长序列建模方面的高效性，我们在预训练后进行了微调，微调过程持续30个epoch，并采用了长序列设置。具体而言，我们保持补丁大小不变，学习率设为$10^{-5}$，权重衰减为$10^{-8}$。在实验中，补丁大小是多实例学习（MIL）效果的关键参数，因此我们将补丁大小作为超参数，通过实验调整与验证，最终确定了最优的补丁大小。
	
\textbf{Training settings: }All models were trained on the training set, and the best accuracy on the validation set was reported. To ensure fair comparison, our training setup follows the benchmark model Vision Mamba\cite{zhu2024vision}. Specifically, we applied data augmentation techniques such as random cropping, random horizontal flipping, label smoothing regularization, mixup, and random erasing. During training, we used ResNet50\cite{he2016deep} as a Pre-trained Feature Extraction to extract features for each patch. The optimizer used was AdamW\cite{loshchilov2017decoupled}, with a momentum of 0.9, a batch size of 12, and a weight decay of 0.05. The model training employed a cosine learning rate scheduler with an initial learning rate of $1 \times 10^{-3}$, and the Exponential Moving Average (EMA) training strategy was applied for 300 epochs.
	
	During testing, we applied center cropping to the validation set. To fully leverage the efficient long-sequence modeling capabilities of the FMamba model, we performed fine-tuning after pre-training for 30 epochs, using a long-sequence setting. Specifically, we kept the patch size constant, set the learning rate to $10^{-5}$, and the weight decay to $10^{-8}$. In our experiments, patch size is a critical parameter for the effectiveness of multi-instance learning (MIL). Thus, we treated the patch size as a hyperparameter, adjusting and validating it experimentally to determine the optimal patch size.
	%在第二阶段的 CAM-SEG 分割任务中，我们采用 U-Net 作为网络骨干，训练参数设置为初始学习率 0.01，动量 0.9，权重衰减率 0.0001，进行了 150 轮训练以获得最终的预测模型，其中前 20 轮用于训练的前期调节。我们在实验中分析了不同损失平衡参数设置下模型的表现。
	
	In the second stage of the CAM-SEG segmentation task, we employed U-Net as the network backbone. The training parameters were set with an initial learning rate of 0.01, momentum of 0.9, and weight decay rate of 0.0001, with a total of 150 training epochs to obtain the final prediction model. The first 20 epochs were used for preliminary adjustment during training. We analyzed the model's performance under different loss balancing parameter $\alpha$ settings in the experiments.

	%在本研究中，为全面评估改进的多实例学习在弱监督病变分割任务中的表现，我们采用了多个国际标准的评价指标，包括精度（Accuracy）、召回率（Recall）、精确率（Precision）、F1分数（F1 Score）、交并比（IoU）、Dice系数和平均精度（mAP）。其中，精度衡量分类的整体正确率，召回率和精确率分别评估模型识别病变区域的能力和准确性，F1分数综合考虑了精确率和召回率，IoU和Dice系数则用于衡量分割区域与真实病变区域的重叠程度，mAP则评估模型在多类别任务中的整体性能。此外，我们还通过概率热图可视化，特别关注病变区域的分割质量，以进一步验证模型在实际病变识别中的有效性。
	
\textbf{Performance metrics:} In this study, to comprehensively evaluate the performance of FMaMIL in the weakly-supervised lesion segmentation task, we adopted several internationally recognized evaluation metrics, including Accuracy, F1 Score, Intersection over Union (IoU), Dice coefficient. Additionally, we paid particular attention to the segmentation quality of lesion areas, using probability heatmaps for visualization to further validate the model's effectiveness in real-world lesion recognition.
	
	% Table generated by Excel2LaTeX from sheet 'compare'
	\begin{table*}[htbp]
		\centering
		\caption{Comparison of our model with state-of-the-art methods on two datasets.}
		\resizebox{\linewidth}{!}{
			\begin{tabular}{cccccccccccc}
				 \toprule
				 \multirow{3}[2]{*}{Type} & \multirow{3}[2]{*}{Method} & \multirow{3}[2]{*}{Seg.Net} & \multirow{3}[2]{*}{Publication} & \multicolumn{4}{c}{Glomerular lesion} & \multicolumn{4}{c}{Camelyon16} \\
				 \cmidrule(lr){5-8}\cmidrule(lr){9-12}
				    &       &       &       & \multicolumn{2}{c}{Classfication} & \multicolumn{2}{c}{Segmentation} & \multicolumn{2}{c}{Classfication} & \multicolumn{2}{c}{Segmentation} \\
				    \cmidrule(lr){5-6}\cmidrule(lr){7-8}\cmidrule(lr){9-10}\cmidrule(lr){11-12}

				    &       &       &       & Acc   & AUC   & mIoU  & Dice  & Acc   & AUC   & mIoU  & Dice \\
				    \midrule
				\multirow{4}[1]{*}{WSS} & OAA\cite{jiang2021online}   & -     & ICCV2019 & -     & -     & 0.668  & 0.712  & -     & -     & 0.607  & 0.652  \\
				& Group-WSSS\cite{li2021group} & -     & AAAI2021 & -     & -     & 0.760  & 0.842  & -     & -     & 0.735  & 0.754  \\
				& Auxsegnet\cite{xu2021leveraging} & -     & ICCV2021 & -     & -     & 0.784  & 0.736  & -     & -     & 0.778  & 0.804  \\
				& L2G\cite{jiang2022l2g}   & U-Net & CVPR2022 & -     & -     & 0.786  & 0.863  & -     & -     & 0.786  & 0.844  \\
				\cmidrule(lr){1-12}
				\multirow{6}[0]{*}{MIL-WSS} & DeepAttnMISL\cite{yao2020whole} & -     & MIA2020 & -     & -     & 0.706  & 0.759  & -     & -     & 0.714  & 0.722  \\
				& SA-MIL\cite{li2023weakly} & -     & MIA2023 & -     & -     & 0.813  & 0.891  & -     & -     & 0.795  & 0.876  \\
				& CLAM-SB\cite{lu2021data} & -     & \multirow{2}[0]{*}{Nat. Biomed. Eng2021} & 0.940  & 0.952  & 0.728  & 0.802  & 0.873  & 0.884  & 0.717  & 0.734  \\
				& CLAM-MB\cite{lu2021data} & -     &       & 0.951  & 0.954  & 0.742  & 0.815  & 0.894  & 0.903  & 0.725  & 0.746  \\
				& AB-MIL\cite{ilse2018attention} & -     & PMLR2018 & 0.923  & 0.930  & 0.721  & 0.790  & 0.885  & 0.889  & 0.703  & 0.709  \\
				& Trans-MIL\cite{shao2021transmil} & -     & NIPS2021 & 0.963  & 0.971  & 0.765  & 0.847  & 0.937  & 0.941  & 0.768  & 0.781  \\
				\cmidrule(lr){1-12}
				\multirow{2}[0]{*}{Cam-base CA} & Resnet34\cite{he2016deep} & U-Net & CVPR2016 & 0.956  & 0.959  & 0.655  & 0.664  & 0.953  & 0.960  & 0.604  & 0.631  \\
				& ConvNeXt\cite{liu2022convnet} & U-Net & CVPR2022 & 0.988  & 0.992  & 0.712  & 0.776  & 0.975  & 0.967  & 0.727  & 0.754  \\
				\cmidrule(lr){1-12}
				\multirow{6}[0]{*}{Mamba} & Vision-mamba\cite{vim} & -     & ICML 2024 & 0.946  & 0.956  & 0.706  & 0.715  & 0.965  & 0.961  & 0.731  & 0.746  \\
				& Vision-mamba\cite{vim} & U-Net & ICML 2024 & 0.946  & 0.956  & 0.734  & 0.760  & 0.965  & 0.961  & 0.755  & 0.763  \\
				& MambaMIL\cite{yang2024mambamil} & -     & MICCAI2024 & 0.993  & 0.995  & 0.831  & 0.860  & 0.986  & 0.989  & 0.814  & 0.838  \\
				& MambaMIL\cite{yang2024mambamil} & U-Net & MICCAI2024 & 0.993  & 0.995  & 0.856  & 0.879  & 0.986  & 0.989  & 0.846  & 0.850  \\
				& MamMIL\cite{fang2024mammil}  & -     & BIBM2024 & 0.990  & 0.993  & 0.825  & 0.861  & 0.981  & 0.986  & 0.807  & 0.834  \\
				& MamMIL\cite{fang2024mammil}  & U-Net & BIBM2024 & 0.990  & 0.993  & 0.853  & 0.873  & 0.981  & 0.986  & 0.843  & 0.852  \\
				\cmidrule(lr){1-12}
				\multirow{3}[0]{*}{\textbf{Ours}} & \textbf{FMaMIL} & -     & -     & 0.996 & 0.998 & 0.849  & 0.891 & 0.993 & 0.992 & 0.836 & 0.884 \\
				& \textbf{FMaMIL} & U-Net & -     & 0.996 & 0.998 & 0.870  & 0.914 & 0.993 & 0.992 & 0.856 & 0.927 \\
				& \textbf{FMaMIL} & Cam-SEG & -     & \textcolor[rgb]{ 1,  0,  0}{\textbf{0.996}} & \textcolor[rgb]{ 1,  0,  0}{\textbf{0.998}} & \textcolor[rgb]{ 1,  0,  0}{\textbf{0.887}} & \textcolor[rgb]{ 1,  0,  0}{\textbf{0.934}} & \textcolor[rgb]{ 1,  0,  0}{\textbf{0.993}} & \textcolor[rgb]{ 1,  0,  0}{\textbf{0.992}} & \textcolor[rgb]{ 1,  0,  0}{\textbf{0.869}} & \textcolor[rgb]{ 1,  0,  0}{\textbf{0.957}} \\
				\cmidrule(lr){1-12}
				\multirow{2}[1]{*}{FSS} & U-Net\cite{ronneberger2015u} & -     & MICCAI2015 & -     & -     & 0.826  & 0.854  & -     & -     & 0.830  & 0.931  \\
				& DeeplabV3+\cite{chen2018encoder} & -     & CVPR2017 & -     & -     & 0.835  & 0.860  & -     & -     & 0.849  & 0.923  \\
				\bottomrule
				
			\end{tabular}%
		}
		\label{compare}%
	\end{table*}%

	\subsection{Comparison with state-of-the-art methods}
	
	%在本节中，我们将我们的模型与当前最先进的分类分割方法进行比较，结果如表 X 所示。所有实验结果均在验证集上进行测试。为了更加客观和精确地衡量模型的性能，我们邀请了专业医生对验证数据集进行了真实且详尽的标注。通过观察实验结果，可以发现，相较于传统的弱监督病变分割方法，我们的模型在性能上有显著提升。
	In this section, we compare our model with the current state-of-the-art classification and segmentation methods, with the results shown in Tab.~\ref{compare}. All experimental results were tested on the validation set. To assess the model's performance more objectively and accurately, we invited professional doctors to provide real and detailed annotations for the validation dataset. By examining the experimental results, it is evident that our model shows significant improvement in performance compared to traditional weakly supervised lesion segmentation methods.

	%具体来说，尽管以多示例学习分类为代表的弱监督分割方法在一定程度上表现出了优势，但由于存在序列依赖性和资源限制，这些方法在性能上仍然存在瓶颈。而我们的模型充分考虑了示例间的相关性，并通过挖掘示例与包之间的关系，提取出了更适用于分割任务的伪标签特征。此外，与采用传统分类模型并利用 CAM 图作为伪标签的两阶段方法相比，我们的方法在性能上实现了显著超越。甚至与基于 MAMBA 的最先进方法相比，我们的模型在分类任务中也取得了更高的得分，且在分割任务中得到了更好的性能指标。这一成果得益于各个模块之间的有效协同配合。
	
	Specifically, although WSS methods, represented by MIL-WSSS, have demonstrated advantages to some extent, these methods still face performance bottlenecks due to sequential dependencies and resource limitations. In contrast, our model fully considers the correlations between instances and, by exploring the relationships between instances and bags, extracts pseudo-label features that are more suitable for segmentation tasks. Furthermore, compared to the two-stage approach that uses traditional classification models and utilizes CAM maps as pseudo-labels, our method achieves significant performance improvement. Even when compared to the state-of-the-art method based on Mamba, our model not only achieves higher scores in the classification task but also delivers better performance metrics in the segmentation task. This achievement is attributed to the effective collaboration between the various modules.
	
	%在与强监督学习方法的比较中，我们的模型展现出了意想不到的优势。我们深入分析了这一结果的原因，发现尽管强监督训练通常依赖于真实标签，但在医学图像这一复杂的任务中，标签往往并不完全详尽，部分病变区域可能未被精准标注，尤其是在稀疏标注情况下。这些标签误差可能导致模型忽略一些病变特征，从而影响训练效果。而我们的弱监督分割模型能够更有效地挖掘图像中的隐性特征，从而提升对稀疏标注数据的处理能力，最终实现了更优的模型性能。
	
	In comparison with fully supervised segmentation (FSS) methods,, our model demonstrates unexpected advantages. We conducted an in-depth analysis of the reasons behind this result and found that, although supervised training typically relies on ground truth labels, in the case of medical images, labels are often not fully comprehensive. Some lesion areas may not be precisely annotated, particularly under sparse annotation conditions. These labeling errors can cause the model to overlook certain lesion features, thereby affecting the training outcome. In contrast, our WSS model is more effective at uncovering latent features in the image, thereby enhancing its ability to handle sparse annotation data, ultimately achieving superior model performance.

	\subsection{Hyper-parameter analysis}
	%在这一部分，我们将分别探索FmambaMIL的分类与分割阶段的关键参数对模型性能的影响，并通过分析这些参数确定最佳值，从而最大化模型效果。
	In this section, we will examine the impact of key parameters of the FMaMIL model on performance during both the classification and segmentation stages. By systematically analyzing these parameters, our goal is to identify their optimal values, thereby optimizing the model's performance and maximizing its efficacy.
	\subsubsection{Hyper-parameters  for Different patch Sizes}

	\begin{figure}[!t]
		\centering
		\includegraphics[width=\linewidth]{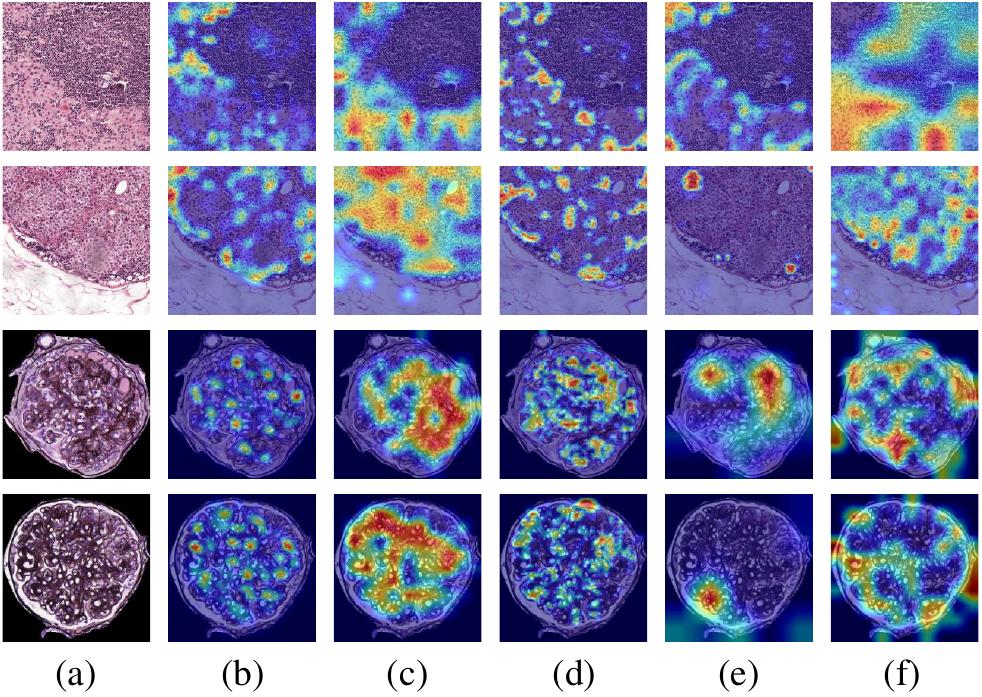}
		\caption{\small{Visualization results of lesion areas with different patch sizes.	(a) Original image; (b) Patch size = 50; (c) Patch size = 100 (d)	Patch size = 150; (e) Patch size = 200; (f) Patch size = 220;}}
		\label{patchimg}
	\end{figure}

	%对于多实例学习来说，我们将输入图像分为若干大小的示例块，通过挖掘示例特征并捕获示例相关性，从而得到模型的预测，选择一个合适的示例大小对于深入理解图像，提取图像的细节特征有着重要的作用。我们在实验中风分别测试了5种不同的尺寸，包括50*50，100*100，150*150，200*200，220*220，对于不能整除的情况，我们使用空白填充的方案。我们在两个数据集上分别测试了分割与分类的结果，实验结果如表x所示，实验结果表明，当patch大小为100*100时，模型达到了最佳的分类分割性能。
	%我们可视化了我们的模型在不同的patch大小大小下的cam图，结果被展示在图x中。当示例较小时，虽然能够提供更精细的局部信息，使得模型能够捕捉微观结构和细粒度特征，但是过小的示例块可能导致全局结构信息的丢失，并增加模型计算开销。相反，较大的示例块能够保留更多的全局语义信息，提高模型对整体模式的理解能力，但可能忽略某些关键的局部细节。以肾小球数据集为例，较小的示例块导致了许多系膜增生的区域并未识别到。而较大的patch导致了几乎所有的肾小球区域都被确定为病变。而在patch大小为100时，cam与真实的结果最为接近，综合性能指标与可视化结果，我们最终选择了patch大小为100作为最佳参数，以确保既能保留足够的细节，又不会影响全局信息的完整性。
	
	In our work, input images are divided into patches of varying sizes, and features are extracted from these patches while capturing inter-instance relationships to generate the final model prediction. Selecting an appropriate patch size is crucial for comprehensively understanding image content and extracting detailed features.
	
	In our experiments, we evaluated five different patch sizes: 50×50, 100×100, 150×150, 200×200, and 220×220. For cases where the image dimensions were not evenly divisible, we applied zero-padding to ensure consistency in input size. The experiments were conducted on two datasets to assess the impact of different patch sizes on segmentation and classification tasks, with the results presented in Tab.~\ref{patch}. The findings indicate that a patch size of 100×100 yielded the best performance in both classification and segmentation tasks.
	
	Furthermore, we visualized the Grad-Cam results under different patch sizes, as shown in Fig.~\ref{patchimg}. The analysis reveals that smaller patches provide finer-grained local information, enabling the model to capture microscopic structures and fine details. However, excessively small patches may lead to the loss of global structural information and significantly increase computational cost. Conversely, larger patches retain more global semantic information, improving the model’s ability to comprehend overall patterns but potentially overlooking critical local details.
	
	Taking the glomerulus dataset as an example, when the patch size was small (e.g., 50×50), many mesangial proliferative regions were not identified, whereas larger patches (e.g., 220×220) resulted in nearly all glomerular regions being misclassified as diseased. In contrast, a patch size of 100×100 produced CAM results that most closely matched the ground truth and achieved the best overall performance. Based on the quantitative experimental results and visualization analysis, we selected 100×100 as the optimal patch size to balance detail preservation and global information integrity.
	
	% Table generated by Excel2LaTeX from sheet 'patch'
	\begin{table}[htbp]
		\centering
		\caption{The performance of FMaMIL with different patch sizes.}
		\resizebox{\linewidth}{!}{
			\begin{tabular}{ccccccccccc}
				\toprule
				\multirow{3}[2]{*}{Patch size} & \multirow{3}[2]{*}{Patch\_size} & \multirow{3}[2]{*}{Num\_patchs} & \multicolumn{4}{c}{Glomerular lesion} & \multicolumn{4}{c}{Camelyon16} \\
				\cmidrule(lr){4-7}\cmidrule(lr){8-11}
				&       &       & \multicolumn{2}{c}{Segmentation} & \multicolumn{2}{c}{Classification} & \multicolumn{2}{c}{Segmentation} & \multicolumn{2}{c}{Classification} \\
				\cmidrule(lr){4-5}\cmidrule(lr){6-7}\cmidrule(lr){8-9}\cmidrule(lr){10-11}
				&       &       & mIoU  & Dice  & Acc   & AUC   & mIoU  & Dice  & Acc   & AUC \\
				\midrule
				50    & 1100  & 22*22 & 0.861 & 0.886 & 0.986 & 0.991 & 0.835 & 0.903 & 0.993 & 0.991 \\
				150   & 1100  &  8*8(padding) & 0.872 & 0.914 & 0.993 & 0.992 & 0.857 & 0.914 & 0.99  & 0.991 \\
				200   & 1100  & 6*6(padding) & 0.79  & 0.891 & 0.991 & 0.994 & 0.803 & 0.868 & 0.987 & 0.989 \\
				220   & 1100  & 5*5   & 0.771 & 0.853   & 0.989 & 0.99  & 0.761 & 0.806 & 0.984 & 0.986 \\
				\cmidrule(lr){1-11}
				FMaMIL(100) & 1100  & 11*11 & \textbf{0.887} & \textbf{0.934} & \textbf{0.996} & \textbf{0.998} & \textbf{0.869} & \textbf{0.957} & \textbf{0.993} & \textbf{0.992} \\
				\bottomrule
			\end{tabular}%
		}
		\label{patch}%
	\end{table}%

	\subsubsection{Hyper-parameters selection for classification Loss of balance $\lambda$}
\begin{figure}[!t]
		\centering
		\includegraphics[width=\linewidth]{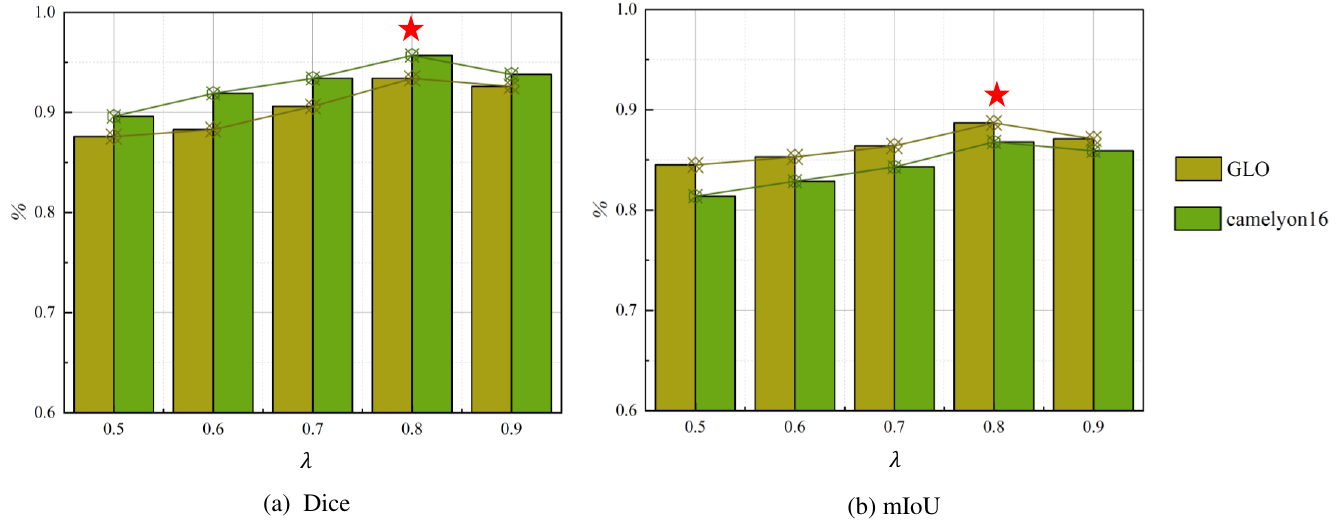}
		\caption{\small{classification Loss of balance $\lambda$}}
		\label{lossinstance}
	\end{figure}

	%在第一阶段的分类任务中，我们设计了示例到包的分类器，分别计算了示例级别的预测损失与包级别的预测损失，之后通过参数lamdba来平衡示例与包的预测影响，我们设计了相关实验来选取最佳的参数lamdba。我们在区间【0.1,0.9】内对lamdba进行调节，由于加入示例级别的损失并不会对模型的分类结果产生显著的影响，因此我们在两个数据集上评估其最终对弱监督分割的影响，实验结果如表x所示，可以观察到，当lamdba=0.2时，取得了最佳的平衡效果。
	In the first-stage classification task, we designed an instance-to-bag classifier that computes both instance-level and bag-level prediction losses. These two losses are balanced using a hyperparameter $\lambda$. We conducted experiments by tuning $\lambda$ within the range of [0.1, 0.9] to determine its optimal value. Since introducing the instance-level loss does not significantly impact the classification performance, we further evaluated its effect on WSS across two datasets. As shown in Fig.~\ref{lossinstance}, the best segmentation performance was achieved when $\lambda = 0.2$, indicating an optimal balance between instance-level learning and bag-level supervision.
	
	\subsubsection{Hyper-parameters for binary threshold $t$}
	%在完成第一阶段的分类任务后，我们利用 Grad-CAM 生成了类别激活图。在第二阶段的监督分割训练过程中，我们需要将其二值化，以生成用于计算硬标签更新损失的掩码。具体而言，当 CAM 中某个像素的值超过阈值 𝑡 时，该像素被归为前景，否则归为背景。阈值 𝑡 的选择对分割结果至关重要，直接影响了第二阶段的分割效果。如果 𝑡 设得过高，可能会导致病变区域的遗漏，而 𝑡 设得过低，则可能引入更多非相关区域，从而影响分割精度。
	
	%为了选择最优的阈值 𝑡，我们在区间 [0,1] 内对其进行调节，并在两个数据集上评估了方法的表现，实验结果如图 x 所示。研究表明，当阈值 𝑡 设为 0.5 时，第二阶段的分割任务在所有数据集上均达到了最佳性能。
	
	After completing the first-stage classification task, we used Grad-CAM to generate CAMs. During the second-stage supervised segmentation training, we needed to binarize these maps to create masks for computing the hard label update loss. Specifically, when a pixel value in the CAM exceeds the threshold $t$, it is classified as foreground; otherwise, it is classified as background. The choice of threshold $t$ is crucial to the segmentation outcome and directly impacts the performance of the second-stage segmentation. If $t$ is set too high, some lesion regions may be missed, whereas if 
	$t$ is set too low, more irrelevant regions may be identified as lesions, reducing segmentation accuracy.
	
\begin{figure}[!t]
		\centering
		\includegraphics[width=\linewidth]{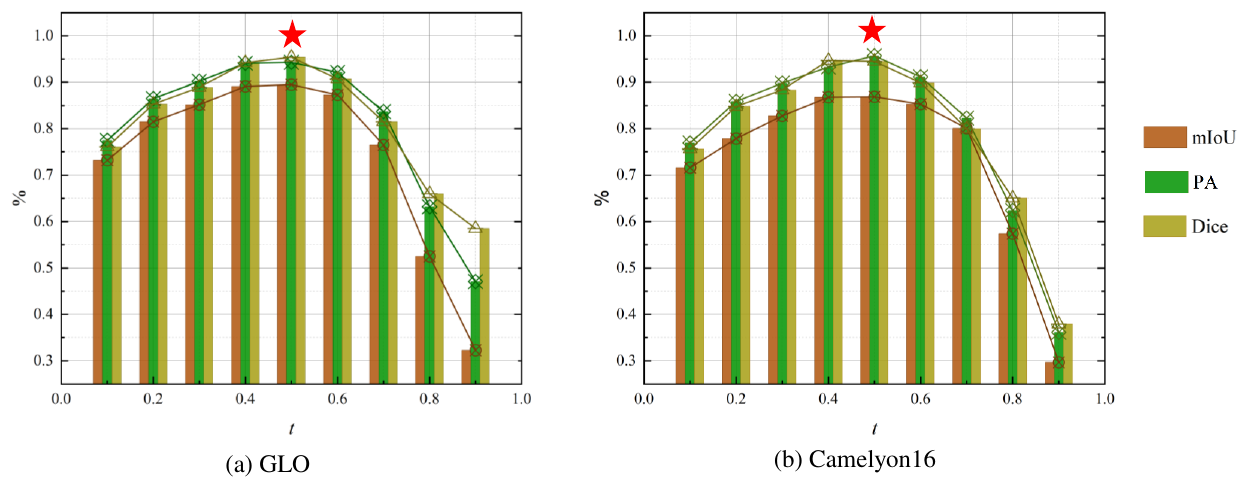}
		\caption{\small{binary threshold $t$}}
		\label{thresholdt}
	\end{figure}

	To determine the optimal threshold $t$, we adjusted its value within the range [0,1] and evaluated the method on two datasets. The experimental results are shown in Fig.~\ref{thresholdt}. The study indicates that when the threshold $t$ is set to 0.5, the second-stage segmentation task achieves the best performance across all datasets.
	
	\subsubsection{Hyper-parameters for Loss of balance  $\alpha$}
	
	In the second-stage segmentation task, we employ both soft label loss, computed from the CAM, and hard label loss, derived from the binarized CAM mask. To balance these two loss components, we introduce the loss balance parameter $\alpha$, which controls the relative contribution of the two terms in the overall loss function. Specifically, a higher $\alpha$ increases the influence of the hard label loss, emphasizing more confident regions, while a lower $\alpha$ gives greater weight to the soft label loss, preserving more uncertainty in the learning process.
	
	The choice of $\alpha$ significantly affects the segmentation performance. A large $\alpha$ may lead to over-reliance on the binarized masks, potentially amplifying errors from thresholding, whereas a small $\alpha$ may cause excessive dependence on the soft labels, reducing the model's ability to refine its predictions.
	
\begin{figure}[!t]
		\centering
		\includegraphics[width=\linewidth]{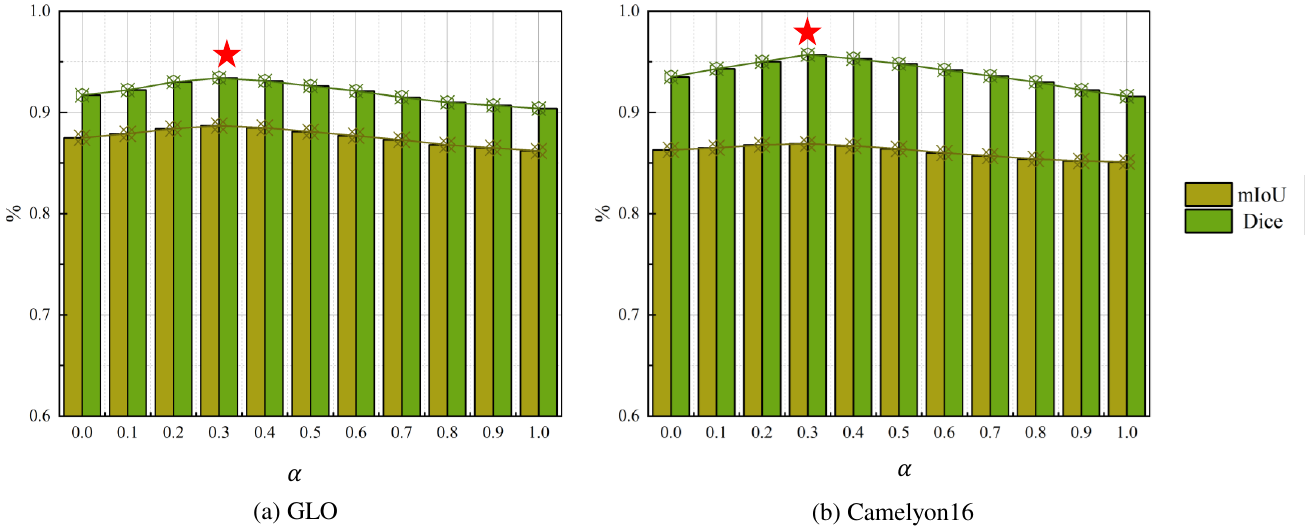}
		\caption{\small{segmentation loss of balance  $\alpha$}}
		\label{segloss}
	\end{figure}

	To determine the optimal $\alpha$, we conduct experiments by varying it within the range [0,1] and evaluate the model performance on two datasets. The results, as shown in Fig.~\ref{segloss}, indicate that setting within the range [0,1] to an 0.3 achieves the best segmentation accuracy across datasets.
	
	\subsection{Visualization comparison}
\begin{figure*}[!t]
		\centering
		\includegraphics[width=\linewidth]{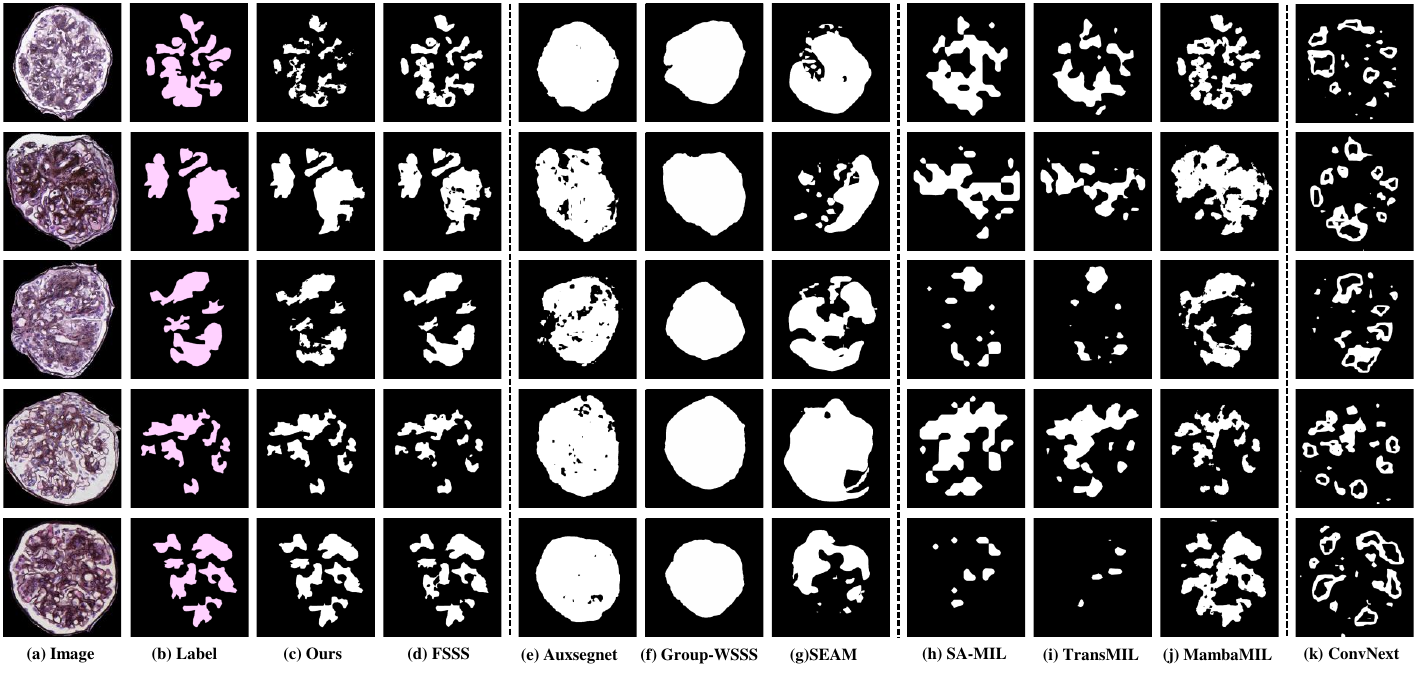}
		\caption{\small{Visualization comparison of lesion segmentation of Glo}}
		\label{gloresult}
	\end{figure*}

\begin{figure*}[!t]
		\centering
		\includegraphics[width=\linewidth]{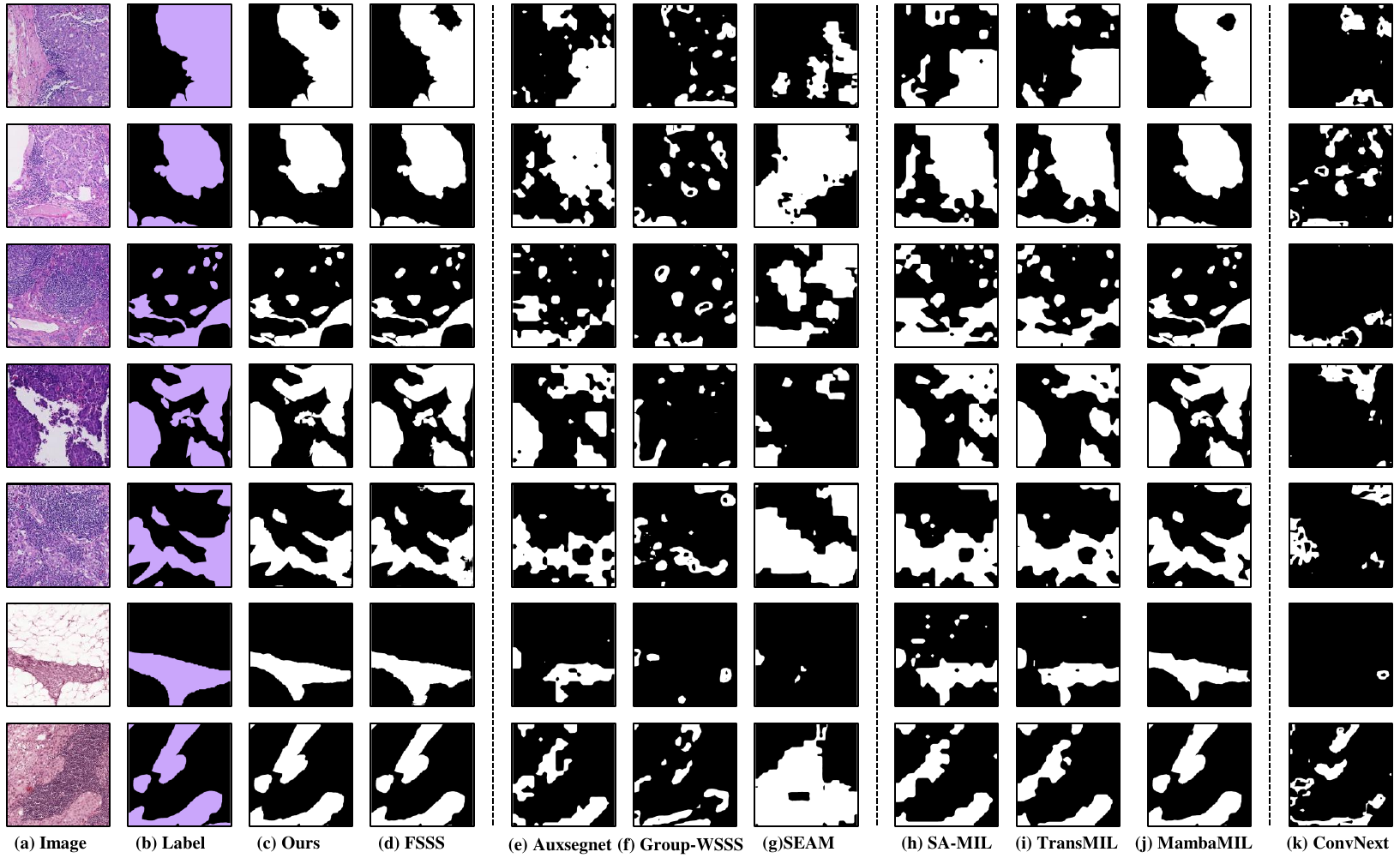}
		\caption{\small{Visualization comparison of lesion segmentation of PatchCamelyon16}}
		\label{cameresult}
	\end{figure*}

	%对于病理图像中的病变识别任务，由于病灶区域通常表现为弥散分布、边界模糊，给精确定位带来了较大挑战。为直观评估所提方法在病变区域定位方面的有效性，我们对部分验证样本进行了分割结果的可视化分析。如图 x 所示，展示了本方法的中间过程、最终分割结果以及与若干先进方法的对比结果。从可视化结果可以看出，所提方法能够更准确地定位病变区域的空间位置，生成的热图在病灶区域呈现出更为清晰且集中的响应。相比传统多实例学习中由于示例之间相互独立所导致的病灶位置偏移、上下文语义信息不完整等问题，我们的方法在空间一致性和语义表达上表现出明显优势。尤其在病灶形态复杂、背景干扰显著的K-W结节识别任务中，本方法仍然保持了良好的定位能力，体现出较强的区域判别能力。值得注意的是，尽管本方法完全基于图像级标签训练，无需任何像素级标注，依然能够获得与专家标注高度一致的分割结果。此外，在专家标注中容易遗漏的微小病灶区域上，所提方法同样表现出较好的感知与定位能力。这些结果不仅进一步验证了模型在弱监督条件下的区域建模能力，也为后续的病灶量化分析与临床辅助诊断提供了可靠的区域信息支持。
	In pathological image analysis, lesion regions are often characterized by diffuse distributions and indistinct boundaries, which pose significant challenges for accurate localization. To qualitatively assess the effectiveness of our proposed method in lesion localization, we visualize the segmentation results on selected validation samples. As illustrated in Fig.~\ref{gloresult} and  Fig.~\ref{cameresult}, we present the intermediate outputs of our method, the final segmentation maps, and comparisons with several state-of-the-art approaches.
	
	The visualizations demonstrate that our method can more precisely localize lesion regions, with the predicted heatmaps exhibiting clearer and more concentrated responses around pathological areas. Compared with conventional MIL methods—where the independence between instances often leads to localization drift and incomplete contextual representation—our approach exhibits superior spatial consistency and semantic relevance. Notably, in the challenging task of K-W nodule recognition, which involves complex lesion morphology and strong background interference, our method maintains robust localization performance, highlighting its strong discriminative capability in spatial regions. It is worth emphasizing that our approach requires only image-level labels during training, yet it achieves lesion predictions that closely align with expert annotations, without any pixel-level supervision.
	
	Furthermore, our method shows promising sensitivity in detecting subtle lesions that are often missed in manual annotations. These results validate the proposed method’s ability to model lesion regions under weak supervision and provide reliable regional information for subsequent quantitative analysis and clinical decision support.

	\subsection{Ablation studies}
	%在本节中，我们将进行消融实验进一步探讨模型各个关键组件对最终分割性能的影响，并分析模型在不同配置下的表现差异。
	In this section, we conduct ablation studies to further investigate the impact of each key component on the final segmentation performance and analyze the model's behavior under different configurations.
	
	\subsubsection{Overall Contribution of Classification Stage}
	%我们首先评估了分类阶段中三个核心模块的性能贡献，包括：（1）双向扫描策略；（2）可学习的频域特征编码；（3）引入双重监督的多实例分类头。我们在两个数据集（Glomerular lesion 和 Camelyon16）上对所提出的方法进行了系统验证。基于Baseline模型 Vision-Mamba，我们依次引入上述三个模块，随后在完整模型的基础上，分别移除各个模块并与完整模型进行对比。实验结果如表 X 所示。
	
	%从结果可以看出，三个关键模块在提升模型整体性能方面均发挥了积极作用。其中，可学习的频域编码显著增强了特征表达能力，是提升最为显著的模块；双向扫描机制有效提升了模型对图像上下文的理解能力；而多实例分类头中的示例级监督进一步优化了CAM的质量，从而提升了分割结果的准确性。相较于原始的Baseline模型，完整模型在分割性能上提升超过15个百分点，充分验证了所提出各个模块在弱监督病变分割任务中的协同增益效果。
	
	We first evaluate the contribution of three core components in the classification stage, including: (1) the bidirectional scanning strategy, (2) the learnable frequency-domain feature encoding, and (3) the dual-supervised multi-instance classification head. The proposed method is systematically validated on two datasets: Glomerular Lesion and Camelyon16. Based on the baseline model Vision-Mamba, we progressively integrate the three components and subsequently perform ablation by individually removing each module from the complete model for comparison. The results are summarized in Tab.~\ref{overallablation}.
	
	The experimental results clearly demonstrate that all three components play an important role in enhancing the overall model performance. Specifically, the learnable frequency-domain encoding significantly improves feature representation and yields the most notable performance gain. The bidirectional scanning strategy enhances contextual understanding, while the instance-level supervision in the MIL head further refines the quality of the CAMs, leading to more accurate segmentation. Compared to the original baseline, the complete model achieves over 15 percentage points improvement in segmentation performance, which strongly validates the effectiveness and synergy of the proposed components in the weakly supervised lesion segmentation task.
	
	\begin{table}[htbp]
		\centering
		\caption{Overall Contribution of Classification Stage Components}
		\resizebox{\linewidth}{!}{
			    \begin{tabular}{ccccccccccccc}
				\toprule
				\multirow{3}[2]{*}{Methods} & \multirow{3}[2]{*}{Two-Scan} & \multirow{3}[2]{*}{Learn-FFT} & \multirow{3}[2]{*}{MIL-Head} & \multirow{3}[2]{*}{Cam-Seg} & \multicolumn{4}{c}{Glomerular lesion} & \multicolumn{4}{c}{Camelyon16} \\
				\cmidrule(lr){6-9}\cmidrule(lr){10-13}
				&       &       &       &       & \multicolumn{2}{c}{Classfication} & \multicolumn{2}{c}{Segmentation} & \multicolumn{2}{c}{Classfication} & \multicolumn{2}{c}{Segmentation} \\
				\cmidrule(lr){6-7}\cmidrule(lr){8-9}\cmidrule(lr){10-11}\cmidrule(lr){12-13}
				&       &       &       &       & Acc   & AUC   & mIoU  & Dice  & Acc   & AUC   & mIoU  & Dice \\
				\midrule
				Baseline(Vision-Mamba) &       &       &       & \checkmark     & 0.946  & 0.956  & 0.734  & 0.760  & 0.965  & 0.961  & 0.755  & 0.763  \\
				Method-A & \checkmark     &       &       & \checkmark     & 0.963  & 0.970  & 0.805  & 0.820  & 0.978  & 0.980  & 0.814  & 0.835  \\
				Method-B &       & \checkmark     &       & \checkmark     & 0.982  & 0.985  & 0.811  & 0.855  & 0.981  & 0.983  & 0.839  & 0.867  \\
				Method-C &       &       & \checkmark     & \checkmark     & 0.959  & 0.965  & 0.763  & 0.801  & 0.972  & 0.974  & 0.781  & 0.809  \\
				\midrule
				FMaMIL(W/O MIL-head) & \checkmark     & \checkmark     &       & \checkmark     & 0.992  & 0.995  & 0.859  & 0.890  & 0.990  & 0.991  & 0.851  & 0.891  \\
				FMaMIL(W/O FFT) & \checkmark     &       & \checkmark     & \checkmark     & 0.983 & 0.988 & 0.843 & 0.875 & 0.985 & 0.987 & 0.846 & 0.909 \\
				FMaMIL(W/O Twoscan) &       & \checkmark     & \checkmark     & \checkmark     & 0.991 & 0.995 & 0.858 & 0.901 & 0.988 & 0.990 & 0.861 & 0.927 \\
				FMaMIL & \checkmark     & \checkmark     & \checkmark     & \checkmark     & \textbf{0.996} & \textbf{0.998} & \textbf{0.887} & \textbf{0.934} & \textbf{0.993} & \textbf{0.992} & \textbf{0.869} & \textbf{0.957} \\
				\bottomrule
			\end{tabular}%
		}
		\label{overallablation}%
	\end{table}%

	\subsubsection{Effect of Learnable Frequency-Domain Encoding}
	%为进一步验证可学习频域编码模块的有效性，我们分别将高频（>0.8）、中频和低频（<0.2）特征单独作为模型输入，以探讨不同频段信息在病变识别与定位任务中的作用。实验结果如表 X 所示。
	
	%结果表明，使用固定频段的频域特征相比于不使用频域编码能够为模型带来一定性能提升，这表明频域信息有助于增强模型的特征判别能力。其中，中频特征对模型性能的提升最为显著，说明其在捕捉病变边缘与结构形态方面具有更强的表达能力。
	
	%相比之下，采用可学习的频域编码能够通过自适应地选择与融合多频段信息，在特征表达和空间定位方面展现出更高的灵活性与准确性，最终获得了最优的分割性能，进一步验证了该模块在弱监督分割任务中的有效性与优势。
	To further verify the effectiveness of the learnable frequency-domain encoding module, we conducted experiments by separately using high-frequency (>0.8), mid-frequency, and low-frequency (<0.2) components as inputs, aiming to investigate the impact of different frequency bands on lesion recognition and localization. The experimental results are presented in Tab.~\ref{freqablation}. The results show that incorporating fixed frequency-domain features brings noticeable performance gains compared to not using frequency encoding, indicating that frequency information can enhance the discriminative power of feature representations. Among the fixed frequency bands, mid-frequency features achieve the most significant improvement, suggesting their strong ability to capture lesion boundaries and structural patterns.
	
	In contrast, the learnable frequency-domain encoding allows the model to adaptively select and fuse multi-band frequency information, offering greater flexibility and precision in feature representation and spatial localization. As a result, it achieves the best segmentation performance, further confirming the effectiveness and superiority of this module in the context of weakly supervised lesion segmentation.
	
	\begin{table}[htbp]
		\centering
		\caption{Effect of Learnable Frequency-Domain Encoding}
		\resizebox{\linewidth}{!}{
			\begin{tabular}{ccccccccccccc}
				\toprule
				\multirow{3}[2]{*}{Methods} & \multirow{3}[2]{*}{Low-freq} & \multirow{3}[2]{*}{Mid-freq} & \multirow{3}[2]{*}{High-freq} & \multirow{3}[2]{*}{Learn-Freq} & \multicolumn{4}{c}{Glomerular lesion} & \multicolumn{4}{c}{Camelyon16} \\
				\cmidrule(lr){6-9}\cmidrule(lr){10-13}
				&       &       &       &       & \multicolumn{2}{c}{Classfication} & \multicolumn{2}{c}{Segmentation} & \multicolumn{2}{c}{Classfication} & \multicolumn{2}{c}{Segmentation} \\
				\cmidrule(lr){6-7}\cmidrule(lr){8-9}\cmidrule(lr){10-11}\cmidrule(lr){12-13}
				&       &       &       &       & ACC   & AUC   & mIoU  & Dice  & ACC   & AUC   & mIoU  & Dice \\
				\midrule
				FMaMIL(W/O FFT) &       &       &       &       & 0.983 & 0.988 & 0.843 & 0.875 & 0.985 & 0.987 & 0.846 & 0.909 \\
				Method-D & \checkmark     &       &       &       & 0.989 & 0.991 & 0.861 & 0.892 & 0.990 & 0.991 & 0.852 & 0.923 \\
				Method-E &       & \checkmark     &       &       & 0.992 & 0.994 & 0.874 & 0.921 & 0.991 & 0.991 & 0.861 & 0.945 \\
				Method-F &       &       & \checkmark     &       & 0.981 & 0.985 & 0.836 & 0.854 & 0.983 & 0.986 & 0.838 & 0.901 \\
				FMaMIL &       &       &       & \checkmark     & \textbf{0.996} & \textbf{0.998} & \textbf{0.887} & \textbf{0.934} & \textbf{0.993} & \textbf{0.992} & \textbf{0.869} & \textbf{0.957} \\
				\bottomrule
			\end{tabular}%
		}
		\label{freqablation}%
	\end{table}%

	\subsubsection{Effect of the Scanning Strategy}
	%为验证不同扫描策略对模型性能的影响，我们分别探讨了是否引入前向与后向的结构状态建模模块（SSM），以及是否采用双向扫描策略。实验结果如表 X 所示。结果表明，前向扫描与后向扫描在信息传递和上下文建模中均具有积极作用，但仅采用单向的逐行扫描策略，其整体性能仍明显落后于双向扫描。这说明双向扫描策略能够有效提升模型对空间结构的一致性建模能力，有助于弥补一维序列在图像空间建模方面的不足，从而为分类阶段提供更清晰、更集中的响应区域。
	To evaluate the impact of different scanning strategies on model performance, we investigate the effects of incorporating forward and backward SSM modules, as well as adopting a bidirectional scanning strategy. The experimental results are presented in Tab.~\ref{scanablation}. The results demonstrate that both forward and backward scanning contribute positively to information propagation and contextual modeling. However, using only a unidirectional row-wise scanning strategy results in inferior performance compared to the bidirectional approach. This indicates that the bidirectional scanning strategy effectively enhances the model’s ability to capture spatial consistency and structural representation, compensating for the spatial information loss inherent in one-dimensional sequences, and providing clearer and more focused activation regions for the classification stage.
	
	\begin{table}[htbp]
		\centering
		\caption{Effect of the Scanning Strategy}
		\resizebox{\linewidth}{!}{
			\begin{tabular}{ccccccccccccc}
				\toprule
				\multirow{3}[2]{*}{Methods} & \multirow{3}[2]{*}{Forward-SSM} & \multirow{3}[2]{*}{Backward-SSM} & \multirow{3}[2]{*}{Single-Scan} & \multirow{3}[2]{*}{Two-Scan} & \multicolumn{4}{c}{Glomerular lesion} & \multicolumn{4}{c}{Camelyon16} \\
				\cmidrule(lr){6-9}\cmidrule(lr){10-13}
				&       &       &       &       & \multicolumn{2}{c}{Classfication} & \multicolumn{2}{c}{Segmentation} & \multicolumn{2}{c}{Classfication} & \multicolumn{2}{c}{Segmentation} \\
				\cmidrule(lr){6-7}\cmidrule(lr){8-9}\cmidrule(lr){10-11}\cmidrule(lr){12-13}
				&       &       &       &       & ACC   & AUC   & mIoU  & Dice  & ACC   & AUC   & mIoU  & Dice \\
				\midrule
				Baseline(W/O B-SSM) & \checkmark     &       & \checkmark     &       & 0.925 & 0.927 & 0.701 & 0.713 & 0.934 & 0.935 & 0.702 & 0.748 \\
				Baseline(W/O F-SSM) &       & \checkmark     & \checkmark     &       & 0.923 & 0.926 & 0.703 & 0.719 & 0.931 & 0.932 & 0.697 & 0.741 \\
				Baseline(Vision-Mamba) & \checkmark     & \checkmark     & \checkmark     &       & 0.946  & 0.956  & 0.734  & 0.760  & 0.965  & 0.961  & 0.755  & 0.763  \\
				\midrule
				Fmamba(W/O B-SSM) & \checkmark     &       &       & \checkmark     & 0.984 & 0.986 & 0.854 & 0.897 & 0.963 & 0.965 & 0.826 & 0.918 \\
				Fmamba(W/O F-SSM) &       & \checkmark     &       & \checkmark     & 0.982 & 0.983 & 0.856 & 0.903 & 0.958 & 0.960 & 0.821 & 0.914 \\
				Fmamba & \checkmark     & \checkmark     &       & \checkmark     & \textbf{0.996} & \textbf{0.998} & \textbf{0.887} & \textbf{0.934} & \textbf{0.993} & \textbf{0.992} & \textbf{0.869} & \textbf{0.957} \\
				\bottomrule
			\end{tabular}%
		}
		\label{scanablation}%
	\end{table}%

	\subsubsection{Effect of Instance-Level Loss in MIL Head}
	%为探究在多示例分割框架中引入示例级监督损失对模型性能的影响，我们设计了相应的消融实验，移除示例级监督损失，仅保留图像级分类目标，其他结构保持一致。实验结果如表 X 所示。可以观察到，示例级监督作为一种细粒度的监督信号，对分类性能影响较小，但在弱监督分割任务中带来了显著性能提升。该损失项有效增强了模型对关键示例的判别能力，从而提升了CAM的生成质量，为后续的分割模块提供了更加准确且集中的引导信息。
	To investigate the impact of introducing instance-level supervision loss in the multi-instance segmentation framework, we conducted an ablation study by removing the instance-level loss while retaining the image-level classification objective, keeping all other components unchanged. As shown in Tab.~\ref{instancelossablation}, the instance-level loss, serving as a fine-grained supervision signal, has limited effect on classification performance but brings significant improvements in the WSS task. This loss effectively enhances the model’s ability to discriminate key instances, thereby improving the quality of the generated CAMs and providing more accurate and focused guidance for the subsequent segmentation module.
	
	\begin{table}[htbp]
		\centering
		\caption{Effect of Instance-Level Loss in MIL Head}
		\resizebox{\linewidth}{!}{
			\begin{tabular}{ccccccc}
				\toprule
				\multirow{2}[2]{*}{Methods} & \multirow{2}[2]{*}{$\mathcal{L}_{\text{bag}}$} & \multirow{2}[2]{*}{$ \mathcal{L}_{\text{instance}}$} & \multicolumn{2}{c}{Glomerular lesion } & \multicolumn{2}{c}{Camelyon16 } \\
				\cmidrule(lr){4-5}\cmidrule(lr){6-7}
				&       &       & mIoU  & Dice  & mIoU  & Dice \\
				\midrule
				Method-G & \checkmark     &       & 0.865 & 0.901 & 0.855 & 0.921 \\
				FMaMIL(ours) & \checkmark     & \checkmark     & \textbf{0.887} & \textbf{0.934} & \textbf{0.869} & \textbf{0.957} \\
				\bottomrule
			\end{tabular}%
		}
		\label{instancelossablation}%
	\end{table}%

	\subsubsection{Overall Contribution in the Segmentation Stage}
	%我们整体评估了第二阶段分割任务中所引入的分阶段训练策略与伪标签优化模块（pseudo label refinement, PLR）的有效性，此外，还对比了不同损失函数对模型性能的影响，实验结果如表 X 所示。当仅使用硬监督损失时，模型只能依赖于初始粗略的伪标签进行训练，导致分割性能明显下降。而引入基于CAM的软标签损失后，能够有效补充病灶区域的语义位置信息，从而提升分割的准确性。分阶段训练策略与PLR模块通过“由粗到细”的优化过程，引导模型逐步收敛，同时动态更新伪标签，有助于提升监督信号的质量与稳定性。此外，引入一致性损失能够在CAM高响应区域内对模型输出施加一致性约束，增强预测结果在关键区域的稳定性与边界清晰度。
	
	We comprehensively evaluate the effectiveness of the staged training strategy(STS) and the pseudo label refinement (PLR) module introduced in the second-stage segmentation task. In addition, we analyze the impact of different loss functions on the model performance. The experimental results are presented in Tab~\ref{segablation}. When only the hard supervision loss is applied, the model relies solely on the initial coarse pseudo labels for training, resulting in a noticeable decline in segmentation performance. In contrast, introducing the CAM-guided soft supervision loss effectively complements the semantic spatial information of lesion regions and significantly improves segmentation accuracy. The staged training strategy, combined with the PLR module, guides the optimization process in a coarse-to-fine manner. The dynamic refinement of pseudo labels during training helps enhance the quality and stability of the supervision signal. Furthermore, the consistency loss constrains the model's predictions in high-response CAM regions, improving the stability of responses and the sharpness of lesion boundaries. 
	
	\begin{table}[htbp]
		\centering
		\caption{Overall Contribution and Effect of Loss Functions in the Segmentation Stage}
		\resizebox{\linewidth}{!}{
			\begin{tabular}{ccc|ccccccc}
				\toprule
				\multirow{2}[2]{*}{Methods} & \multirow{2}[2]{*}{STS} & \multirow{2}[2]{*}{PLR} & \multirow{2}[2]{*}{$\mathcal{L}_{hard}$} & \multirow{2}[2]{*}{$\mathcal{L}_{soft}$} & \multirow{2}[2]{*}{$\mathcal{L}_{con}$} & \multicolumn{2}{c}{Glomerular lesion } & \multicolumn{2}{c}{Camelyon16 } \\
				\cmidrule(lr){7-8}\cmidrule(lr){9-10}
				&       &       &       &       &       & mIoU  & Dice  & mIoU  & Dice \\
				\midrule
				Baseline-A &       &       & \checkmark     &       &       & 0.870 &0.914 & 0.856 & 0.927 \\
				Baseline -B &       &       &       & \checkmark     &       & 0.872 & 0.918 & 0.859 & 0.934 \\
				\cmidrule(lr){2-10}
				FMaMIL(w/o PLR) & \checkmark     &       & \checkmark    & \checkmark     &       & 0.878 & 0.927 & 0.862 & 0.942 \\
				FMaMIL(w/o $\mathcal{L}_{con}$) & \checkmark     & \checkmark     & \checkmark     & \checkmark     &       & 0.883  & 0.926 & 0.865 & 0.946 \\
				FMaMIL(ours) & \checkmark     & \checkmark     & \checkmark     & \checkmark    & \checkmark     & \textbf{0.887} & \textbf{0.934} & \textbf{0.869} & \textbf{0.957} \\
				\bottomrule
			\end{tabular}%
		}
		\label{segablation}%
	\end{table}%

	\section{Discussion}
	\label{discussion}
	%在本研究中，我们提出了一个面向病理图像的两阶段弱监督病变分割框架。该框架遵循“先分类、后分割”的设计思路，旨在充分利用图像级标签条件下的语义信息，以实现精准的病灶定位与分割。在第一阶段的分类任务中，我们基于多示例学习方法，引入了线性注意力结构 Mamba 以挖掘示例间的深层相关性；同时引入可学习的频域编码以增强特征的表达能力，并结合示例级与包级双重监督机制，生成了高质量且响应集中的类激活图（CAM）。在第二阶段的分割任务中，我们采用分阶段训练策略，首先使用基于 CAM 的软标签损失充分利用分类阶段的语义先验，随后引入伪标签优化机制与一致性损失，进一步提升伪标签的准确性与模型预测的稳定性，从而在无需任何像素级标注的前提下，实现了优于现有方法的病灶分割性能。
	
	%在医学图像分析任务中，模型的可解释性对于结果的可信度和临床可用性具有重要意义。在我们的实验中，基于可视化分析与大量定量评估，我们发现模型生成的分割区域与真实病灶位置具有高度一致性，证明了模型确实聚焦于与病变相关的重要区域。尤其值得注意的是，引入的示例级监督机制显著增强了模型对关键区域的感知能力，突破了传统分类网络只能关注于判别性特征的局限，使模型能够发现更多潜在病灶区域。此外，可学习的频域编码与双向扫描策略从频域建模与空间上下文建模两个维度共同提升了模型对复杂病灶区域的识别能力，解决了传统空域方法在边缘与纹理建模中的不足，进一步增强了模型的结构可解释性与稳定性。
	
	%针对现有弱监督方法常将粗糙的CAM图直接作为分割伪标签、而忽视其潜在噪声的问题，我们在本研究中从两个方向进行了改进：一方面优化CAM结构以获得更集中、更语义一致的响应区域；另一方面，我们探索引入软标签损失作为优化目标，并基于模型预测结果动态更新软标签。这一策略显著提升了伪标签的监督质量，表明软监督是缓解标签噪声问题的可行路径。我们认为未来可进一步探索知识蒸馏机制，引入教师–学生网络结构，通过教师模型稳定引导学生模型学习，从而有效缓解噪声伪标签带来的性能偏移问题。
	
	%当然，我们也认识到当前框架仍存在一定的局限性。例如，我们对图像级标签的质量仍较为依赖；在部分边界模糊、形态复杂的病灶区域，模型仍存在响应不够精准的问题。尽管我们通过Mamba结构提升了局部相关性建模能力，但全局结构信息的利用仍较为有限，未来可进一步引入全局建模机制（如长距离注意力或图结构建模）以提升模型对整体图像语义的理解能力。
	
	%综上所述，我们提出的FmaMIL方法通过创新性地融合频域信息与空域结构，在多示例学习框架下有效挖掘了示例之间的内在联系，并通过基于CAM的软监督分割策略，显著提升了弱监督分割的性能。该方法为数字病理图像中的低成本高质量分割提供了一种新的技术路径。未来工作将致力于进一步提升该框架在临床实际场景中的适应性与推广性，推动病理图像智能分析在更多疾病场景中的深入应用。
	
	In this study, we propose a two-stage weakly supervised lesion segmentation framework tailored for pathological images. The framework follows a “classification-then-segmentation” paradigm that leverages image-level annotations to enable precise lesion localization and segmentation. In the first stage, we adopt a MIL strategy, where a linear attention module (Mamba) is introduced to capture correlations between instances. Additionally, a learnable frequency-domain encoding module is employed to enhance feature representation, and dual-level supervision—at both the instance and bag levels—is used to generate high-quality and focused CAMs. In the second stage, we utilize a staged training strategy. Specifically, a soft label loss guided by CAMs is used to exploit semantic priors from the classification stage, followed by pseudo label refinement and consistency loss to progressively improve label quality and prediction stability. As a result, our method achieves superior segmentation performance without relying on any pixel-level annotations.
	
	Model interpretability is of great significance in medical image analysis, especially for ensuring clinical reliability and trust. Through comprehensive visualizations and quantitative evaluations, we observe that the predicted segmentation regions closely align with the actual lesion locations, demonstrating that the model effectively attends to clinically relevant areas. Notably, the introduction of instance-level supervision significantly enhances the model’s ability to detect key regions, overcoming the common limitation of conventional classification networks that tend to focus only on the most discriminative regions. Moreover, the learnable frequency-domain encoding and bidirectional scanning strategies jointly improve the model's perception of complex lesion structures. These modules complement each other by enhancing spatial contextual modeling and capturing detailed frequency information, thereby addressing the limitations of conventional spatial-only approaches and improving both structural interpretability and prediction stability.
	
	While many existing weakly supervised methods directly use coarse CAM maps as segmentation pseudo labels—often resulting in noisy supervision—our framework addresses this from two directions:
	\begin{itemize}
		\item \textbf{Refining CAM Structure}: By enhancing the quality of CAMs, we generate more focused and semantically consistent activation regions, improving the spatial precision of pseudo labels.
		
		\item \textbf{Dynamic Soft Supervision}: We introduce soft label losses and iteratively refine pseudo labels using the model’s own predictions, enabling adaptive learning under label noise.
	\end{itemize}
	
	This dual strategy not only improves the reliability of supervision but also highlights the potential of soft supervision in mitigating the adverse effects of noisy labels. Looking ahead, knowledge distillation—particularly in a teacher–student framework—may provide a more robust solution for further reducing bias introduced by imperfect pseudo annotations.

	Nevertheless, some limitations remain. Our method still relies on relatively accurate image-level labels, and challenges persist in accurately segmenting lesions with ambiguous boundaries or complex morphologies. Although the Mamba structure improves local dependency modeling, global semantic understanding is still limited. Future work may explore global modeling techniques, such as long-range attention or graph-based methods, to further enhance the model’s holistic perception capabilities.
	
	In summary, the proposed FMaMIL method represents a significant advancement in weakly supervised lesion segmentation by integrating frequency- and spatial-domain information within a MIL framework. It effectively models inter-instance correlations and employs CAM-based soft supervision to enable high-quality segmentation using only image-level labels. Beyond achieving state-of-the-art performance, FMaMIL offers a lightweight and generalizable architecture that addresses key challenges in digital pathology, particularly under annotation-scarce conditions. Looking forward, future work will focus on improving clinical applicability and robustness through domain adaptation and knowledge distillation, paving the way for broader deployment in real-world pathological imaging scenarios.
	
	\section{Conclusion}
	\label{Conclusion}
	
	%本文提出了一种面向病理图像的两阶段弱监督病变分割框架 FmaMIL，能够在仅依赖图像级标签的条件下，实现对病灶区域的精准定位与分割。该框架有效融合了频域与空域信息，结合 Mamba 结构在长序列建模方面的优势，在多实例学习范式下充分挖掘示例间的相关性，并基于生成的高质量 CAM 图引导分割过程。同时，本文还初步探索了软标签与自我纠正机制在缓解伪标签噪声方面的潜力。大量实验证明了该方法在多个真实病理图像数据集上的有效性与鲁棒性，为未来实现低成本、高精度的病理图像自动分析提供了一条可行路径。未来的研究将进一步探索知识蒸馏、全局语义建模等方向，以提升该框架在更复杂场景下的泛化能力与适应性，推动弱监督病变分割在数字病理学中的深入应用。
	
	This paper proposes a two-stage weakly supervised lesion segmentation framework, FMaMIL, designed for pathological images. The framework achieves accurate lesion localization and segmentation using only image-level annotations. By effectively integrating frequency-domain and spatial-domain information, and leveraging the long-sequence modeling capabilities of Mamba, the framework captures instance correlations under the MIL paradigm. The high-quality CAMs generated in the classification stage are further utilized to guide the segmentation process. Additionally, we explore the potential of soft label supervision and self-correction mechanisms in mitigating the impact of noisy pseudo labels. Extensive experiments demonstrate the effectiveness and robustness of the proposed method across multiple real-world pathological datasets, providing a practical path toward low-cost and high-precision pathological image analysis. Future work will focus on incorporating knowledge distillation and global semantic modeling to further enhance the generalizability and adaptability of the framework in more complex scenarios, promoting the advancement of weakly supervised lesion segmentation in digital pathology.
	
	\section*{Acknowledgment}
	This work was supported by National Natural Science Foundation of China under Grant No. 61901292, The Natural Science Foundation of Shanxi Province, China under Grant No. 202303021211082. The Graduate Scientific Research and Innovation Project of Shanxi Province, Grant No. RC2400005593.
	
	\printcredits
	
	\newpage
	
	\bibliographystyle{cas-model2-names} % 设置参考文献的样式
	\bibliography{references}

\begin{thebibliography}{76}
\expandafter\ifx\csname natexlab\endcsname\relax\def\natexlab#1{#1}\fi
\providecommand{\url}[1]{\texttt{#1}}
\providecommand{\href}[2]{#2}
\providecommand{\path}[1]{#1}
\providecommand{\DOIprefix}{doi:}
\providecommand{\ArXivprefix}{arXiv:}
\providecommand{\URLprefix}{URL: }
\providecommand{\Pubmedprefix}{pmid:}
\providecommand{\doi}[1]{\href{http://dx.doi.org/#1}{\path{#1}}}
\providecommand{\Pubmed}[1]{\href{pmid:#1}{\path{#1}}}
\providecommand{\bibinfo}[2]{#2}
\ifx\xfnm\relax \def\xfnm[#1]{\unskip,\space#1}\fi
%Type = Inproceedings
\bibitem[{Araslanov and Roth(2020)}]{araslanov2020single}
\bibinfo{author}{Araslanov, N.}, \bibinfo{author}{Roth, S.},
  \bibinfo{year}{2020}.
\newblock \bibinfo{title}{Single-stage semantic segmentation from image
  labels}, in: \bibinfo{booktitle}{Proceedings of the IEEE/CVF conference on
  computer vision and pattern recognition}, pp. \bibinfo{pages}{4253--4262}.
%Type = Article
\bibitem[{Bakalo et~al.(2021)Bakalo, Goldberger and Ben-Ari}]{bakalo2021weakly}
\bibinfo{author}{Bakalo, R.}, \bibinfo{author}{Goldberger, J.},
  \bibinfo{author}{Ben-Ari, R.}, \bibinfo{year}{2021}.
\newblock \bibinfo{title}{Weakly and semi supervised detection in medical
  imaging via deep dual branch net}.
\newblock \bibinfo{journal}{Neurocomputing} \bibinfo{volume}{421},
  \bibinfo{pages}{15--25}.
%Type = Inproceedings
\bibitem[{Bearman et~al.(2016)Bearman, Russakovsky, Ferrari and
  Fei-Fei}]{bearman2016s}
\bibinfo{author}{Bearman, A.}, \bibinfo{author}{Russakovsky, O.},
  \bibinfo{author}{Ferrari, V.}, \bibinfo{author}{Fei-Fei, L.},
  \bibinfo{year}{2016}.
\newblock \bibinfo{title}{What’s the point: Semantic segmentation with point
  supervision}, in: \bibinfo{booktitle}{European conference on computer
  vision}, \bibinfo{organization}{Springer}. pp. \bibinfo{pages}{549--565}.
%Type = Article
\bibitem[{Campanella et~al.(2019)Campanella, Hanna, Geneslaw, Miraflor, Werneck
  Krauss~Silva, Busam, Brogi, Reuter, Klimstra and
  Fuchs}]{campanella2019clinical}
\bibinfo{author}{Campanella, G.}, \bibinfo{author}{Hanna, M.G.},
  \bibinfo{author}{Geneslaw, L.}, \bibinfo{author}{Miraflor, A.},
  \bibinfo{author}{Werneck Krauss~Silva, V.}, \bibinfo{author}{Busam, K.J.},
  \bibinfo{author}{Brogi, E.}, \bibinfo{author}{Reuter, V.E.},
  \bibinfo{author}{Klimstra, D.S.}, \bibinfo{author}{Fuchs, T.J.},
  \bibinfo{year}{2019}.
\newblock \bibinfo{title}{Clinical-grade computational pathology using weakly
  supervised deep learning on whole slide images}.
\newblock \bibinfo{journal}{Nature medicine} \bibinfo{volume}{25},
  \bibinfo{pages}{1301--1309}.
%Type = Article
\bibitem[{Cao et~al.(2024)Cao, Pan, Ren, Lu and Zhang}]{cao2024multi}
\bibinfo{author}{Cao, L.}, \bibinfo{author}{Pan, K.}, \bibinfo{author}{Ren,
  Y.}, \bibinfo{author}{Lu, R.}, \bibinfo{author}{Zhang, J.},
  \bibinfo{year}{2024}.
\newblock \bibinfo{title}{Multi-branch spectral channel attention network for
  breast cancer histopathology image classification}.
\newblock \bibinfo{journal}{Electronics} \bibinfo{volume}{13},
  \bibinfo{pages}{459}.
%Type = Inproceedings
\bibitem[{Chang et~al.(2024)Chang, Zeng, Huang and Ni}]{chang2024net}
\bibinfo{author}{Chang, A.}, \bibinfo{author}{Zeng, J.},
  \bibinfo{author}{Huang, R.}, \bibinfo{author}{Ni, D.}, \bibinfo{year}{2024}.
\newblock \bibinfo{title}{Em-net: Efficient channel and frequency learning with
  mamba for 3d medical image segmentation}, in:
  \bibinfo{booktitle}{International Conference on Medical Image Computing and
  Computer-Assisted Intervention}, \bibinfo{organization}{Springer}. pp.
  \bibinfo{pages}{266--275}.
%Type = Inproceedings
\bibitem[{Chen et~al.(2018)Chen, Zhu, Papandreou, Schroff and
  Adam}]{chen2018encoder}
\bibinfo{author}{Chen, L.C.}, \bibinfo{author}{Zhu, Y.},
  \bibinfo{author}{Papandreou, G.}, \bibinfo{author}{Schroff, F.},
  \bibinfo{author}{Adam, H.}, \bibinfo{year}{2018}.
\newblock \bibinfo{title}{Encoder-decoder with atrous separable convolution for
  semantic image segmentation}, in: \bibinfo{booktitle}{Proceedings of the
  European conference on computer vision (ECCV)}, pp.
  \bibinfo{pages}{801--818}.
%Type = Article
\bibitem[{Cheplygina et~al.(2015)Cheplygina, Tax and
  Loog}]{cheplygina2015multiple}
\bibinfo{author}{Cheplygina, V.}, \bibinfo{author}{Tax, D.M.},
  \bibinfo{author}{Loog, M.}, \bibinfo{year}{2015}.
\newblock \bibinfo{title}{Multiple instance learning with bag dissimilarities}.
\newblock \bibinfo{journal}{Pattern recognition} \bibinfo{volume}{48},
  \bibinfo{pages}{264--275}.
%Type = Inproceedings
\bibitem[{Dai et~al.(2015)Dai, He and Sun}]{dai2015boxsup}
\bibinfo{author}{Dai, J.}, \bibinfo{author}{He, K.}, \bibinfo{author}{Sun, J.},
  \bibinfo{year}{2015}.
\newblock \bibinfo{title}{Boxsup: Exploiting bounding boxes to supervise
  convolutional networks for semantic segmentation}, in:
  \bibinfo{booktitle}{Proceedings of the IEEE international conference on
  computer vision}, pp. \bibinfo{pages}{1635--1643}.
%Type = Article
\bibitem[{Dietterich et~al.(1997)Dietterich, Lathrop and
  Lozano-P{\'e}rez}]{dietterich1997solving}
\bibinfo{author}{Dietterich, T.G.}, \bibinfo{author}{Lathrop, R.H.},
  \bibinfo{author}{Lozano-P{\'e}rez, T.}, \bibinfo{year}{1997}.
\newblock \bibinfo{title}{Solving the multiple instance problem with
  axis-parallel rectangles}.
\newblock \bibinfo{journal}{Artificial intelligence} \bibinfo{volume}{89},
  \bibinfo{pages}{31--71}.
%Type = Inproceedings
\bibitem[{Ding et~al.(2023)Ding, Lu, Cai, Zhang and Shang}]{ding2023slf}
\bibinfo{author}{Ding, H.}, \bibinfo{author}{Lu, J.}, \bibinfo{author}{Cai,
  J.}, \bibinfo{author}{Zhang, Y.}, \bibinfo{author}{Shang, Y.},
  \bibinfo{year}{2023}.
\newblock \bibinfo{title}{Slf-unet: Improved unet for brain mri segmentation by
  combining spatial and low-frequency domain features}, in:
  \bibinfo{booktitle}{Computer Graphics International Conference},
  \bibinfo{organization}{Springer}. pp. \bibinfo{pages}{415--426}.
%Type = Inproceedings
\bibitem[{Dong et~al.(2018)Dong, Kampffmeyer, Liang, Wang, Dai and
  Xing}]{dong2018reinforced}
\bibinfo{author}{Dong, N.}, \bibinfo{author}{Kampffmeyer, M.},
  \bibinfo{author}{Liang, X.}, \bibinfo{author}{Wang, Z.},
  \bibinfo{author}{Dai, W.}, \bibinfo{author}{Xing, E.}, \bibinfo{year}{2018}.
\newblock \bibinfo{title}{Reinforced auto-zoom net: towards accurate and fast
  breast cancer segmentation in whole-slide images}, in:
  \bibinfo{booktitle}{Deep Learning in Medical Image Analysis and Multimodal
  Learning for Clinical Decision Support: 4th International Workshop, DLMIA
  2018, and 8th International Workshop, ML-CDS 2018, Held in Conjunction with
  MICCAI 2018, Granada, Spain, September 20, 2018, Proceedings 4},
  \bibinfo{organization}{Springer}. pp. \bibinfo{pages}{317--325}.
%Type = Article
\bibitem[{Ehteshami~Bejnordi et~al.(2017)Ehteshami~Bejnordi, Veta, Johannes~van
  Diest, van Ginneken, Karssemeijer, Litjens, van~der Laak,  and the
  CAMELYON16~Consortium}]{10.1001/jama.2017.14585}
\bibinfo{author}{Ehteshami~Bejnordi, B.}, \bibinfo{author}{Veta, M.},
  \bibinfo{author}{Johannes~van Diest, P.}, \bibinfo{author}{van Ginneken, B.},
  \bibinfo{author}{Karssemeijer, N.}, \bibinfo{author}{Litjens, G.},
  \bibinfo{author}{van~der Laak, J.A.W.M.}, , \bibinfo{author}{the
  CAMELYON16~Consortium}, \bibinfo{year}{2017}.
\newblock \bibinfo{title}{Diagnostic assessment of deep learning algorithms for
  detection of lymph node metastases in women with breast cancer}.
\newblock \bibinfo{journal}{JAMA} \bibinfo{volume}{318},
  \bibinfo{pages}{2199--2210}.
\newblock \URLprefix \url{https://doi.org/10.1001/jama.2017.14585},
  \DOIprefix\doi{10.1001/jama.2017.14585},
  \href{http://arxiv.org/abs/https://jamanetwork.com/journals/jama/articlepdf/2665774/jama\_ehteshami\_bejnordi\_2017\_oi\_170113.pdf}{\tt
  arXiv:https://jamanetwork.com/journals/jama/articlepdf/2665774/jama\_ehteshami\_bejnordi\_2017\_oi\_170113.pdf}.
%Type = Inproceedings
\bibitem[{Fang et~al.(2024a)Fang, Huang, Tang, Huangfu and Liu}]{fang2024sam}
\bibinfo{author}{Fang, H.}, \bibinfo{author}{Huang, S.}, \bibinfo{author}{Tang,
  W.}, \bibinfo{author}{Huangfu, L.}, \bibinfo{author}{Liu, B.},
  \bibinfo{year}{2024}a.
\newblock \bibinfo{title}{Sam-mil: A spatial contextual aware multiple instance
  learning approach for whole slide image classification}, in:
  \bibinfo{booktitle}{Proceedings of the 32nd ACM International Conference on
  Multimedia}, pp. \bibinfo{pages}{6083--6092}.
%Type = Inproceedings
\bibitem[{Fang et~al.(2024b)Fang, Wang, Zhang, Wang, Zhang, Ji and
  Zhang}]{fang2024mammil}
\bibinfo{author}{Fang, Z.}, \bibinfo{author}{Wang, Y.}, \bibinfo{author}{Zhang,
  Y.}, \bibinfo{author}{Wang, Z.}, \bibinfo{author}{Zhang, J.},
  \bibinfo{author}{Ji, X.}, \bibinfo{author}{Zhang, Y.}, \bibinfo{year}{2024}b.
\newblock \bibinfo{title}{Mammil: Multiple instance learning for whole slide
  images with state space models}, in: \bibinfo{booktitle}{2024 IEEE
  International Conference on Bioinformatics and Biomedicine (BIBM)},
  \bibinfo{organization}{IEEE}. pp. \bibinfo{pages}{3200--3205}.
%Type = Inproceedings
\bibitem[{Feng et~al.(2022)Feng, Ma, Zhang, Zhao, Xia and Tao}]{feng2022fiba}
\bibinfo{author}{Feng, Y.}, \bibinfo{author}{Ma, B.}, \bibinfo{author}{Zhang,
  J.}, \bibinfo{author}{Zhao, S.}, \bibinfo{author}{Xia, Y.},
  \bibinfo{author}{Tao, D.}, \bibinfo{year}{2022}.
\newblock \bibinfo{title}{Fiba: Frequency-injection based backdoor attack in
  medical image analysis}, in: \bibinfo{booktitle}{Proceedings of the IEEE/CVF
  Conference on Computer Vision and Pattern Recognition}, pp.
  \bibinfo{pages}{20876--20885}.
%Type = Article
\bibitem[{Gu and Dao(2023)}]{gu2023mamba}
\bibinfo{author}{Gu, A.}, \bibinfo{author}{Dao, T.}, \bibinfo{year}{2023}.
\newblock \bibinfo{title}{Mamba: Linear-time sequence modeling with selective
  state spaces}.
\newblock \bibinfo{journal}{arXiv preprint arXiv:2312.00752} .
%Type = Article
\bibitem[{Gu et~al.(2020)Gu, Dao, Ermon, Rudra and R{\'e}}]{gu2020hippo}
\bibinfo{author}{Gu, A.}, \bibinfo{author}{Dao, T.}, \bibinfo{author}{Ermon,
  S.}, \bibinfo{author}{Rudra, A.}, \bibinfo{author}{R{\'e}, C.},
  \bibinfo{year}{2020}.
\newblock \bibinfo{title}{Hippo: Recurrent memory with optimal polynomial
  projections}.
\newblock \bibinfo{journal}{Advances in neural information processing systems}
  \bibinfo{volume}{33}, \bibinfo{pages}{1474--1487}.
%Type = Article
\bibitem[{Han et~al.(2022)Han, Wang, Chen, Chen, Guo, Liu, Tang, Xiao, Xu, Xu
  et~al.}]{han2022survey}
\bibinfo{author}{Han, K.}, \bibinfo{author}{Wang, Y.}, \bibinfo{author}{Chen,
  H.}, \bibinfo{author}{Chen, X.}, \bibinfo{author}{Guo, J.},
  \bibinfo{author}{Liu, Z.}, \bibinfo{author}{Tang, Y.}, \bibinfo{author}{Xiao,
  A.}, \bibinfo{author}{Xu, C.}, \bibinfo{author}{Xu, Y.}, et~al.,
  \bibinfo{year}{2022}.
\newblock \bibinfo{title}{A survey on vision transformer}.
\newblock \bibinfo{journal}{IEEE transactions on pattern analysis and machine
  intelligence} \bibinfo{volume}{45}, \bibinfo{pages}{87--110}.
%Type = Article
\bibitem[{Hao et~al.(2024)Hao, He and Hung}]{hao2024t}
\bibinfo{author}{Hao, J.}, \bibinfo{author}{He, L.}, \bibinfo{author}{Hung,
  K.F.}, \bibinfo{year}{2024}.
\newblock \bibinfo{title}{T-mamba: Frequency-enhanced gated long-range
  dependency for tooth 3d cbct segmentation}.
\newblock \bibinfo{journal}{arXiv preprint arXiv:2404.01065} .
%Type = Inproceedings
\bibitem[{He et~al.(2016)He, Zhang, Ren and Sun}]{he2016deep}
\bibinfo{author}{He, K.}, \bibinfo{author}{Zhang, X.}, \bibinfo{author}{Ren,
  S.}, \bibinfo{author}{Sun, J.}, \bibinfo{year}{2016}.
\newblock \bibinfo{title}{Deep residual learning for image recognition}, in:
  \bibinfo{booktitle}{Proceedings of the IEEE conference on computer vision and
  pattern recognition}, pp. \bibinfo{pages}{770--778}.
%Type = Inproceedings
\bibitem[{Hou et~al.(2016)Hou, Samaras, Kurc, Gao, Davis and
  Saltz}]{hou2016patch}
\bibinfo{author}{Hou, L.}, \bibinfo{author}{Samaras, D.},
  \bibinfo{author}{Kurc, T.M.}, \bibinfo{author}{Gao, Y.},
  \bibinfo{author}{Davis, J.E.}, \bibinfo{author}{Saltz, J.H.},
  \bibinfo{year}{2016}.
\newblock \bibinfo{title}{Patch-based convolutional neural network for whole
  slide tissue image classification}, in: \bibinfo{booktitle}{Proceedings of
  the IEEE conference on computer vision and pattern recognition}, pp.
  \bibinfo{pages}{2424--2433}.
%Type = Article
\bibitem[{Hu et~al.(2024)Hu, Daryakenari, Shen, Kawaguchi and
  Karniadakis}]{hu2024state}
\bibinfo{author}{Hu, Z.}, \bibinfo{author}{Daryakenari, N.A.},
  \bibinfo{author}{Shen, Q.}, \bibinfo{author}{Kawaguchi, K.},
  \bibinfo{author}{Karniadakis, G.E.}, \bibinfo{year}{2024}.
\newblock \bibinfo{title}{State-space models are accurate and efficient neural
  operators for dynamical systems}.
\newblock \bibinfo{journal}{arXiv preprint arXiv:2409.03231} .
%Type = Inproceedings
\bibitem[{Ilse et~al.(2018)Ilse, Tomczak and Welling}]{ilse2018attention}
\bibinfo{author}{Ilse, M.}, \bibinfo{author}{Tomczak, J.},
  \bibinfo{author}{Welling, M.}, \bibinfo{year}{2018}.
\newblock \bibinfo{title}{Attention-based deep multiple instance learning}, in:
  \bibinfo{booktitle}{International conference on machine learning},
  \bibinfo{organization}{PMLR}. pp. \bibinfo{pages}{2127--2136}.
%Type = Article
\bibitem[{Jiang et~al.(2021)Jiang, Han, Hou, Cheng and Wei}]{jiang2021online}
\bibinfo{author}{Jiang, P.T.}, \bibinfo{author}{Han, L.H.},
  \bibinfo{author}{Hou, Q.}, \bibinfo{author}{Cheng, M.M.},
  \bibinfo{author}{Wei, Y.}, \bibinfo{year}{2021}.
\newblock \bibinfo{title}{Online attention accumulation for weakly supervised
  semantic segmentation}.
\newblock \bibinfo{journal}{IEEE Transactions on Pattern Analysis and Machine
  Intelligence} \bibinfo{volume}{44}, \bibinfo{pages}{7062--7077}.
%Type = Inproceedings
\bibitem[{Jiang et~al.(2019)Jiang, Hou, Cao, Cheng, Wei and
  Xiong}]{jiang2019integral}
\bibinfo{author}{Jiang, P.T.}, \bibinfo{author}{Hou, Q.}, \bibinfo{author}{Cao,
  Y.}, \bibinfo{author}{Cheng, M.M.}, \bibinfo{author}{Wei, Y.},
  \bibinfo{author}{Xiong, H.K.}, \bibinfo{year}{2019}.
\newblock \bibinfo{title}{Integral object mining via online attention
  accumulation}, in: \bibinfo{booktitle}{Proceedings of the IEEE/CVF
  international conference on computer vision}, pp.
  \bibinfo{pages}{2070--2079}.
%Type = Inproceedings
\bibitem[{Jiang et~al.(2022)Jiang, Yang, Hou and Wei}]{jiang2022l2g}
\bibinfo{author}{Jiang, P.T.}, \bibinfo{author}{Yang, Y.},
  \bibinfo{author}{Hou, Q.}, \bibinfo{author}{Wei, Y.}, \bibinfo{year}{2022}.
\newblock \bibinfo{title}{L2g: A simple local-to-global knowledge transfer
  framework for weakly supervised semantic segmentation}, in:
  \bibinfo{booktitle}{Proceedings of the IEEE/CVF conference on computer vision
  and pattern recognition}, pp. \bibinfo{pages}{16886--16896}.
%Type = Article
\bibitem[{Kanavati et~al.(2020)Kanavati, Toyokawa, Momosaki, Rambeau, Kozuma,
  Shoji, Yamazaki, Takeo, Iizuka and Tsuneki}]{kanavati2020weakly}
\bibinfo{author}{Kanavati, F.}, \bibinfo{author}{Toyokawa, G.},
  \bibinfo{author}{Momosaki, S.}, \bibinfo{author}{Rambeau, M.},
  \bibinfo{author}{Kozuma, Y.}, \bibinfo{author}{Shoji, F.},
  \bibinfo{author}{Yamazaki, K.}, \bibinfo{author}{Takeo, S.},
  \bibinfo{author}{Iizuka, O.}, \bibinfo{author}{Tsuneki, M.},
  \bibinfo{year}{2020}.
\newblock \bibinfo{title}{Weakly-supervised learning for lung carcinoma
  classification using deep learning}.
\newblock \bibinfo{journal}{Scientific reports} \bibinfo{volume}{10},
  \bibinfo{pages}{9297}.
%Type = Article
\bibitem[{Khan et~al.(2022)Khan, Naseer, Hayat, Zamir, Khan and
  Shah}]{khan2022transformers}
\bibinfo{author}{Khan, S.}, \bibinfo{author}{Naseer, M.},
  \bibinfo{author}{Hayat, M.}, \bibinfo{author}{Zamir, S.W.},
  \bibinfo{author}{Khan, F.S.}, \bibinfo{author}{Shah, M.},
  \bibinfo{year}{2022}.
\newblock \bibinfo{title}{Transformers in vision: A survey}.
\newblock \bibinfo{journal}{ACM computing surveys (CSUR)} \bibinfo{volume}{54},
  \bibinfo{pages}{1--41}.
%Type = Inproceedings
\bibitem[{Lerousseau et~al.(2020)Lerousseau, Vakalopoulou, Classe, Adam,
  Battistella, Carr{\'e}, Estienne, Henry, Deutsch and
  Paragios}]{lerousseau2020weakly}
\bibinfo{author}{Lerousseau, M.}, \bibinfo{author}{Vakalopoulou, M.},
  \bibinfo{author}{Classe, M.}, \bibinfo{author}{Adam, J.},
  \bibinfo{author}{Battistella, E.}, \bibinfo{author}{Carr{\'e}, A.},
  \bibinfo{author}{Estienne, T.}, \bibinfo{author}{Henry, T.},
  \bibinfo{author}{Deutsch, E.}, \bibinfo{author}{Paragios, N.},
  \bibinfo{year}{2020}.
\newblock \bibinfo{title}{Weakly supervised multiple instance learning
  histopathological tumor segmentation}, in: \bibinfo{booktitle}{Medical Image
  Computing and Computer Assisted Intervention--MICCAI 2020: 23rd International
  Conference, Lima, Peru, October 4--8, 2020, Proceedings, Part V 23},
  \bibinfo{organization}{Springer}. pp. \bibinfo{pages}{470--479}.
%Type = Inproceedings
\bibitem[{Li et~al.(2025)Li, Li, Wang, He, Wang, Wang and
  Qiao}]{li2025videomamba}
\bibinfo{author}{Li, K.}, \bibinfo{author}{Li, X.}, \bibinfo{author}{Wang, Y.},
  \bibinfo{author}{He, Y.}, \bibinfo{author}{Wang, Y.}, \bibinfo{author}{Wang,
  L.}, \bibinfo{author}{Qiao, Y.}, \bibinfo{year}{2025}.
\newblock \bibinfo{title}{Videomamba: State space model for efficient video
  understanding}, in: \bibinfo{booktitle}{European Conference on Computer
  Vision}, \bibinfo{organization}{Springer}. pp. \bibinfo{pages}{237--255}.
%Type = Article
\bibitem[{Li et~al.(2023)Li, Qian, Han, Eric, Chang, Wei, Lai, Liao, Fan and
  Xu}]{li2023weakly}
\bibinfo{author}{Li, K.}, \bibinfo{author}{Qian, Z.}, \bibinfo{author}{Han,
  Y.}, \bibinfo{author}{Eric, I.}, \bibinfo{author}{Chang, C.},
  \bibinfo{author}{Wei, B.}, \bibinfo{author}{Lai, M.}, \bibinfo{author}{Liao,
  J.}, \bibinfo{author}{Fan, Y.}, \bibinfo{author}{Xu, Y.},
  \bibinfo{year}{2023}.
\newblock \bibinfo{title}{Weakly supervised histopathology image segmentation
  with self-attention}.
\newblock \bibinfo{journal}{Medical Image Analysis} \bibinfo{volume}{86},
  \bibinfo{pages}{102791}.
%Type = Inproceedings
\bibitem[{Li et~al.(2021)Li, Zhou, Li, Zhou and Zhang}]{li2021group}
\bibinfo{author}{Li, X.}, \bibinfo{author}{Zhou, T.}, \bibinfo{author}{Li, J.},
  \bibinfo{author}{Zhou, Y.}, \bibinfo{author}{Zhang, Z.},
  \bibinfo{year}{2021}.
\newblock \bibinfo{title}{Group-wise semantic mining for weakly supervised
  semantic segmentation}, in: \bibinfo{booktitle}{Proceedings of the AAAI
  conference on artificial intelligence}, pp. \bibinfo{pages}{1984--1992}.
%Type = Article
\bibitem[{Liao et~al.(2024)Liao, Hu, Xie and Xia}]{liao2024modeling}
\bibinfo{author}{Liao, Z.}, \bibinfo{author}{Hu, S.}, \bibinfo{author}{Xie,
  Y.}, \bibinfo{author}{Xia, Y.}, \bibinfo{year}{2024}.
\newblock \bibinfo{title}{Modeling annotator preference and stochastic
  annotation error for medical image segmentation}.
\newblock \bibinfo{journal}{Medical Image Analysis} \bibinfo{volume}{92},
  \bibinfo{pages}{103028}.
%Type = Inproceedings
\bibitem[{Lin et~al.(2018)Lin, Chen, Dou, Wang, Qin and Heng}]{lin2018scannet}
\bibinfo{author}{Lin, H.}, \bibinfo{author}{Chen, H.}, \bibinfo{author}{Dou,
  Q.}, \bibinfo{author}{Wang, L.}, \bibinfo{author}{Qin, J.},
  \bibinfo{author}{Heng, P.A.}, \bibinfo{year}{2018}.
\newblock \bibinfo{title}{Scannet: A fast and dense scanning framework for
  metastastic breast cancer detection from whole-slide image}, in:
  \bibinfo{booktitle}{2018 IEEE winter conference on applications of computer
  vision (WACV)}, \bibinfo{organization}{IEEE}. pp. \bibinfo{pages}{539--546}.
%Type = Article
\bibitem[{Litjens et~al.(2017)Litjens, Kooi, Bejnordi, Setio, Ciompi,
  Ghafoorian, Van Der~Laak, Van~Ginneken and S{\'a}nchez}]{litjens2017survey}
\bibinfo{author}{Litjens, G.}, \bibinfo{author}{Kooi, T.},
  \bibinfo{author}{Bejnordi, B.E.}, \bibinfo{author}{Setio, A.A.A.},
  \bibinfo{author}{Ciompi, F.}, \bibinfo{author}{Ghafoorian, M.},
  \bibinfo{author}{Van Der~Laak, J.A.}, \bibinfo{author}{Van~Ginneken, B.},
  \bibinfo{author}{S{\'a}nchez, C.I.}, \bibinfo{year}{2017}.
\newblock \bibinfo{title}{A survey on deep learning in medical image analysis}.
\newblock \bibinfo{journal}{Medical image analysis} \bibinfo{volume}{42},
  \bibinfo{pages}{60--88}.
%Type = Inproceedings
\bibitem[{Liu et~al.(2021a)Liu, Chen, Qin, Dou and Heng}]{liu2021feddg}
\bibinfo{author}{Liu, Q.}, \bibinfo{author}{Chen, C.}, \bibinfo{author}{Qin,
  J.}, \bibinfo{author}{Dou, Q.}, \bibinfo{author}{Heng, P.A.},
  \bibinfo{year}{2021}a.
\newblock \bibinfo{title}{Feddg: Federated domain generalization on medical
  image segmentation via episodic learning in continuous frequency space}, in:
  \bibinfo{booktitle}{Proceedings of the IEEE/CVF conference on computer vision
  and pattern recognition}, pp. \bibinfo{pages}{1013--1023}.
%Type = Article
\bibitem[{Liu et~al.(2021b)Liu, Song, Liu and Zhang}]{liu2021review}
\bibinfo{author}{Liu, X.}, \bibinfo{author}{Song, L.}, \bibinfo{author}{Liu,
  S.}, \bibinfo{author}{Zhang, Y.}, \bibinfo{year}{2021}b.
\newblock \bibinfo{title}{A review of deep-learning-based medical image
  segmentation methods}.
\newblock \bibinfo{journal}{Sustainability} \bibinfo{volume}{13},
  \bibinfo{pages}{1224}.
%Type = Article
\bibitem[{Liu et~al.(2024)Liu, Tian, Zhao, Yu, Xie, Wang, Ye and
  Liu}]{DBLP:journals/corr/abs-2401-10166}
\bibinfo{author}{Liu, Y.}, \bibinfo{author}{Tian, Y.}, \bibinfo{author}{Zhao,
  Y.}, \bibinfo{author}{Yu, H.}, \bibinfo{author}{Xie, L.},
  \bibinfo{author}{Wang, Y.}, \bibinfo{author}{Ye, Q.}, \bibinfo{author}{Liu,
  Y.}, \bibinfo{year}{2024}.
\newblock \bibinfo{title}{Vmamba: Visual state space model}.
\newblock \bibinfo{journal}{CoRR} \bibinfo{volume}{abs/2401.10166}.
\newblock \URLprefix \url{https://doi.org/10.48550/arXiv.2401.10166}.
%Type = Inproceedings
\bibitem[{Liu et~al.(2022)Liu, Mao, Wu, Feichtenhofer, Darrell and
  Xie}]{liu2022convnet}
\bibinfo{author}{Liu, Z.}, \bibinfo{author}{Mao, H.}, \bibinfo{author}{Wu,
  C.Y.}, \bibinfo{author}{Feichtenhofer, C.}, \bibinfo{author}{Darrell, T.},
  \bibinfo{author}{Xie, S.}, \bibinfo{year}{2022}.
\newblock \bibinfo{title}{A convnet for the 2020s}, in:
  \bibinfo{booktitle}{Proceedings of the IEEE/CVF conference on computer vision
  and pattern recognition}, pp. \bibinfo{pages}{11976--11986}.
%Type = Article
\bibitem[{Loshchilov(2017)}]{loshchilov2017decoupled}
\bibinfo{author}{Loshchilov, I.}, \bibinfo{year}{2017}.
\newblock \bibinfo{title}{Decoupled weight decay regularization}.
\newblock \bibinfo{journal}{arXiv preprint arXiv:1711.05101} .
%Type = Article
\bibitem[{Lu et~al.(2021)Lu, Williamson, Chen, Chen, Barbieri and
  Mahmood}]{lu2021data}
\bibinfo{author}{Lu, M.Y.}, \bibinfo{author}{Williamson, D.F.},
  \bibinfo{author}{Chen, T.Y.}, \bibinfo{author}{Chen, R.J.},
  \bibinfo{author}{Barbieri, M.}, \bibinfo{author}{Mahmood, F.},
  \bibinfo{year}{2021}.
\newblock \bibinfo{title}{Data-efficient and weakly supervised computational
  pathology on whole-slide images}.
\newblock \bibinfo{journal}{Nature biomedical engineering} \bibinfo{volume}{5},
  \bibinfo{pages}{555--570}.
%Type = Article
\bibitem[{Maron and Lozano-P{\'e}rez(1997)}]{maron1997framework}
\bibinfo{author}{Maron, O.}, \bibinfo{author}{Lozano-P{\'e}rez, T.},
  \bibinfo{year}{1997}.
\newblock \bibinfo{title}{A framework for multiple-instance learning}.
\newblock \bibinfo{journal}{Advances in neural information processing systems}
  \bibinfo{volume}{10}.
%Type = Article
\bibitem[{Qu et~al.(2024)Qu, Ning, An, Fan, Derr, Liu, Xu and
  Li}]{qu2024survey}
\bibinfo{author}{Qu, H.}, \bibinfo{author}{Ning, L.}, \bibinfo{author}{An, R.},
  \bibinfo{author}{Fan, W.}, \bibinfo{author}{Derr, T.}, \bibinfo{author}{Liu,
  H.}, \bibinfo{author}{Xu, X.}, \bibinfo{author}{Li, Q.},
  \bibinfo{year}{2024}.
\newblock \bibinfo{title}{A survey of mamba}.
\newblock \bibinfo{journal}{arXiv preprint arXiv:2408.01129} .
%Type = Article
\bibitem[{Qu et~al.(2022)Qu, Wang, Song et~al.}]{qu2022bi}
\bibinfo{author}{Qu, L.}, \bibinfo{author}{Wang, M.}, \bibinfo{author}{Song,
  Z.}, et~al., \bibinfo{year}{2022}.
\newblock \bibinfo{title}{Bi-directional weakly supervised knowledge
  distillation for whole slide image classification}.
\newblock \bibinfo{journal}{Advances in Neural Information Processing Systems}
  \bibinfo{volume}{35}, \bibinfo{pages}{15368--15381}.
%Type = Article
\bibitem[{Quellec et~al.(2017)Quellec, Cazuguel, Cochener and
  Lamard}]{quellec2017multiple}
\bibinfo{author}{Quellec, G.}, \bibinfo{author}{Cazuguel, G.},
  \bibinfo{author}{Cochener, B.}, \bibinfo{author}{Lamard, M.},
  \bibinfo{year}{2017}.
\newblock \bibinfo{title}{Multiple-instance learning for medical image and
  video analysis}.
\newblock \bibinfo{journal}{IEEE reviews in biomedical engineering}
  \bibinfo{volume}{10}, \bibinfo{pages}{213--234}.
%Type = Inproceedings
\bibitem[{Ronneberger et~al.(2015)Ronneberger, Fischer and
  Brox}]{ronneberger2015u}
\bibinfo{author}{Ronneberger, O.}, \bibinfo{author}{Fischer, P.},
  \bibinfo{author}{Brox, T.}, \bibinfo{year}{2015}.
\newblock \bibinfo{title}{U-net: Convolutional networks for biomedical image
  segmentation}, in: \bibinfo{booktitle}{Medical image computing and
  computer-assisted intervention--MICCAI 2015: 18th international conference,
  Munich, Germany, October 5-9, 2015, proceedings, part III 18},
  \bibinfo{organization}{Springer}. pp. \bibinfo{pages}{234--241}.
%Type = Article
\bibitem[{Ruan et~al.(2025)Ruan, Gao, Xie and Xiang}]{ruan2025learning}
\bibinfo{author}{Ruan, J.}, \bibinfo{author}{Gao, J.}, \bibinfo{author}{Xie,
  M.}, \bibinfo{author}{Xiang, S.}, \bibinfo{year}{2025}.
\newblock \bibinfo{title}{Learning multi-axis representation in frequency
  domain for medical image segmentation}.
\newblock \bibinfo{journal}{Machine Learning} \bibinfo{volume}{114},
  \bibinfo{pages}{10}.
%Type = Article
\bibitem[{Shao et~al.(2021)Shao, Bian, Chen, Wang, Zhang, Ji
  et~al.}]{shao2021transmil}
\bibinfo{author}{Shao, Z.}, \bibinfo{author}{Bian, H.}, \bibinfo{author}{Chen,
  Y.}, \bibinfo{author}{Wang, Y.}, \bibinfo{author}{Zhang, J.},
  \bibinfo{author}{Ji, X.}, et~al., \bibinfo{year}{2021}.
\newblock \bibinfo{title}{Transmil: Transformer based correlated multiple
  instance learning for whole slide image classification}.
\newblock \bibinfo{journal}{Advances in neural information processing systems}
  \bibinfo{volume}{34}, \bibinfo{pages}{2136--2147}.
%Type = Article
\bibitem[{Vaswani(2017)}]{vaswani2017attention}
\bibinfo{author}{Vaswani, A.}, \bibinfo{year}{2017}.
\newblock \bibinfo{title}{Attention is all you need}.
\newblock \bibinfo{journal}{Advances in Neural Information Processing Systems}
  .
%Type = Article
\bibitem[{Waleffe et~al.(2024)Waleffe, Byeon, Riach, Norick, Korthikanti, Dao,
  Gu, Hatamizadeh, Singh, Narayanan et~al.}]{waleffe2024empirical}
\bibinfo{author}{Waleffe, R.}, \bibinfo{author}{Byeon, W.},
  \bibinfo{author}{Riach, D.}, \bibinfo{author}{Norick, B.},
  \bibinfo{author}{Korthikanti, V.}, \bibinfo{author}{Dao, T.},
  \bibinfo{author}{Gu, A.}, \bibinfo{author}{Hatamizadeh, A.},
  \bibinfo{author}{Singh, S.}, \bibinfo{author}{Narayanan, D.}, et~al.,
  \bibinfo{year}{2024}.
\newblock \bibinfo{title}{An empirical study of mamba-based language models}.
\newblock \bibinfo{journal}{arXiv preprint arXiv:2406.07887} .
%Type = Inproceedings
\bibitem[{Wang et~al.(2017)Wang, Lu, Wang, Feng, Wang, Yin and
  Ruan}]{wang2017learning}
\bibinfo{author}{Wang, L.}, \bibinfo{author}{Lu, H.}, \bibinfo{author}{Wang,
  Y.}, \bibinfo{author}{Feng, M.}, \bibinfo{author}{Wang, D.},
  \bibinfo{author}{Yin, B.}, \bibinfo{author}{Ruan, X.}, \bibinfo{year}{2017}.
\newblock \bibinfo{title}{Learning to detect salient objects with image-level
  supervision}, in: \bibinfo{booktitle}{Proceedings of the IEEE conference on
  computer vision and pattern recognition}, pp. \bibinfo{pages}{136--145}.
%Type = Article
\bibitem[{Wang et~al.(2021)Wang, Li, Wang, Liu, Wang, Tan, Wu, Liu, Sun, Yang
  et~al.}]{wang2021annotation}
\bibinfo{author}{Wang, S.}, \bibinfo{author}{Li, C.}, \bibinfo{author}{Wang,
  R.}, \bibinfo{author}{Liu, Z.}, \bibinfo{author}{Wang, M.},
  \bibinfo{author}{Tan, H.}, \bibinfo{author}{Wu, Y.}, \bibinfo{author}{Liu,
  X.}, \bibinfo{author}{Sun, H.}, \bibinfo{author}{Yang, R.}, et~al.,
  \bibinfo{year}{2021}.
\newblock \bibinfo{title}{Annotation-efficient deep learning for automatic
  medical image segmentation}.
\newblock \bibinfo{journal}{Nature communications} \bibinfo{volume}{12},
  \bibinfo{pages}{5915}.
%Type = Article
\bibitem[{Wang et~al.(2019a)Wang, Yang, Rong, Zhan and
  Xiao}]{wang2019pathology}
\bibinfo{author}{Wang, S.}, \bibinfo{author}{Yang, D.M.},
  \bibinfo{author}{Rong, R.}, \bibinfo{author}{Zhan, X.},
  \bibinfo{author}{Xiao, G.}, \bibinfo{year}{2019}a.
\newblock \bibinfo{title}{Pathology image analysis using segmentation deep
  learning algorithms}.
\newblock \bibinfo{journal}{The American journal of pathology}
  \bibinfo{volume}{189}, \bibinfo{pages}{1686--1698}.
%Type = Article
\bibitem[{Wang et~al.(2019b)Wang, Chen, Gan, Lin, Dou, Tsougenis, Huang, Cai
  and Heng}]{wang2019weakly}
\bibinfo{author}{Wang, X.}, \bibinfo{author}{Chen, H.}, \bibinfo{author}{Gan,
  C.}, \bibinfo{author}{Lin, H.}, \bibinfo{author}{Dou, Q.},
  \bibinfo{author}{Tsougenis, E.}, \bibinfo{author}{Huang, Q.},
  \bibinfo{author}{Cai, M.}, \bibinfo{author}{Heng, P.A.},
  \bibinfo{year}{2019}b.
\newblock \bibinfo{title}{Weakly supervised deep learning for whole slide lung
  cancer image analysis}.
\newblock \bibinfo{journal}{IEEE transactions on cybernetics}
  \bibinfo{volume}{50}, \bibinfo{pages}{3950--3962}.
%Type = Article
\bibitem[{Wang et~al.(2020)Wang, Tang, Chen, Luo, Tang, Ran, Cheung and
  Heng}]{wang2020ud}
\bibinfo{author}{Wang, X.}, \bibinfo{author}{Tang, F.}, \bibinfo{author}{Chen,
  H.}, \bibinfo{author}{Luo, L.}, \bibinfo{author}{Tang, Z.},
  \bibinfo{author}{Ran, A.R.}, \bibinfo{author}{Cheung, C.Y.},
  \bibinfo{author}{Heng, P.A.}, \bibinfo{year}{2020}.
\newblock \bibinfo{title}{Ud-mil: uncertainty-driven deep multiple instance
  learning for oct image classification}.
\newblock \bibinfo{journal}{IEEE journal of biomedical and health informatics}
  \bibinfo{volume}{24}, \bibinfo{pages}{3431--3442}.
%Type = Article
\bibitem[{Wang et~al.(2024)Wang, Wang, Ding, Li, Wu, Rong, Kong, Huang, Li,
  Yang et~al.}]{wang2024state}
\bibinfo{author}{Wang, X.}, \bibinfo{author}{Wang, S.}, \bibinfo{author}{Ding,
  Y.}, \bibinfo{author}{Li, Y.}, \bibinfo{author}{Wu, W.},
  \bibinfo{author}{Rong, Y.}, \bibinfo{author}{Kong, W.},
  \bibinfo{author}{Huang, J.}, \bibinfo{author}{Li, S.}, \bibinfo{author}{Yang,
  H.}, et~al., \bibinfo{year}{2024}.
\newblock \bibinfo{title}{State space model for new-generation network
  alternative to transformers: A survey}.
\newblock \bibinfo{journal}{arXiv preprint arXiv:2404.09516} .
%Type = Article
\bibitem[{Wang et~al.(2018)Wang, Yan, Tang, Bai and Liu}]{wang2018revisiting}
\bibinfo{author}{Wang, X.}, \bibinfo{author}{Yan, Y.}, \bibinfo{author}{Tang,
  P.}, \bibinfo{author}{Bai, X.}, \bibinfo{author}{Liu, W.},
  \bibinfo{year}{2018}.
\newblock \bibinfo{title}{Revisiting multiple instance neural networks}.
\newblock \bibinfo{journal}{Pattern recognition} \bibinfo{volume}{74},
  \bibinfo{pages}{15--24}.
%Type = Inproceedings
\bibitem[{Wei et~al.(2017)Wei, Feng, Liang, Cheng, Zhao and
  Yan}]{wei2017object}
\bibinfo{author}{Wei, Y.}, \bibinfo{author}{Feng, J.}, \bibinfo{author}{Liang,
  X.}, \bibinfo{author}{Cheng, M.M.}, \bibinfo{author}{Zhao, Y.},
  \bibinfo{author}{Yan, S.}, \bibinfo{year}{2017}.
\newblock \bibinfo{title}{Object region mining with adversarial erasing: A
  simple classification to semantic segmentation approach}, in:
  \bibinfo{booktitle}{Proceedings of the IEEE conference on computer vision and
  pattern recognition}, pp. \bibinfo{pages}{1568--1576}.
%Type = Inproceedings
\bibitem[{Wu et~al.(2024)Wu, Ye, Yang, Li and Li}]{wu2024dupl}
\bibinfo{author}{Wu, Y.}, \bibinfo{author}{Ye, X.}, \bibinfo{author}{Yang, K.},
  \bibinfo{author}{Li, J.}, \bibinfo{author}{Li, X.}, \bibinfo{year}{2024}.
\newblock \bibinfo{title}{Dupl: Dual student with trustworthy progressive
  learning for robust weakly supervised semantic segmentation}, in:
  \bibinfo{booktitle}{Proceedings of the IEEE/CVF Conference on Computer Vision
  and Pattern Recognition}, pp. \bibinfo{pages}{3534--3543}.
%Type = Inproceedings
\bibitem[{Xiao et~al.(2015)Xiao, Xia, Yang, Huang and Wang}]{xiao2015learning}
\bibinfo{author}{Xiao, T.}, \bibinfo{author}{Xia, T.}, \bibinfo{author}{Yang,
  Y.}, \bibinfo{author}{Huang, C.}, \bibinfo{author}{Wang, X.},
  \bibinfo{year}{2015}.
\newblock \bibinfo{title}{Learning from massive noisy labeled data for image
  classification}, in: \bibinfo{booktitle}{Proceedings of the IEEE conference
  on computer vision and pattern recognition}, pp. \bibinfo{pages}{2691--2699}.
%Type = Article
\bibitem[{Xu et~al.(2024)Xu, Usuyama, Bagga, Zhang, Rao, Naumann, Wong, Gero,
  Gonz{\'a}lez, Gu et~al.}]{xu2024whole}
\bibinfo{author}{Xu, H.}, \bibinfo{author}{Usuyama, N.},
  \bibinfo{author}{Bagga, J.}, \bibinfo{author}{Zhang, S.},
  \bibinfo{author}{Rao, R.}, \bibinfo{author}{Naumann, T.},
  \bibinfo{author}{Wong, C.}, \bibinfo{author}{Gero, Z.},
  \bibinfo{author}{Gonz{\'a}lez, J.}, \bibinfo{author}{Gu, Y.}, et~al.,
  \bibinfo{year}{2024}.
\newblock \bibinfo{title}{A whole-slide foundation model for digital pathology
  from real-world data}.
\newblock \bibinfo{journal}{Nature} , \bibinfo{pages}{1--8}.
%Type = Inproceedings
\bibitem[{Xu et~al.(2020)Xu, Qin, Sun, Wang, Chen and Ren}]{xu2020learning}
\bibinfo{author}{Xu, K.}, \bibinfo{author}{Qin, M.}, \bibinfo{author}{Sun, F.},
  \bibinfo{author}{Wang, Y.}, \bibinfo{author}{Chen, Y.K.},
  \bibinfo{author}{Ren, F.}, \bibinfo{year}{2020}.
\newblock \bibinfo{title}{Learning in the frequency domain}, in:
  \bibinfo{booktitle}{Proceedings of the IEEE/CVF conference on computer vision
  and pattern recognition}, pp. \bibinfo{pages}{1740--1749}.
%Type = Inproceedings
\bibitem[{Xu et~al.(2021)Xu, Ouyang, Bennamoun, Boussaid, Sohel and
  Xu}]{xu2021leveraging}
\bibinfo{author}{Xu, L.}, \bibinfo{author}{Ouyang, W.},
  \bibinfo{author}{Bennamoun, M.}, \bibinfo{author}{Boussaid, F.},
  \bibinfo{author}{Sohel, F.}, \bibinfo{author}{Xu, D.}, \bibinfo{year}{2021}.
\newblock \bibinfo{title}{Leveraging auxiliary tasks with affinity learning for
  weakly supervised semantic segmentation}, in: \bibinfo{booktitle}{Proceedings
  of the IEEE/CVF international conference on computer vision}, pp.
  \bibinfo{pages}{6984--6993}.
%Type = Article
\bibitem[{Xu et~al.(2017)Xu, Jia, Wang, Ai, Zhang, Lai and Chang}]{xu2017large}
\bibinfo{author}{Xu, Y.}, \bibinfo{author}{Jia, Z.}, \bibinfo{author}{Wang,
  L.B.}, \bibinfo{author}{Ai, Y.}, \bibinfo{author}{Zhang, F.},
  \bibinfo{author}{Lai, M.}, \bibinfo{author}{Chang, E.I.C.},
  \bibinfo{year}{2017}.
\newblock \bibinfo{title}{Large scale tissue histopathology image
  classification, segmentation, and visualization via deep convolutional
  activation features}.
\newblock \bibinfo{journal}{BMC bioinformatics} \bibinfo{volume}{18},
  \bibinfo{pages}{1--17}.
%Type = Article
\bibitem[{Xu et~al.(2014)Xu, Zhu, Eric, Chang, Lai and Tu}]{xu2014weakly}
\bibinfo{author}{Xu, Y.}, \bibinfo{author}{Zhu, J.Y.}, \bibinfo{author}{Eric,
  I.}, \bibinfo{author}{Chang, C.}, \bibinfo{author}{Lai, M.},
  \bibinfo{author}{Tu, Z.}, \bibinfo{year}{2014}.
\newblock \bibinfo{title}{Weakly supervised histopathology cancer image
  segmentation and classification}.
\newblock \bibinfo{journal}{Medical image analysis} \bibinfo{volume}{18},
  \bibinfo{pages}{591--604}.
%Type = Inproceedings
\bibitem[{Yang et~al.(2024)Yang, Wang and Chen}]{yang2024mambamil}
\bibinfo{author}{Yang, S.}, \bibinfo{author}{Wang, Y.}, \bibinfo{author}{Chen,
  H.}, \bibinfo{year}{2024}.
\newblock \bibinfo{title}{Mambamil: Enhancing long sequence modeling with
  sequence reordering in computational pathology}, in:
  \bibinfo{booktitle}{International Conference on Medical Image Computing and
  Computer-Assisted Intervention}, \bibinfo{organization}{Springer}. pp.
  \bibinfo{pages}{296--306}.
%Type = Article
\bibitem[{Yao et~al.(2020)Yao, Zhu, Jonnagaddala, Hawkins and
  Huang}]{yao2020whole}
\bibinfo{author}{Yao, J.}, \bibinfo{author}{Zhu, X.},
  \bibinfo{author}{Jonnagaddala, J.}, \bibinfo{author}{Hawkins, N.},
  \bibinfo{author}{Huang, J.}, \bibinfo{year}{2020}.
\newblock \bibinfo{title}{Whole slide images based cancer survival prediction
  using attention guided deep multiple instance learning networks}.
\newblock \bibinfo{journal}{Medical image analysis} \bibinfo{volume}{65},
  \bibinfo{pages}{101789}.
%Type = Article
\bibitem[{Yue and Li(2024)}]{yue2024medmamba}
\bibinfo{author}{Yue, Y.}, \bibinfo{author}{Li, Z.}, \bibinfo{year}{2024}.
\newblock \bibinfo{title}{Medmamba: Vision mamba for medical image
  classification}.
\newblock \bibinfo{journal}{arXiv preprint arXiv:2403.03849} .
%Type = Inproceedings
\bibitem[{Zhang et~al.(2025)Zhang, Li, Sun, Zheng, Zhu and
  Yang}]{zhang2025attention}
\bibinfo{author}{Zhang, Y.}, \bibinfo{author}{Li, H.}, \bibinfo{author}{Sun,
  Y.}, \bibinfo{author}{Zheng, S.}, \bibinfo{author}{Zhu, C.},
  \bibinfo{author}{Yang, L.}, \bibinfo{year}{2025}.
\newblock \bibinfo{title}{Attention-challenging multiple instance learning for
  whole slide image classification}, in: \bibinfo{booktitle}{European
  Conference on Computer Vision}, \bibinfo{organization}{Springer}. pp.
  \bibinfo{pages}{125--143}.
%Type = Article
\bibitem[{Zhang et~al.(2024)Zhang, Zhang, Wang, Yang, Peng and
  Tong}]{zhang2024mamba2mil}
\bibinfo{author}{Zhang, Y.}, \bibinfo{author}{Zhang, X.},
  \bibinfo{author}{Wang, J.}, \bibinfo{author}{Yang, Y.},
  \bibinfo{author}{Peng, T.}, \bibinfo{author}{Tong, C.}, \bibinfo{year}{2024}.
\newblock \bibinfo{title}{Mamba2mil: State space duality based multiple
  instance learning for computational pathology}.
\newblock \bibinfo{journal}{arXiv preprint arXiv:2408.15032} .
%Type = Article
\bibitem[{Zheng et~al.(2024)Zheng, Sharma, Betke, Beane and
  Kolachalama}]{zheng2024fouriermil}
\bibinfo{author}{Zheng, Y.}, \bibinfo{author}{Sharma, H.},
  \bibinfo{author}{Betke, M.}, \bibinfo{author}{Beane, J.},
  \bibinfo{author}{Kolachalama, V.B.}, \bibinfo{year}{2024}.
\newblock \bibinfo{title}{Fouriermil: Fourier filtering-based multiple instance
  learning for whole slide image analysis}.
\newblock \bibinfo{journal}{bioRxiv} , \bibinfo{pages}{2024--08}.
%Type = Inproceedings
\bibitem[{Zhou et~al.(2016)Zhou, Khosla, Lapedriza, Oliva and
  Torralba}]{zhou2016learning}
\bibinfo{author}{Zhou, B.}, \bibinfo{author}{Khosla, A.},
  \bibinfo{author}{Lapedriza, A.}, \bibinfo{author}{Oliva, A.},
  \bibinfo{author}{Torralba, A.}, \bibinfo{year}{2016}.
\newblock \bibinfo{title}{Learning deep features for discriminative
  localization}, in: \bibinfo{booktitle}{Proceedings of the IEEE conference on
  computer vision and pattern recognition}, pp. \bibinfo{pages}{2921--2929}.
%Type = Article
\bibitem[{Zhou et~al.(2024)Zhou, He, Wu, Yao, Xie and Li}]{zhou2024spatial}
\bibinfo{author}{Zhou, Z.}, \bibinfo{author}{He, A.}, \bibinfo{author}{Wu, Y.},
  \bibinfo{author}{Yao, R.}, \bibinfo{author}{Xie, X.}, \bibinfo{author}{Li,
  T.}, \bibinfo{year}{2024}.
\newblock \bibinfo{title}{Spatial-frequency dual progressive attention network
  for medical image segmentation}.
\newblock \bibinfo{journal}{arXiv preprint arXiv:2406.07952} .
%Type = Article
\bibitem[{Zhu et~al.(2024a)Zhu, Liao, Zhang, Wang, Liu and
  Wang}]{zhu2024vision}
\bibinfo{author}{Zhu, L.}, \bibinfo{author}{Liao, B.}, \bibinfo{author}{Zhang,
  Q.}, \bibinfo{author}{Wang, X.}, \bibinfo{author}{Liu, W.},
  \bibinfo{author}{Wang, X.}, \bibinfo{year}{2024}a.
\newblock \bibinfo{title}{Vision mamba: Efficient visual representation
  learning with bidirectional state space model}.
\newblock \bibinfo{journal}{arXiv preprint arXiv:2401.09417} .
%Type = Article
\bibitem[{Zhu et~al.(2024b)Zhu, Liao, Zhang, Wang, Liu and Wang}]{vim}
\bibinfo{author}{Zhu, L.}, \bibinfo{author}{Liao, B.}, \bibinfo{author}{Zhang,
  Q.}, \bibinfo{author}{Wang, X.}, \bibinfo{author}{Liu, W.},
  \bibinfo{author}{Wang, X.}, \bibinfo{year}{2024}b.
\newblock \bibinfo{title}{Vision mamba: Efficient visual representation
  learning with bidirectional state space model}.
\newblock \bibinfo{journal}{arXiv preprint arXiv:2401.09417} .

\end{thebibliography}

\end{document}